\documentclass[letterpaper, 10 pt, journal, twoside]{IEEEtran}
\IEEEoverridecommandlockouts                              

\usepackage{graphicx}
\usepackage{subcaption}

\usepackage{graphicx}
\usepackage{amsmath}
\usepackage{amsfonts}
\usepackage{amssymb}
\usepackage{textcomp}
\usepackage{array}
\usepackage{booktabs}
\usepackage{pifont}
\usepackage{multirow} 
\usepackage{balance}
\usepackage{tabularx}
\usepackage{caption}
\usepackage{hyperref}
\usepackage{xcolor} 
\hypersetup{hypertex=true,
    colorlinks=true,
    linkcolor=blue,
    anchorcolor=blue,
    citecolor=blue}

\usepackage{cleveref}
\usepackage[
    backend=biber,
    style=ieee,
    sorting=none
]{biblatex}
\usepackage{biblatex}
\definecolor{darkgreen}{rgb}{0.0, 0.5, 0.0}

\newcommand{\cmark}{\textcolor{green}{\ding{51}}}  
\newcommand{\xmark}{\textcolor{red}{\ding{55}}}    
\newcommand{\degree}{\textdegree}
\renewcommand{\arraystretch}{1.2}
\title{\LARGE \bf DynOPETs: A Versatile Benchmark for Dynamic Object Pose Estimation and Tracking in Moving Camera Scenarios}

\author{Xiangting Meng*$^{1}$, Jiaqi Yang*$^{1}$, Mingshu Chen$^{2}$, Chenxin Yan$^{1}$, \\
Yujiao Shi$^{1}$, Wenchao Ding$^{2}$, Laurent Kneip$^{1\dagger}$
\thanks{* Authors contributed equally to this work}
\thanks{$\dagger$ Corresponding authors
        {\tt lkneip@shanghaitech.edu.cn}
}
\thanks{$^{1}$ ShanghaiTech University, Mobile Peception Lab.}
\thanks{$^{2}$ Fudan Univesity, Multi-Agent Robotic Systems Lab.}
\thanks{Project page: \hyperref[https://stay332.github.io/DynOPETs]{\color{magenta}{https://stay332.github.io/DynOPETs}}}
}
\begin{document}

\maketitle
\thispagestyle{empty}
\pagestyle{empty}
\begin{refsection}
\begin{abstract}
In the realm of object pose estimation, scenarios involving both dynamic objects and moving cameras are prevalent. However, the scarcity of corresponding real-world datasets significantly hinders the development and evaluation of robust pose estimation models. This is largely attributed to the inherent challenges in accurately annotating object poses in dynamic scenes captured by moving cameras. To bridge this gap, this paper presents a novel dataset \textit{DynOPETs} and a dedicated data acquisition and annotation pipeline tailored for object pose estimation and tracking in such unconstrained environments. Our efficient annotation method innovatively integrates pose estimation and pose tracking techniques to generate pseudo-labels, which are subsequently refined through pose graph optimization. The resulting dataset offers accurate pose annotations for dynamic objects observed from moving cameras. To validate the effectiveness and value of our dataset, we perform comprehensive evaluations using 18 state-of-the-art methods, demonstrating its potential to accelerate research in this challenging domain. The dataset will be made publicly available to facilitate further exploration and advancement in the field.
\end{abstract}
\section{INTRODUCTION}
Object pose estimation, which involves determining an object's 6-DoF pose (position and orientation) in 3D space, is a fundamental research challenge in computer vision and robotics.
The task is crucial for various applications in embodied intelligence, augmented reality (AR), and mixed reality (MR) systems~\cite{wen2024foundationpose}, such as robotic manipulation~\cite{ubellacker2024high}, dexterous hand control, and human-object interaction~\cite{banerjee2025hot3d}.
Despite significant advancements in recent years driven by the emergence of benchmark datasets, existing real-world datasets exhibit notable limitations.
Most of the existing datasets predominantly assume static conditions - either fixed camera positions or stationary objects. This assumption contrasts sharply with real-world requirements where simultaneous camera ego-motion and object dynamics must be addressed.

The limited availability of dynamic pose estimation datasets stems primarily from annotation challenges inherent to moving-camera-dynamic-object configurations.
While static-object scenarios allow for indirect pose annotation via the state-of-the-art camera pose estimation technique, such as SLAM algorithm~\cite{teed2021droid} or fiducial marker-based localization method~\cite{olson2011apriltag}, dynamic settings require direct object motion tracking.
In such dynamic settings, existing solutions for pose annotation rely on either labor-intensive manual annotation~\cite{wen2020se} or expensive motion capture (Mocap) systems in controlled environments~\cite{drost2017introducing}. 
\begin{figure}[!t]
\centering
\includegraphics[width=0.48\textwidth]{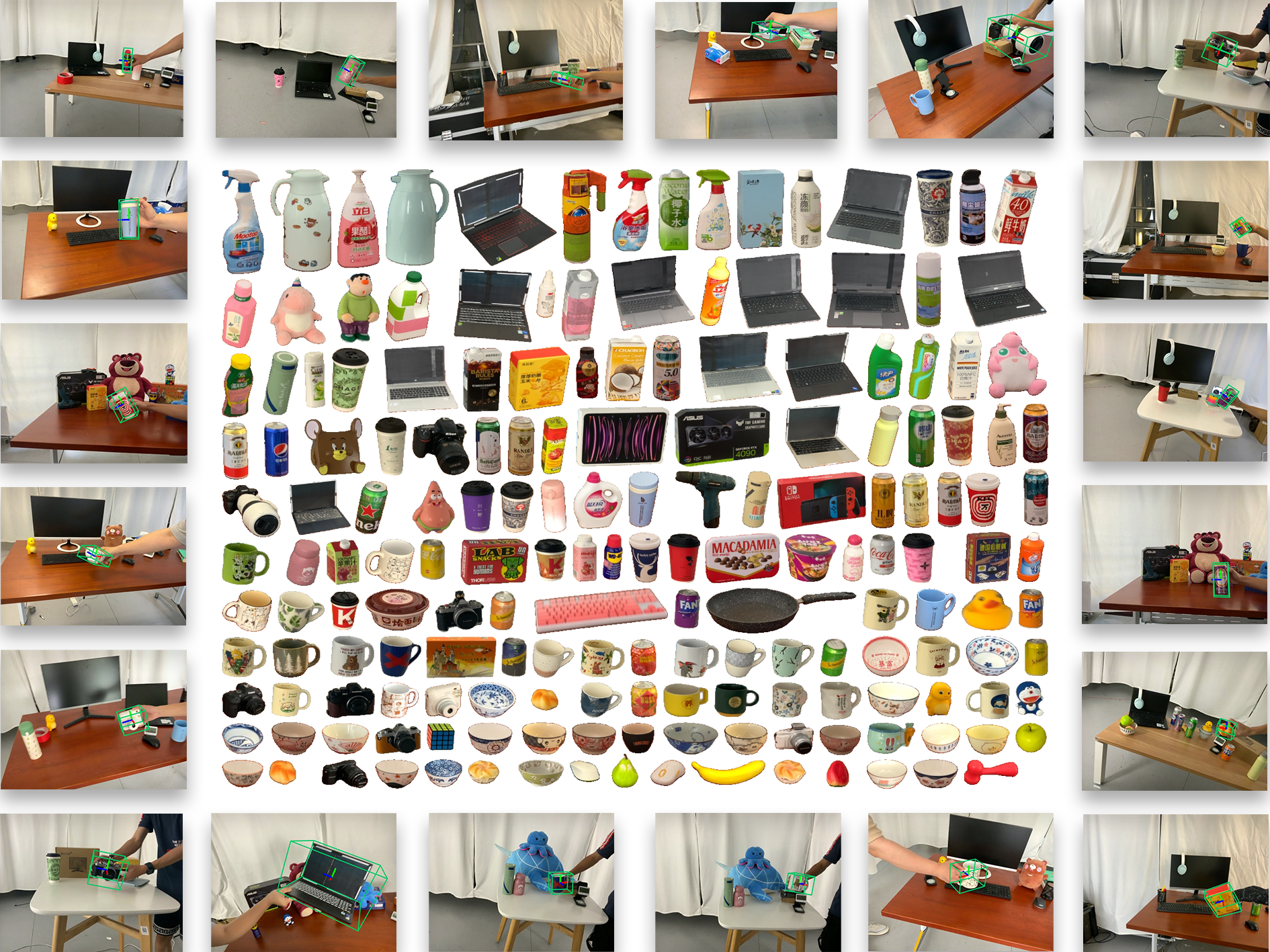}
\caption{The full set of dynamic objects in our dataset.}
\label{fig:overview}
\end{figure}
Another approach~\cite{guo2023handal} involves using state-of-the-art object pose estimation methods~\cite{wen2024foundationpose} to generate high-quality pseudo-labels for annotation. However, many common objects in daily life, such as those with symmetric geometries, transparent or reflective materials, or sparse/repetitive patterns, pose significant challenges for those algorithms. As a result, a significant fraction of the object pose annotations still require manual refinement due to the low quality of the pseudo-labels.

To address the issue of scarce real-world datasets in this challenging scenario, we propose a novel dataset that includes diverse object categories and dynamic scenes. The dataset primarily consists of RGB-D video sequences, object CAD models, and high-quality object and camera pose annotations, supporting various tasks in the field of object pose estimation. To tackle the challenges of pose annotation, we introduce a novel automated pose annotation pipeline leveraging the fusion of absolute and relative object pose estimates, significantly improving data labeling efficiency and reducing human effort. Finally, we systematically investigate existing state-of-the-art methods and comprehensively benchmark their performance in dynamic scenarios, demonstrating the value of our work. We believe that our proposed dataset and efficient pose annotation pipeline can fill the gap in dynamic object pose estimation datasets, thereby promoting the development of applications in the fields of embodied intelligence and AR/MR.
Our main contributions are summarized as follows:
\begin{itemize}
\item We propose an efficient pose annotation approach that integrates an absolute pose estimator with a global EKF smoother, a relative pose estimator based on point tracking, and an object pose graph optimization module. This pipeline enables accurate, marker-free dynamic object pose labeling, without extensive manual effort.
\item We present \textit{DynOPETs}, a comprehensive object pose estimation dataset comprising RGB-D sequences of 175 distinct object instances undergoing simultaneous camera and object motion, with synchronized 6-DoF pose annotations for both camera and object trajectories.
\item We conduct a systematic benchmark of various taxonomy-specific pose estimation and object pose tracking methods on our dataset.
\end{itemize}
\begin{table*}[t]
\caption{Comparison between existing 17 datasets with our proposed dataset. The table summarizes key attributes used for comparing datasets: object categories, whether the data is synthetic or real-world, data modalities (RGB, Depth), object details (number of instances, presence of dynamic objects, presence of moving cameras, CAD model availability), whether the videos are marker-free, and dataset size (number of videos, images, and annotations). }
\label{tab:dataset_comparison}
\centering
\small
\setlength{\tabcolsep}{5pt} 
\begin{tabular}{lcccccccccccccc}
\toprule
\toprule
\multirow{2}{*}[-0.5ex]{Dataset} &\multirow{2}{*}[-0.5ex]{Cat.} & \multirow{2}{*}[-0.5ex]{Real} & \multicolumn{2}{c}{Modality} & \multicolumn{4}{c}{Object Details} & \multirow{2}{*}[-0.5ex]{Marker-Free} & \multirow{2}{*}[-0.5ex]{Vid.} & \multirow{2}{*}[-0.5ex]{Img.} & \multirow{2}{*}[-0.5ex]{Anno.} \\
\cmidrule(lr){4-5} \cmidrule(lr){6-9} 
& &  & RGB & Depth  & Num. & CAD & Dyna-Obj & Mov-Cam &  &  &  &  \\
\midrule
CAMERA25~\cite{wang2019normalized}    & 6   & \xmark & \cmark & \cmark & 1085 & \cmark & \xmark & \cmark & \cmark & - & 300K & 4M \\
Omni6D~\cite{zhang2024omni6d}      & 166  & \xmark & \cmark & \cmark & 4688 & \cmark & \xmark & \cmark & \cmark & - & 0.8M & 5.6M\\
SOPE~\cite{zhang2024omni6dpose}        & 149  & \xmark & \cmark & \cmark & 4162 & \cmark & \xmark & \cmark & \cmark & - & 475K & 5M\\
\midrule
\midrule
YCB-Video~\cite{xiang2018posecnn}   & -   & \cmark & \cmark & \cmark & 21  & \cmark & \xmark & \cmark & \cmark & 92  & 133K & 0.6M \\
T-LESS~\cite{hodan2017t}       & -   & \cmark & \cmark & \cmark & 30  & \cmark & \xmark & \cmark & \xmark & 20  & 48K & 48K \\
Linemod~\cite{hinterstoisser2013model} & -   & \cmark & \cmark & \cmark & 15  & \cmark & \xmark & \xmark & \xmark & -  & 18K &15k  \\
StereoOBJ-1M~\cite{liu2021stereobj} & -   & \cmark & \xmark & \cmark & 18  & \cmark & \xmark & \cmark & \xmark & 182  & 393K & 1.5M\\
YCBInEOAT~\cite{wen2020se}        & -   & \cmark & \cmark & \cmark & 5   & \cmark & \cmark & \xmark & \cmark & 9   & 7449 & 3.7K \\
HOT3D~\cite{banerjee2025hot3d}           & -   & \cmark & \cmark & \xmark & 33  & \cmark & \cmark & \cmark & \xmark & 425  & 3.7M & 6M\\
\midrule
\textbf{UOPE-56 (Ours)}   & -   & \cmark & \cmark & \cmark & 56  & \cmark & \cmark & \cmark & \cmark & 56  & 64K  & 180K \\
\midrule
REAL275~\cite{wang2019normalized}     & 6   & \cmark & \cmark & \cmark & 42  & \cmark & \xmark & \cmark & \xmark & 18  & 8K & 35K  \\
Wild6D~\cite{fu2022category}       & 5   & \cmark & \cmark & \cmark & 162 & \xmark & \xmark & \cmark & \cmark & 486  & 10K & 10K \\
PhoCaL~\cite{wang2022phocal}       & 8   & \cmark & \cmark & \cmark & 60  & \cmark & \xmark & \cmark & \xmark & 24  & 3.9K  & 91K\\
HANDAL~\cite{guo2023handal}       & 17  & \cmark & \cmark & \xmark & 212 & \cmark & \cmark & \xmark & \cmark & 2.2K & 308K & 1.2M\\
HouseCat6D~\cite{jung2024housecat6d} & 10  & \cmark & \cmark & \cmark & 194 & \cmark & \xmark & \cmark & \cmark & 41  & 23.5K & 160K \\
ROPE~\cite{zhang2024omni6dpose}             & 149 & \cmark & \cmark & \cmark & 581 & \cmark & \xmark & \cmark & \cmark & 363  & 332K &1.5M \\
\midrule
\textbf{COPE-119 (Ours)}  & 6   & \cmark & \cmark & \cmark & 119 & \cmark & \cmark & \cmark & \cmark & 119  & 140K & 350K\\
\bottomrule
\bottomrule
\end{tabular}

\end{table*}

\section{RELATED WORK}
\label{sec:related_work}
According to ~\cite{liu2024deep}, We categorize object pose estimation methods into 4 classes: instance-level object pose estimation (IOPE), unseen object pose estimation (UOPE), category-level object pose estimation (COPE), and object pose tracking (OPT). This section discusses representative and state-of-the-art methods within each class, followed by an overview of recently published datasets relevant to these areas.

\subsection{Object Pose Estimation and Tracking Methods}
\subsubsection{Instance-level Object Pose Estimation}
\label{subsubsec:iope} 
These methods predict the pose of only those objects present in the training set, often relying on priors such as CAD models or posed reference views during inference~\cite{wang2019densefusion}. Object pose estimation can be achieved through template matching~\cite{dang2022learning} or feature correspondence matching~\cite{xu2022rnnpose} given these priors. Some approaches directly predict object pose using neural networks~\cite{wang2019densefusion}, while others employ voting schemes~\cite{zhou2023deep}. Instance-level methods are the most mature but also the most restrictive due to their inability to generalize to novel objects. Consequently, research has shifted towards the more generalizable UOPE and COPE methods, which are the primary focus of this paper and our proposed dataset.

\subsubsection{Unseen Object Pose Estimation}
Similar to IOPE methods, UOPE methods rely on CAD models or posed reference views as priors. However, UOPE methods can generalize to novel objects, enabling pose prediction for unseen objects during inference. Recent works in this area are broadly categorized into two main strategies: feature matching and template matching-based methods. 
Feature matching methods~\cite{chen2023zeropose, lin2024sam,caraffa2024freeze,huang2024matchu} extract robust image features and establish correspondences with generic, object-agnostic CAD model features. These approaches enable zero-shot estimation but may rely heavily on feature discriminability.
In contrast, template matching methods~\cite{labbe2022megapose,nguyen2024gigapose,ornek2024foundpose,wen2024foundationpose} leverage extensive sets of object templates to achieve strong generalization across novel objects.

\subsubsection{Category-level Object Pose Estimation}
COPE methods typically handle novel objects within the same category by learning category-specific shape representations, eliminating the reliance on CAD models and reference views. Wang et al.~\cite{wang2019normalized} pioneered this approach by introducing the normalized object coordinate space (NOCS), a canonical representation where all objects within the same category are consistently aligned in terms of size and orientation. Aligning predicted canonical representations with depth points allows for the estimation of object pose.

Recent advancements have explored different network designs, such as using vision transformers~\cite{krishnan2024omninocs} or diffusion models~\cite{ikeda2024diffusionnocs} to predict better canonical representations. Several works~\cite{wang2023query6dof,meng2023kgnet,liu2023net,ikeda2024diffusionnocs,wang2024gs} employ deformed categorical shape templates to improve canonical representations, while others~\cite{li2023generative, zhang2024omni6dpose} employ diffusion models to formulate COPE as a conditional generative task. Furthermore, some methods directly predict object pose, employing either enhanced geometric feature extractors~\cite{li2025gce} or a combination of geometric and semantic priors~\cite{lin2023vi, chen2024secondpose, lin2024instance}.
\subsubsection{Object Pose Tracking}
Unlike single-frame object pose estimation methods, object pose tracking leverage temporal information across video sequences to estimate object poses~\cite{liu2024deep}. Typically, tracking methods also require posed reference views or CAD models. While classical methods~\cite{wen2021bundletrack} often track object poses using geometric or feature-based approaches, recent methods~\cite{wen2023bundlesdf} jointly track and refine object shape through neural implicit representations, providing robust performance. Moreover, Wen et al.~\cite{wen2024foundationpose} leverage large-scale pre-trained models and efficiently track unseen objects using large-scale synthetic training data for enhanced generalization.

\subsection{Object Pose Estimation Datasets}
Object pose estimation datasets have evolved significantly over the years. We show a dataset comparison in table~\ref{tab:dataset_comparison}. The following is a brief overview.

Linemod~\cite{hinterstoisser2013model} introduced a standard instance-level RGB-D benchmark featuring texture-less objects with limited occlusions in relatively simple scenarios. YCB-Video~\cite{xiang2018posecnn} subsequently expanded realism with household objects. 
T-Less~\cite{hodan2017t} introduced texture-less but symmetric objects in challenging industrial scenarios, highlighting pose ambiguity issues.
REAL275~\cite{wang2019normalized} established the first real-world benchmark for category-level pose estimation by introducing canonical object representations across 6 categories. 
StereOBJ-1M~\cite{liu2021stereobj} further tackled challenging materials with a much larger scale, such as transparent and reflective objects.
To further diversify modalities, 
PhoCaL~\cite{wang2022phocal} provided multimodal images focusing on challenging transparent and reflective household items with 8 categories.
Wild6D~\cite{fu2022category} advanced realism further by capturing large-scale videos of objects in the wild. More recently,
HANDAL~\cite{guo2023handal} focused on graspable objects, collecting 17 categories of manipulable items in cluttered scenes via multi-view imagery and annotating poses with~\cite{wen2023bundlesdf}. 
HOT3D~\cite{banerjee2025hot3d} was the first to address egocentric scenarios featuring dense hand–object interactions. However, it incorporates a limited variety of objects and constrained scene settings.
HouseCat6D~\cite{jung2024housecat6d} delivered highly accurate annotations for 10 categories of household objects using polarization imagery to enhance robustness to reflections. OmniNOCS~\cite{krishnan2024omninocs} \textcolor{blue}{is} a unified NOCS dataset that integrated data from multiple domains and featured over 90 object classes, making it the largest NOCS dataset to date. Omni6D~\cite{zhang2024omni6d} and ROPE~\cite{zhang2024omni6dpose} also provided a large-scale dataset for object pose estimation.
\begin{figure*}[ht]
  \centering
  \includegraphics[{width=\textwidth}]{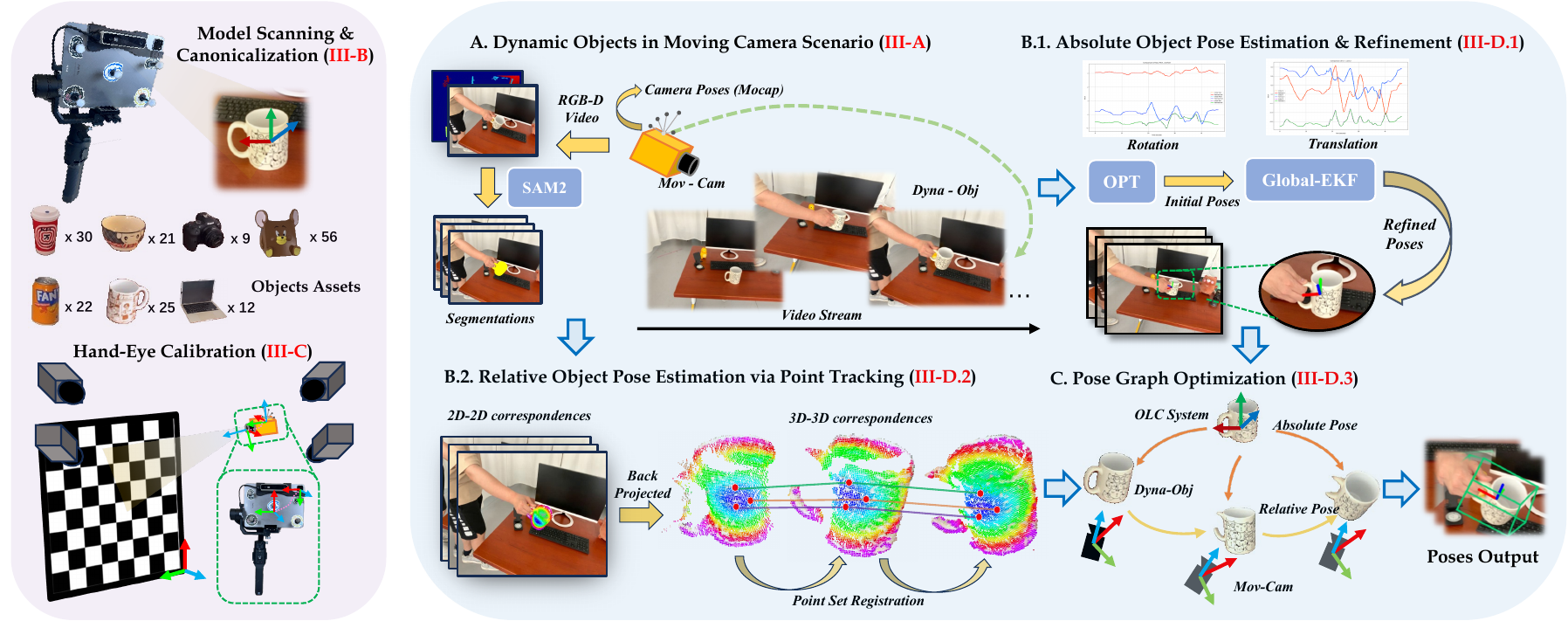}
  \caption{Overview. The left block is CAD model scanning and preprocessing, as well as hand-eye calibration. The right block is the data acquisition and efficient object pose annotation pipeline. The dynamic object is captured by a moving RGB-D camera, and the ground truth object pose annotation is calculated by fusing refined absolute object pose estimates and relative object pose estimates. The camera pose is provided by a Mocap.}
  \label{fig:pipeline}
\end{figure*}
\section{Data Acquisition}
\subsection{Overview}
The proposed \textit{DynOPETs} consists of 175 RGB-D video sequences of distinct everyday objects captured on typical desktop backgrounds with sufficient visual features. Each sequence comprises roughly 500 frames recorded at 15 Hz, totaling approximately 210K RGB-D frames. Additionally, we provide over 530K annotations, included instance segmentations, camera intrinsic and extrinsic parameters, object poses, and camera poses. The dataset is divided into two subsets. The first, COPE-119, comprises 119 sequences of objects belonging to 6 categories commonly found in the COPE benchmark: bottles, bowls, cameras, cans, laptops, and mugs. 
The second, UOPE-56, contains 56 sequences of other common household objects (illustrated in ~\cref{fig:overview}) and is designed for evaluating the performance of UOPE algorithms. 

\subsection{Hardware Setup and Objects}
\label{hardware&objects}
RGB-D data was captured using a consumer-grade iPad Pro with the hardware-synchronized Structure Sensor Pro~\cite{structuresensor} depth camera. CAD models for all 175 objects were scanned using the sensors' integrated RGB-D 3D modelling software. Following mesh reconstruction, objects were manually aligned so that their centers coincided with the origin of the 3D coordinate system, and their front, top, and side views aligned with the x, y, and z axes, respectively. Consistent with the definition in Wang et al.~\cite{wang2019normalized}, the coordinate system of an object's CAD model is defined as Object Local Coordinates (OLC) system, also known as the canonical space in the COPE domain, as illustrated in Figure~\ref{fig:pose_graph}. Additionally, we employed SAM2~\cite{ravi2024sam} to automatically segment objects in these frames.

\subsection{Sensor Calibration and Synchronization}
Intrinsic and extrinsic camera parameters were calibrated using the depth camera SDK, which provides simultaneous calibration of both RGB and depth cameras. We developed a custom camera driver based on the SDK that warps the depth map to the RGB camera's coordinate frame, resulting in co-registered RGB-D video sequences. A Mocap system provided camera poses. The relative transformation between the Mocap marker frame and the camera's optical center was calibrated using a hand-eye calibration procedure, yielding camera poses within the Mocap coordinate system. A temporal offset existed between the Mocap system's timestamps and the RGB-D frame timestamps. To resolve this, we used visual odometry (VO)~\cite{teed2021droid} to estimate the camera pose for each frame and registered these VO-derived poses to the Mocap poses. This registration process established a mapping between each camera frame and the corresponding Mocap timestamp.

\subsection{Efficient Object Pose Annotation}
Illustrated in \cref{fig:pipeline},
To provide context for our approach, we begin with key definitions. Firstly, the absolute object pose, which we aim to annotate, is defined as the Sim(3) transformation (scale, rotation, and translation) required to align an object from its OLC system to the camera frame. Since our CAD models and corresponding point clouds are scaled to real-world dimensions, this Sim(3) simplifies to an SE(3) transformation (rotation and translation). Secondly, the relative object pose describes the SE(3) transformation between the object's poses at two different time steps in the camera coordinate system. These two definitions are directly relevant to the following two approaches for obtaining the object pose annotations.

\subsubsection{Absolute Object Pose Estimation and Refinement}
The first approach involves directly generating high-quality pseudo pose annotations using existing state-of-the-art algorithms. Model-based method FoundationPose~\cite{wen2024foundationpose} stands out in this category, leveraging CAD models to output object motion. 
Although generally accurate, FoundationPose~\cite{wen2024foundationpose} can occasionally generate discontinuous pose estimates. To address this issue, we adopt a global EKF~\cite{bishop2001introduction} with Rauch-Tung-Striebel (RTS) smoother~\cite{rauch1965maximum} to refine the object pose predictions, mitigating pose jitters. The RTS smoother employs a constant velocity model for both translation $p\in \mathbf{R}^3$ and rotational quaternion $q\in \mathbb{S}^3$, where the state variables are defined as $\mathbf{x}=[p, q, \upsilon, \omega]$, where $\upsilon\in\mathbb{R}^3, \omega\in\mathbb{R}^3$ refers the velocity and angular velocity. We first perform a forward pass~\cref{eq: ekf} using the standard EKF prediction and update equations to obtain filtered state estimates $\mathbf{x}_{t}$ from step $t-1$ to step $t$,
then, a backward pass~\cref{eq: rts smoother} using the RTS smoother refines these estimates from step $t+1$ to $t$:
\begin{equation}
\label{eq: state transfer}
\hat{\mathbf{x}}_{t} = \mathbf F\mathbf{x}_{t-1}, 
\end{equation}
\begin{equation}
\label{eq: ekf}
\mathbf{x}_{t} = \hat{\mathbf{x}}_{t} + \mathbf K_t\, (\mathbf{z}_t - \mathbf H\hat{\mathbf{x}}_{t}),
\end{equation}
\begin{equation}
\label{eq: rts smoother}
\mathbf{x}_{t} = \mathbf{x}_{t} + \mathbf G_t(\mathbf{x}_{t+1} - \hat{\mathbf{x}}_{t+1}),
\end{equation}
\begin{equation}
\label{eq: rts gain}
\mathbf{G}_t = \mathbf{P}_{t}\mathbf{F}^T \mathbf{P}_{t+1}^{-1},
\end{equation}
where $\mathbf{z}$ refers to the observed pose, and $\mathbf F$ and $\mathbf H$ represent the state transition and observation matrices, respectively. $K$ is the extended Kalman gain used in the update step~\cite{bishop2001introduction}, $\mathbf P$ denotes the covariance matrix, and $\mathbf G$ represents the smoothing gain.
After our refinement, the pose annotations of objects with rich textures and distinctive shapes are now accurate and smooth, while for those objects exhibiting significant symmetry or lacking distinctive texture, such as objects within the bowl category, the pose annotation is still not high-quality enough.
\subsubsection{Relative Object Pose Estimation via Point Tracking}
To address the challenges posed by these objects, we explore a second approach that involves first estimating the absolute object pose in the initial frame and then iteratively computing the frame-to-frame relative object pose to derive the pose annotation for the entire sequence. 

BundleTrack~\cite{wen2021bundletrack} serves as a representative example of this approach, employing keypoint detection and matching to establish 2D correspondences. These 2D correspondences are then back-projected into 3D space using sensor depths, forming 3D-3D correspondences that enable the computation of relative object poses via point registration techniques. Our method adopts a similar strategy but replaces the keypoint detection and matching component with a state-of-the-art Track Any Point (TAP) model~\cite{karaev2024cotracker3}. 
TAP models accept a video sequence as input and output 2D correspondence points for a group of specified query points in the reference frame across all frames. Compared to traditional feature extraction and matching methods, TAP model~\cite{karaev2024cotracker3} is better equipped to leverage the temporal information inherent in video sequences, enabling them to effectively handle objects with limited texture.  
While this method produces smoother object pose estimates and is more robust to symmetry ambiguities, it is susceptible to cumulative errors, leading to drift in the estimated trajectory over time.
\subsubsection{Object Pose Graph Optimization}
\begin{figure}[thbp]
    \centering
    \includegraphics[width=0.38\textwidth]{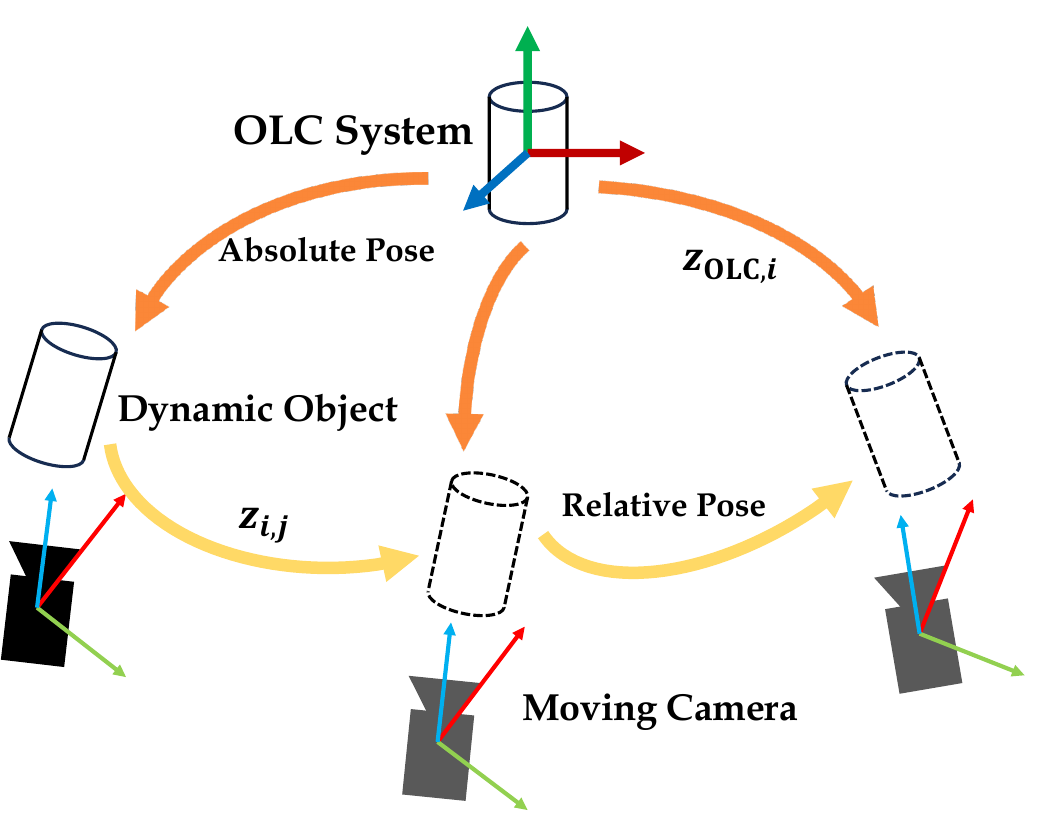}
    \caption{Pose graph Optimization. The orange arrows represent absolute pose edges $(\text{OLC}, i)$ and the yellow arrows represent relative pose edges $(i,j)$. In the OLC space, the object's center aligns with the origin and has a semantically meaningful orientation. For example, the opening of a cup or bottle aligns with the y-axis, while the front and sides of the object are aligned with the x-axis and z-axis, respectively. }
    \label{fig:pose_graph}
\end{figure}

Recognizing the complementary nature of the two annotation methods discussed before, we construct a pose graph $\mathcal{G} = (\mathcal{V}, \mathcal{E})$ to integrate them into a unified framework.  The graph consists of nodes $\mathcal{V}$ and edges $\mathcal{E}$(illustrated in~\cref{fig:pose_graph}). Each node $i \in \mathcal{V}$ represents the pose of an object at a specific time step, denoted as $\mathbf{T}_i \in SE(3)$, where $\mathbf{T}_i$ represents the transformation from the object's local coordinate system to the camera coordinate system at time $i$. In essence, objects in OLC and their corresponding camera frames at each time step become the nodes in this graph.

The edges $\mathcal{E}$ represent constraints between these object poses. We utilize two types of edges: absolute pose edges and relative pose edges.  Absolute pose edges, denoted as $(\text{OLC}, i) \in \mathcal{E}$, are derived from FoundationPose refined by the EKF/RTS smoother.  Relative pose edges, denoted as $(i, j) \in \mathcal{E}$, connect two nodes representing the same object at different time steps. These are derived from the point tracking method (TAP). We use $\mathbf{z}_{\text{OLC}, i}, \mathbf{z}_{i, j} \in SE(3)$ to represent the associated measurements of these edges, and $\mathbf{\Omega}_{i, j}$ and $\mathbf{\Omega}_{\text{OLC}, i}$ for the corresponding information matrices. Note that in our implementation, $\mathbf{\Omega}_{\text{OLC}, i}$ is constant and $\mathbf{\Omega}_{i, j}$ are calculated from the Jacobian of the point set registration. Also, absolute pose edges with poor measurements can be easily removed manually.

By encoding both absolute and relative poses as edges connecting these nodes, we form a graph with closed loops. Then we need to finding the set of object poses $\mathbf{T} = \{\mathbf{T}_i\}_{i \in \mathcal{V}}$ that minimizes the following cost function:
\begin{align}
\label{eq:pose_graph_cost}
\mathcal{F}(\mathbf{T}) & = \sum_{(\text{OLC}, i) \in \mathcal{E}} \mathbf{r}_{\text{OLC}, i}^T(\mathbf{T}) \mathbf{\Omega}_{\text{OLC}, i} \mathbf{r}_{\text{OLC}, i}(\mathbf{T}) \\
& + \sum_{(i, j) \in \mathcal{E}} \mathbf{r}_{i, j}^T(\mathbf{T}) \mathbf{\Omega}_{i, j} \mathbf{r}_{i, j}(\mathbf{T}),
\end{align}
where $\mathbf{r}_{\text{OLC}, i}(\mathbf{T})$ and $\mathbf{r}_{i, j}(\mathbf{T})$ are the residual error terms for absolute and relative pose edges, respectively, defined as:
\begin{equation}
\label{eq:absolute_residual}
\mathbf{r}_{\text{OLC}, i}(\mathbf{T}) = \log(\mathbf{z}_{\text{OLC}, i}^{-1} \mathbf{T}_i),
\end{equation}
\begin{equation}
\label{eq:relative_residual}
\mathbf{r}_{i, j}(\mathbf{T}) = \log(\mathbf{z}_{i, j}^{-1} \mathbf{T}_i^{-1} \mathbf{T}_j),
\end{equation}
Here, $\log(\cdot)$ denotes the matrix logarithm, mapping an $SE(3)$ transformation to its Lie algebra representation.

The optimization problem, therefore, is to find:
\begin{equation}
\label{eq:optimization_problem}
\mathbf{T}^* = \arg\min_{\mathbf{T}} \mathcal{F}(\mathbf{T}).
\end{equation}

This non-linear least squares problem can be solved using Gauss-Newton or Levenberg-Marquardt algorithms. By minimizing the cost function, we obtain refined object poses that are globally consistent, leveraging the strengths of both absolute and relative pose estimation methods. The manual effort required by our method is primarily limited to identifying and removing inaccurate absolute pose estimates, which minimizes human intervention and promotes annotation efficiency.

\renewcommand{\arraystretch}{1.3}
\begin{table*}[htbp]
    \centering
    \caption{Comparison of COPE methods on the COPE-119 dataset. Methods are categorized into those that predict NOCS maps and those that directly predict object poses. \textbf{*} indicates that only the NOCS map prediction component was used. \textbf{\dag} denotes that the method does not predict scale. }
    \label{cope_results}
    \begin{tabular*}{0.95\textwidth}{l@{\extracolsep{\fill}}cccccccccc}
    \hline
    \hline
    \textbf{Method} & \textbf{IoU25 $\uparrow$} & \textbf{IoU50 $\uparrow$} & \textbf{IoU75 $\uparrow$} & \textbf{5\degree{}2cm $\uparrow$} & \textbf{5\degree{}5cm $\uparrow$} & \textbf{10\degree{}2cm $\uparrow$} & \textbf{10\degree{}5cm $\uparrow$} \\
    \hline
    NOCS~\cite{wang2019normalized} & 73.50 & 48.75 & 20.47 & 11.57 & 16.71 & 22.43 & 32.98 \\
    NOCS (MV-ROPE)\textbf{*}~\cite{yang2024mv} & 88.74 & 71.67 & 38.19 &  28.84  & 31.94 & 48.50 & 55.53 \\
    DiffusionNOCS~\cite{ikeda2024diffusionnocs} & 83.00 & 38.89 & 4.57 & 39.87 & 43.83 & 55.82 & 64.80 \\
    \hline
    GCASP~\cite{li2023generative} & 89.13 & 52.84 & 14.79 & 50.22 & 52.74 & 65.89 & 70.97 \\
    VI-Net~\cite{lin2023vi} & 95.66 & 87.34 & 52.35 & 55.09 & 59.98 & 72.35 & 79.70 \\
    IST-Net~\cite{liu2023net} & 97.52 & 90.68 & 58.99 & 51.57 & 52.85 & 70.90 & 73.41 \\
    GenPose\textbf{\dag}~\cite{zhang2023generative} & - & - & - & 47.48 & 54.30 & 60.96 & 71.84 \\
    SecondPose~\cite{chen2024secondpose} & \textbf{98.55} & 90.90 & 53.83 & 58.57 & \textbf{64.00} & 75.13 & \textbf{83.29} \\
    AGPose~\cite{lin2024instance} & 98.47 & \textbf{92.94} & \textbf{66.34} & \textbf{61.61} & 63.24 & \textbf{80.02} & 82.06 \\     
    \hline
    \hline
    \end{tabular*}
\end{table*}
\renewcommand{\arraystretch}{1.3}
\begin{table*}[htbp]
    \centering
    \caption{Comparison of Fine-tuned VS. Original COPE methods on COPE-119 test split, \textbf{*} indicates fine-tuned methods.}
    \label{cope_finetuned_results}
    \begin{tabular*}{0.95\textwidth}{l@{\extracolsep{\fill}}cccccccccc}
    \hline
    \hline
    \textbf{Method} & \textbf{IoU25 $\uparrow$} & \textbf{IoU50 $\uparrow$} & \textbf{IoU75 $\uparrow$} & \textbf{5\degree{}2cm $\uparrow$} & \textbf{5\degree{}5cm $\uparrow$} & \textbf{10\degree{}2cm $\uparrow$} & \textbf{10\degree{}5cm $\uparrow$} \\
    \hline
    SecondPose~\cite{chen2024secondpose} & 98.53 & 92.24 & 57.42 & 61.88 & 67.05 & 78.24 & 84.08  \\
    AGPose~\cite{lin2024instance}  & 98.52 & 92.10 & 68.53 & 67.11 & 67.50 & 81.23 & 81.92 \\
    \hline
    SecondPose\textbf{*}~\cite{chen2024secondpose} & 99.38(\textcolor[rgb]{0.0,0.5,0.0}{+0.85}) & 96.23(\textcolor[rgb]{0.0,0.5,0.0}{+3.99}) & 59.80(\textcolor[rgb]{0.0,0.5,0.0}{+2.38}) & \textbf{76.33}(\textcolor[rgb]{0.0,0.5,0.0}{+14.45}) & \textbf{82.91}(\textcolor[rgb]{0.0,0.5,0.0}{+15.86}) & 89.43(\textcolor[rgb]{0.0,0.5,0.0}{+11.19}) & \textbf{97.19}(\textcolor[rgb]{0.0,0.5,0.0}{+13.11}) \\
    AGPose\textbf{*}~\cite{lin2024instance} & \textbf{99.91}(\textcolor[rgb]{0.0,0.5,0.0}{+1.39}) & \textbf{99.76}(\textcolor[rgb]{0.0,0.5,0.0}{+7.66}) & \textbf{84.42}(\textcolor[rgb]{0.0,0.5,0.0}{+15.89}) & 75.10(\textcolor[rgb]{0.0,0.5,0.0}{+8.99}) & 75.74(\textcolor[rgb]{0.0,0.5,0.0}{+8.24}) & \textbf{94.35}(\textcolor[rgb]{0.0,0.5,0.0}{+13.12}) & 96.46(\textcolor[rgb]{0.0,0.5,0.0}{+14.54}) \\
    \hline
    \hline
    \end{tabular*}
\end{table*}
\section{EVALUATION}
\subsection{Comparsions with Pose from Motion Capture System}
Before conducting benchmarks, we need to verify the accuracy of the pose labels obtained from the proposed annotation algorithm. For this purpose, we collected an additional 10 sequences of data, where the objects in these sequences were equipped with motion capture system markers, allowing us to obtain ground truth with millimeter-level accuracy. We use the Absolute Trajectory Error (ATE) and Relative Pose Error (RPE) \cite{sturm2012benchmark} to evaluate the pose annotation compared to the ground truth object pose. The results show that the average rotation error between our poses and those obtained from the motion capture system is less than one degree, and the average translation error is less than 1 cm, which is sufficiently accurate. Due to space limitations, please refer to the PDF supplementary material on the project page for details of this experiment.
\subsection{Overview}
As discussed in ~\cref{subsubsec:iope}, IOPE methods are not our primary focus, instead, we adopt the COPE and UOPE methods for benchmarking, which are capable of generalizing across intra-class variations and unseen objects without the need for individual retraining for each object. These methods can directly leverage open-source pre-trained weights. This approach serves two purposes: first, it allows us to verify the validity and accuracy of our data; second, it helps evaluate the generalization capability of the open-source algorithm on our dataset.
Furthermore, we fine-tuned several variants of the COPE method on the COPE-119 training set and evaluated their performance on the test set to validate the effectiveness of our ground truth labels for network training. 
Additionally, we assessed various object-pose-tracking methods to further broaden our analysis. 
Finally, it is important to note that due to space limitations, detailed experimental analysis and in-depth discussion can be found in the PDF supplementary material on the project page.

\subsection{Evaluation Metrics}
We follow \cite{wang2019normalized} and adopt the evaluation protocol commonly used for COPE methods, which includes the mean precision of 3D intersection over union (IoU) to jointly assess rotation, translation, and size prediction.
3D IoU measures the overlap between the 3D bounding box of an object derived from the predicted pose and the ground truth bounding box, providing a comprehensive metric. For rotation and translation errors, we typically report the accuracy within a specified threshold. For instance, Rot$_{m}$Trans$_{n}$ represents the percentage of predicted poses with a rotation error less than $m$ degrees and a translation error less than $n$ centimeters. 

We evaluated the Average Recall (AR)~\cite{hodan2023bop} for UOPE methods using three error functions: Visible Surface Discrepancy (VSD), Maximum Symmetry-Aware Surface Distance (MSSD), and Maximum Symmetry-Aware Projection Distance (MSPD). In addition, we report the frame rate (FPS) for all UOPE methods. 

The evaluation of OPT methods performance relies on two complementary metrics: We evaluate object pose tracking performance using ATE for global accuracy and  RPE~\cite{sturm2012benchmark} for local accuracy, reporting average results across all 175 sequences. We also report the AUC of ADD and ADD-S~\cite{xiang2018posecnn}, as well as the average recall of ADD(S) within 10\% of the object diameter (termed ADD(S)-0.1d)~\cite{he2022oneposeplusplus}.

\subsection{Category-level Pose Estimation}
\subsubsection{COPE-119 Baseline Evaluation}
As shown in \cref{cope_results}, for COPE task, we evaluate 8 methods: ~\cite{wang2019normalized}, ~\cite{yang2024mv}, and ~\cite{ikeda2024diffusionnocs} predict canonical representations directly, while~\cite{li2023generative} ~\cite{lin2023vi}, ~\cite{liu2023net}, \textcolor{blue}{~\cite{zhang2023generative}}, ~\cite{chen2024secondpose}, and ~\cite{lin2024instance} predict object pose and category. Except for NOCS~\cite{wang2019normalized}, which performs joint object detection and pose estimation, all methods utilize identical SAM2~\cite{ravi2024sam} segmentation results to ensure a fair benchmark. 
Notably, DiffusionNOCS~\cite{ikeda2024diffusionnocs} achieves a relatively lower IoU score but exhibits a very small pose error. A possible reason is that it has never been trained on real-world data, leading to instability in object scale prediction.
\subsubsection{COPE-119 Fine-Tuning Evaluation}
The purpose of this experiment is to demonstrate the accuracy of pose annotations, based on the assumption that the metric improves only when the ground truth pose annotations in both the training and test sets are correct. We fine-tuned AGPose~\cite{lin2024instance} and SecondPose~\cite{chen2024secondpose} on our 84-sequence training set for 2 epochs and evaluated them on our 35-sequence test set. As shown in ~\cref{cope_finetuned_results}, the fine-tuned results significantly outperform their original test set performance by a large margin in merely 2 epochs of fine-tuning, highlighting the accuracy of pose annotations in our dataset.

\renewcommand{\arraystretch}{1.2}
\begin{table}[htbp]
    \centering
    \caption{Comparison of UOPE methods on the UOPE-56 datasets. \textbf{*} indicates pose estimation without refinement.}
    \label{UOPE_methods_comparison}
    \resizebox{0.48\textwidth}{!}{ 
    \begin{tabular}{lcccccc}
    \hline
    \hline
    \textbf{Method} & \textbf{Input} & \textbf{VSD} $\uparrow$ & \textbf{MSSD} $\uparrow$ &  \textbf{MSPD} $\uparrow$ & \textbf{AR} $\uparrow$& \textbf{FPS} $\downarrow$ \\
    \hline
    FoundPose\textbf{*}~\cite{ornek2024foundpose} & RGB & 50.12 & 52.02 & 79.25 & 60.46 & 0.50 \\
    MegaPose~\cite{labbe2022megapose} & RGB & 64.56 & 68.45 & \textbf{83.64} & \textbf{72.23} & 2.25 \\
    GigaPose~\cite{nguyen2024gigapose} & RGB & \textbf{64.84} & 61.32 & 83.20 & 69.79 & 3.37 \\
    \hline
    MegaPose~\cite{labbe2022megapose} & RGB-D & 76.34 & 80.48 & 84.64 & 80.49 & 2.61 \\
    SAM6D~\cite{lin2024sam} & RGB-D & 79.67 & 77.95 & 81.65 & 79.75 & 3.75 \\
    FoundationPose~\cite{wen2024foundationpose} & RGB-D & \textbf{92.74} & \textbf{90.09} & \textbf{94.40} & \textbf{92.41} & 0.98 \\
    \hline
    \hline
    
    \end{tabular}
    }
\end{table}

\subsection{Unseen Object Pose Estimation}
We evaluate 5 UOPE methods in \cref{UOPE_methods_comparison}, grouped by their input modalities into RGB-based and RGB-D-based approaches. To ensure fairness, each sequence is paired with a specific CAD model and uniform segmentations produced by SAM2~\cite{ravi2024sam}. Moreover, methods that rely on CNOS~\cite{nguyen2023cnos} for detection or segmentation also utilize SAM2. Except for SAM6D~\cite{lin2024sam}, which experiences performance bottlenecks due to its intrinsic segmentation component and fixed threshold settings that fail to generalize in our dataset’s cluttered backgrounds, we apply dynamic threshold adjustments to maintain accurate pose estimation. MegaPose~\cite{labbe2022megapose}, available in RGB-only and RGB-D configurations, leverages extensive pre-processing to extract rich template information, demonstrating a significant advantage, especially on single-object sequences with ground-truth detection. Notably, FoundPose~\cite{ornek2024foundpose} does not open-source its core refinement code, limiting its evaluation of the results, so we only report its coarse results.

\subsection{Object Pose Tracking}
For object pose tracking methods, we benchmark ~\cite{runz2018maskfusion}, ~\cite{wen2021bundletrack}, ~\cite{teed2021droid}, ~\cite{wen2023bundlesdf}, and ~\cite{ornek2024foundpose}. 
We mainly employed the original algorithms, but with DROID-SLAM~\cite{teed2021droid}, we integrated SAM2-derived masks~\cite{ravi2024sam} to zero out weights of non-object regions directly within the DBA layer to derive the camera-to-object poses, effectively adapting it for object pose tracking. We report FoundationPose~\cite{wen2024foundationpose} results under the model-based mode. 
\renewcommand{\arraystretch}{1.2}
\begin{table}[htb]
    \centering
    \caption{Comparison on the DynOPETs dataset. The evaluation of average ATE (m), RPE on rotation (Rad), and average RPE on translation (m).}
    \label{OPT_trajectory_metrics}
    \resizebox{0.48\textwidth}{!}{ 
    \begin{tabular}{l >{\centering\arraybackslash}p{1.7cm} >{\centering\arraybackslash}p{1.7cm} >{\centering\arraybackslash}p{1.7cm}}
    \hline
    \hline
    \textbf{Method} & \textbf{ATE $\downarrow$} & \textbf{RPE Rot. $\downarrow$} & \textbf{RPE Trans. $\downarrow$} \\
    \hline
    MaskFusion~\cite{runz2018maskfusion} & 0.320 & 0.060 & 0.033 \\
    DROID-SLAM~\cite{teed2021droid} & 0.173 & 0.063 & 0.040 \\
    BundleTrack~\cite{wen2021bundletrack} & 0.068 & 0.037 & 0.021 \\
    BundleSDF~\cite{wen2023bundlesdf} & 0.094 & 0.044 & 0.026 \\
    FoundationPose~\cite{wen2024foundationpose} & \textbf{0.037} & \textbf{0.018} & \textbf{0.010} \\
    \hline
    \hline
    \end{tabular}
    }
\end{table}
\begin{table}[htb]
    \centering
    \caption{Comparison on DynOPETs dataset. ADD(S) represents AUC (0, 0.1m), and ADD(S)-0.1d is the average recall within 10\% of the object diameter.}
    \label{OPT_AUC_metrics}
    \resizebox{0.48\textwidth}{!}{ 
    \begin{tabular}{l >{\centering\arraybackslash}p{1.7cm} >{\centering\arraybackslash}p{1.7cm} >{\centering\arraybackslash}p{1.7cm} >{\centering\arraybackslash}p{1.7cm}}
    \hline
    \hline
    \textbf{Method} & \textbf{ADD $\uparrow$} & \textbf{ADDS $\uparrow$} & \textbf{ADD-0.1d $\uparrow$} & \textbf{ADDS-0.1d $\uparrow$} \\
    \hline
    MaskFusion~\cite{runz2018maskfusion} & 12.23 & 34.47 & 28.75 & 64.65 \\
    DROID-SLAM~\cite{teed2021droid} & 34.27 & 57.20 & 62.20 & 86.41 \\
    BundleTrack~\cite{wen2021bundletrack} & 76.39 & 89.36 & \textbf{94.30} & 99.54 \\
    BundleSDF~\cite{wen2023bundlesdf} & 70.64 & 89.41 & 92.30 & \textbf{99.98} \\
    FoundationPose~\cite{wen2024foundationpose} & \textbf{89.80} & \textbf{96.40} & 92.42 & 99.29 \\
    \hline
    \hline
    \end{tabular}
    }
\end{table}

By considering both ATE and RPE, we obtain a comprehensive understanding of the pose estimation performance in table~\cref{OPT_trajectory_metrics}, encompassing both global consistency and local accuracy.
Besides, we benchmark each method’s performance in table~\cref{OPT_AUC_metrics} across all 175 sequences (COPE-119 \& UOPE-56) using the AUC of ADD(S) and ADD(S)-0.1d. Due to failure cases in some sequences, FoundationPose~\cite{wen2024foundationpose} underperforms—especially in ADD(S)-0.1d compared to BundleTrack~\cite{wen2021bundletrack} and BundleSDF~\cite{wen2023bundlesdf}. Similarly, MaskFusion~\cite{runz2018maskfusion} shows poor performance, consistent with the trajectory error metrics. 

\section{CONCLUSION}
In this work, we propose an efficient framework for acquiring and annotating object pose estimation datasets in environments where both objects and cameras are in motion. Leveraging this approach and our environmental assumptions, we introduce a novel dataset tailored for various object pose tasks. We evaluate 18 state-of-the-art algorithms and provide reference performance benchmarks, validating the high value of the dataset. Future work includes deploying this data acquisition pipeline on a real-world robotic platform to further advance research in robot grasping and manipulation. Another promising direction is to augment the dataset with human hand pose annotations to facilitate research in AR/MR interaction.

{
\printbibliography[heading=bibintoc,title={References}]
}
\end{refsection}

\clearpage
\begin{refsection} 
\clearpage
\setcounter{page}{1}

\section{Supplementary}
\label{sec:supplementary}

\subsection{Experiments with Motion Capture System}
\subsubsection{Experiment Setup}
In addition to the original 175 sequences in our dataset, we collected 10 extra validation sequences shown in~\cref{fig:overview} with precise object pose ground truth using a motion capture system. We made every effort to ensure these 10 additional sequences closely mirror the experimental conditions of the original 175 sequences. These 10 sequences feature 10 different objects: 6 objects come from 6 categories in COPE-119, and 4 objects are from UOPE-56. 
To provide ground truth poses using the motion capture system, each object was affixed with at least three markers, forming a rigid body frame within the motion capture system coordinate system. We refer to the rigid body frames formed by markers attached to the object and the camera as the object body frame and camera body frame, respectively.
The position and orientation of object body in the motion capture system is denoted as $\textbf{T}_{OB, M}$. It is important to note that the orientation and position of this object body frame differ from those of the object's own coordinate system; there is a fixed but unknown transformation $\textbf{T}_{O, OB}$ between them. We determined this unknown transformation based on the geometric relationship
\begin{equation}
    \textbf{T}_{O,OB} =  (\textbf{T}_{OB,M})^{-1}\textbf{T}_{CB,M} \textbf{T}_{C,CB}\textbf{T}_{O,C} 
\end{equation},
where $\textbf{T}_{C,CB}$ is the handeye calibration result representing the transformation from camera to camera body frame, $\textbf{T}_{CB,M} ,\textbf{T}_{OB,M}$ are motion capture reading of camera body and object body position and orentation in motion capture system world coordinate, and $\textbf{T}^{O}_{C}$ is the object pose which can be estimated by algorithms like FoundationPose~\cite{wen2024foundationpose}. Note that, to ensure accuracy, in our setup, $\textbf{T}_{O,C}$ is estimated by ~\cite{wen2024foundationpose} first then refined by manual alignment of the object point cloud and its CAD model. Finally, the ground truth object pose is given by 
\begin{equation}    \textbf{T}_{O,C}= ( \textbf{T}_{C,CB} )^{-1}(\textbf{T}_{CB,M})^{-1}\textbf{T}_{OB,M} \textbf{T}_{O,OB}
\end{equation}. Please note that after the markers are attached to these 10 objects, we reconstructed object CAD models using an RGB-D scanner. However, because the markers themselves reflect infrared light, they interfered with the RGB-D scanner's depth readings, leading to imperfections in certain areas of the object mesh. Despite this, the overall quality remains usable. 

\begin{figure*}[!t]
\centering
\includegraphics[width=0.95\textwidth]{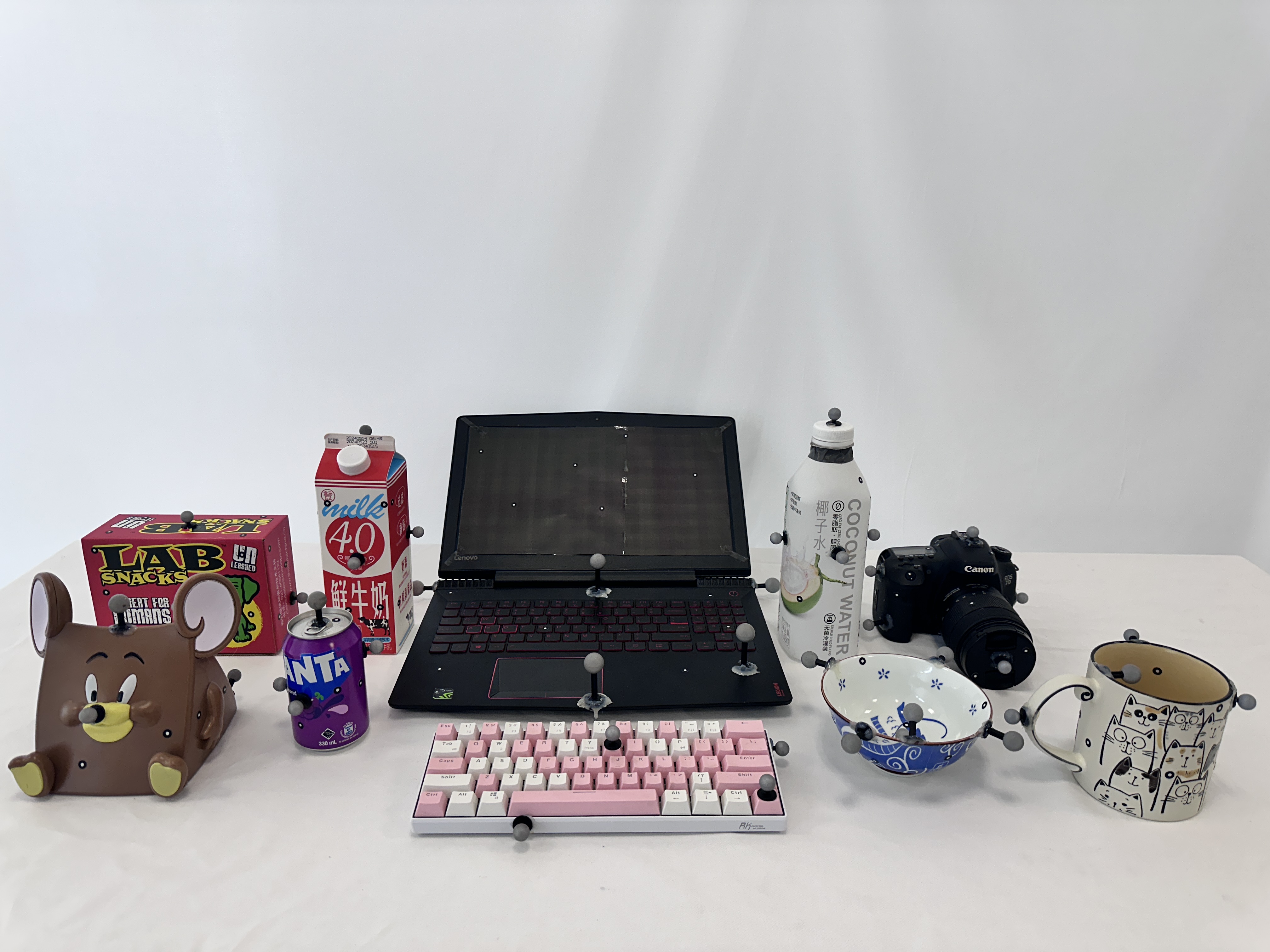}
\caption{The 10 objects in Mocap experiments we used.}
\label{fig:overview}
\end{figure*}
\subsubsection{Implementation Details}
We benchmark several key outputs against the ground truth object pose: the raw, unprocessed FoundationPose~\cite{wen2024foundationpose} (referred to as \textbf{Raw Abs-Pose}), the refined and smoothed ~\cite{wen2024foundationpose} by global Kalman Filter (\textbf{Refined Abs-Pose}), as well as the outputs of our relative pose estimator (\textbf{Rel-Pose}) and the pose graph optimization results (\textbf{PGO-Pose}). 
In pose graph optimization, the information matrix of the absolute pose edge is set heuristically. First, for each frame, we reproject the object bounding box and axis onto the 2D image based on the \textbf{Refined Abs-Pose} estimation. Then, we stitch these images into a video. By observing whether the projected bounding box in the video exhibits jitter, we can assess the quality of the absolute pose estimation. Typically, 2D projection jitter indicates an inconsistency between the object poses in consecutive frames. This means that the object poses across the few frames exhibiting this jitter will inevitably vary in quality—some may be good, some bad, or all of them might be poor. Consequently, these poses are considered unreliable. 

In general, the quality of absolute pose estimation is better than that of relative pose estimation. Therefore, we adopt a conservative parameter strategy, setting the diagonal elements of the absolute pose information matrix to 1e5 to reduce the negative impact of the relative pose edge. 

For those visually unreliable absolute edges, the information matrix elements are set to 1e2-1e3. Note that the elements of the relative pose edge information matrix are generally around 1e2. This allows the relative pose to effectively compensate for poor absolute pose estimations.
Finnally, for our comparative analysis, we utilize ATE (Absolute Trajectory Error) and RPE (Relative Pose Error). The unit of translation error is millimeters and the unit of rotation error is degrees.

\begin{table*}[h!]
    \centering
    \caption{ATE (mm) results.}
    \renewcommand{\arraystretch}{1.2}
    \resizebox{\textwidth}{!}{%
    \begin{tabular}{c|ccc|ccc|ccc|ccc}
    \toprule
    \hline
    \multirow{2}{*}{\textbf{Sequence ID}} & \multicolumn{3}{c|}{\textbf{Rel-Pose}} & \multicolumn{3}{c|}{\textbf{Raw Abs-Pose}} & \multicolumn{3}{c|}{\textbf{Refined Abs-Pose}} & \multicolumn{3}{c}{\textbf{PGO Pose}} \\
    \cline{2-13}
     & \textbf{ATE.} & \textbf{ATE.} & \textbf{ATE.} & \textbf{ATE.} & \textbf{ATE.} & \textbf{ATE.} & \textbf{ATE.} & \textbf{ATE.} & \textbf{ATE.} & \textbf{ATE.} & \textbf{ATE.} & \textbf{ATE.} \\
    & \textbf{max} & \textbf{mean} & \textbf{median} & \textbf{max} & \textbf{mean} & \textbf{median} & \textbf{max} & \textbf{mean} & \textbf{median} & \textbf{max} & \textbf{mean} & \textbf{median} \\
    \midrule
    Seq0 & 484.70 & 159.37 & 146.14 & 24.17 & 6.15 & 5.60 & \textbf{\color{darkgreen}16.85} & \textbf{\color{darkgreen}5.98} & \textbf{\color{darkgreen}5.56} & \textbf{\color{darkgreen}16.85} & 5.99 & \textbf{\color{darkgreen}5.56} \\
    Seq1 & 364.25 & 176.07 & 166.79 & \textbf{\color{darkgreen}17.01} & 4.90 & \textbf{\color{darkgreen}4.84} & \textbf{\color{darkgreen}17.01} & {4.89} & 4.86 & \textbf{\color{darkgreen}17.01} & \textbf{\color{darkgreen}4.88} & 4.86 \\
    Seq2 & 534.79 & 274.89 & 280.95 & \textbf{\color{darkgreen}17.16} & 5.15 & 4.92 & \textbf{\color{darkgreen}17.16} & \textbf{\color{darkgreen}5.06} & \textbf{\color{darkgreen}4.78} & \textbf{\color{darkgreen}17.16} & 5.07 & \textbf{\color{darkgreen}4.78} \\
    Seq3 & 228.37 & 85.35 & 83.48 & \textbf{\color{darkgreen}16.79} & 6.35 & \textbf{\color{darkgreen}5.37} & \textbf{\color{darkgreen}16.79} & 6.35 & 5.38 & \textbf{\color{darkgreen}16.79} & 6.35 & 5.38 \\
    Seq4 & 357.11 & 165.51 & 160.60 & \textbf{\color{darkgreen}16.87} & \textbf{\color{darkgreen}5.91} & \textbf{\color{darkgreen}5.84} & 16.88 & 5.92 & \textbf{\color{darkgreen}5.84} & 16.88 & 5.93 & \textbf{\color{darkgreen}5.84} \\
    Seq5 & 257.58 & 111.89 & 90.20 & \textbf{\color{darkgreen}7.59} & \textbf{\color{darkgreen}3.14} & \textbf{\color{darkgreen}3.00} & \textbf{\color{darkgreen}7.59} & 3.15 & 3.01 & \textbf{\color{darkgreen}7.59} & 3.15 & 3.01 \\
    Seq6 & 209.48 & 105.43 & 99.38 & 16.35 & \textbf{\color{darkgreen}5.70} & \textbf{\color{darkgreen}5.38} & \textbf{\color{darkgreen}16.34} & 5.71 & 5.39 & 16.35 & 5.72 & 5.41 \\
    Seq7 & 250.53 & 117.48 & 121.19 & 19.91 & 4.86 & 3.93 & \textbf{\color{darkgreen}18.48} & \textbf{\color{darkgreen}4.71} & \textbf{\color{darkgreen}3.87} & 18.50 & \textbf{\color{darkgreen}4.71} & \textbf{\color{darkgreen}3.87} \\
    Seq8 & 292.44 & 92.27 & 83.14 & 12.74 & 3.72 & 3.47 & \textbf{\color{darkgreen}12.73} & \textbf{\color{darkgreen}3.71} & \textbf{\color{darkgreen}3.46} & \textbf{\color{darkgreen}12.73} & \textbf{\color{darkgreen}3.71} & \textbf{\color{darkgreen}3.46} \\
    Seq9 & 296.40 & 153.10 & 165.70 & \textbf{\color{darkgreen}13.95} & 3.90 & 3.33 & \textbf{\color{darkgreen}13.95} & \textbf{\color{darkgreen}3.85} & \textbf{\color{darkgreen}3.25} & \textbf{\color{darkgreen}13.95} & \textbf{\color{darkgreen}3.85} & \textbf{\color{darkgreen}3.25} \\
    \hline
    \textbf{Avg.} & 327.37 & 144.62 & 139.76 & 16.65 & 4.98 & \textbf{\color{darkgreen}4.53} & \textbf{\color{darkgreen}15.38} & \textbf{\color{darkgreen}4.93} & 4.54 & \textbf{\color{darkgreen}15.38} & 4.94 & 4.54 \\
    \midrule
    \bottomrule
    \end{tabular}%
    }
    \label{tab:ate}
\end{table*}

\begin{table*}[h!]
    \centering
    \renewcommand{\arraystretch}{1.2}
    \caption{RPE results for the rot. (degs).}
    \resizebox{\textwidth}{!}{%
    \begin{tabular}{c|ccc|ccc|ccc|ccc}
    \toprule
    \hline
    \multirow{2}{*}{\textbf{Sequence ID}} & \multicolumn{3}{c|}{\textbf{Rel-Pose}} & \multicolumn{3}{c|}{\textbf{Raw Abs-Pose}} & \multicolumn{3}{c|}{\textbf{Refined Abs-Pose}} & \multicolumn{3}{c}{\textbf{PGO Pose}} \\
    \cline{2-13}
    & \textbf{RPE.} & \textbf{RPE.} & \textbf{RPE.} & \textbf{RPE.} & \textbf{RPE.} & \textbf{RPE.} & \textbf{RPE.} & \textbf{RPE.} & \textbf{RPE.} & \textbf{RPE.} & \textbf{RPE.} & \textbf{RPE.} \\
    & \textbf{rot.max} & \textbf{rot.mean} & \textbf{rot.median} & \textbf{rot.max} & \textbf{rot.mean} & \textbf{rot.median} & \textbf{rot.max} & \textbf{rot.mean} & \textbf{rot.median} & \textbf{rot.max} & \textbf{rot.mean} & \textbf{rot.median} \\
    \midrule
    Seq0 & 5.31 & 1.15 & 0.80 & 5.54 & 1.12 & 0.89 & 3.24 & \textbf{\textcolor{darkgreen}{0.21}} & \textbf{\textcolor{darkgreen}{0.14}} & \textbf{\textcolor{darkgreen}{3.21}} & \textbf{\textcolor{darkgreen}{0.21}} & \textbf{\textcolor{darkgreen}{0.14}} \\
    Seq1 & 7.54 & 1.05 & 0.78 & 9.80 & 0.95 & 0.75 & 5.36 & 0.15 & \textbf{\textcolor{darkgreen}{0.10}} & \textbf{\textcolor{darkgreen}{3.69}} & \textbf{\textcolor{darkgreen}{0.14}} & \textbf{\textcolor{darkgreen}{0.10}} \\
    Seq2 & 139.58 & 1.90 & 0.86 & 2.95 & 0.97 & 0.87 & \textbf{\textcolor{darkgreen}{0.45}} & \textbf{\textcolor{darkgreen}{0.13}} & \textbf{\textcolor{darkgreen}{0.12}} & 0.71 & \textbf{\textcolor{darkgreen}{0.13}} & \textbf{\textcolor{darkgreen}{0.12}} \\
    Seq3 & 8.71 & 0.96 & 0.80 & 5.64 & 1.31 & 1.11 & \textbf{\textcolor{darkgreen}{1.60}} & \textbf{\textcolor{darkgreen}{0.43}} & \textbf{\textcolor{darkgreen}{0.33}} & 2.88 & \textbf{\textcolor{darkgreen}{0.43}} & \textbf{\textcolor{darkgreen}{0.33}} \\
    Seq4 & 3.59 & 0.63 & 0.52 & 3.00 & 0.76 & 0.63 & \textbf{\textcolor{darkgreen}{2.86}} & \textbf{\textcolor{darkgreen}{0.36}} & \textbf{\textcolor{darkgreen}{0.27}} & \textbf{\textcolor{darkgreen}{2.86}} & \textbf{\textcolor{darkgreen}{0.36}} & \textbf{\textcolor{darkgreen}{0.27}} \\
    Seq5 & 3.21 & 0.77 & 0.64 & 3.29 & 0.83 & 0.72 & \textbf{\textcolor{darkgreen}{0.26}} & \textbf{\textcolor{darkgreen}{0.10}} & \textbf{\textcolor{darkgreen}{0.09}} & \textbf{\textcolor{darkgreen}{0.26}} & \textbf{\textcolor{darkgreen}{0.10}} & \textbf{\textcolor{darkgreen}{0.09}} \\
    Seq6 & 2.91 & 0.72 & 0.59 & 5.69 & 0.81 & 0.58 & 2.37 & \textbf{\textcolor{darkgreen}{0.14}} & \textbf{\textcolor{darkgreen}{0.09}} & \textbf{\textcolor{darkgreen}{1.44}} & \textbf{\textcolor{darkgreen}{0.14}} & \textbf{\textcolor{darkgreen}{0.09}} \\
    Seq7 & 3.86 & 0.78 & 0.64 & 4.94 & 0.80 & 0.61 & 0.57 & \textbf{\textcolor{darkgreen}{0.13}} & \textbf{\textcolor{darkgreen}{0.11}} & \textbf{\textcolor{darkgreen}{0.55}} & \textbf{\textcolor{darkgreen}{0.13}} & \textbf{\textcolor{darkgreen}{0.11}} \\
    Seq8 & 3.37 & 0.74 & 0.68 & 2.50 & 0.73 & 0.62 & 0.52 & \textbf{\textcolor{darkgreen}{0.11}} & \textbf{\textcolor{darkgreen}{0.09}} & \textbf{\textcolor{darkgreen}{0.49}} & \textbf{\textcolor{darkgreen}{0.11}} & \textbf{\textcolor{darkgreen}{0.09}} \\
    Seq9 & 5.44 & 0.85 & 0.58 &3.97& 0.79& 0.66 & 2.78 & \textbf{\textcolor{darkgreen}{0.16}} & \textbf{\textcolor{darkgreen}{0.13}} & \textbf{\textcolor{darkgreen}{1.77}} & \textbf{\textcolor{darkgreen}{0.16}} & \textbf{\textcolor{darkgreen}{0.13}} \\
    \hline
    \textbf{Avg.} & 18.05 & 0.95 & 0.69 & 4.93 & 0.91 & 0.74 & 2.00 & \textbf{\textcolor{darkgreen}{0.19}} & \textbf{\textcolor{darkgreen}{0.15}} & \textbf{\textcolor{darkgreen}{1.79}} & \textbf{\textcolor{darkgreen}{0.19}} & \textbf{\textcolor{darkgreen}{0.15}} \\
    \midrule
    \bottomrule
    \end{tabular}%
    }
    \label{tab:rpe-rot}
\end{table*}

\begin{table*}[hbtp]
    \centering
    \renewcommand{\arraystretch}{1.2}
    \caption{RPE results for the trans. (mm).}
    \resizebox{\textwidth}{!}{%
    \begin{tabular}{c|ccc|ccc|ccc|ccc}
    \toprule
    \hline
    \multirow{2}{*}{\textbf{Sequence ID}} & \multicolumn{3}{c|}{\textbf{Rel-Pose}} & \multicolumn{3}{c|}{\textbf{Raw Abs-Pose}} & \multicolumn{3}{c|}{\textbf{Refined Abs-Pose}} & \multicolumn{3}{c}{\textbf{PGO Pose}} \\
    \cline{2-13}
    & \textbf{RPE.} & \textbf{RPE.} & \textbf{RPE.} & \textbf{RPE.} & \textbf{RPE.} & \textbf{RPE.} & \textbf{RPE.} & \textbf{RPE.} & \textbf{RPE.} & \textbf{RPE.} & \textbf{RPE.} & \textbf{RPE.} \\
    & \textbf{trans.max} & \textbf{trans.mean} & \textbf{trans.median} & \textbf{trans.max} & \textbf{trans.mean} & \textbf{trans.median} & \textbf{trans.max} & \textbf{trans.mean} & \textbf{trans.median} & \textbf{trans.max} & \textbf{trans.mean} & \textbf{trans.median} \\
    \midrule
    Seq0 & 70.27 & 12.72 & 9.16 & 18.49 & 3.80 & 3.25 & 18.45 & \textbf{\color{darkgreen}3.72} & \textbf{\color{darkgreen}3.17} & \textbf{\color{darkgreen}13.98} & 3.73 & 3.18 \\
    Seq1 & 56.04 & 10.16 & 7.97 & 18.60 & 2.17 & 1.66 & 18.39 & \textbf{\color{darkgreen}2.11} & \textbf{\color{darkgreen}1.64} & \textbf{\color{darkgreen}17.18 } & 2.12 & \textbf{\color{darkgreen}1.64}\\
    Seq2 & 656.47 & 16.70 & 9.84 & 11.71 & 2.89 & 2.38 & \textbf{\color{darkgreen}11.67} & \textbf{\color{darkgreen}2.87} & \textbf{\color{darkgreen}2.33} & \textbf{\color{darkgreen}11.67} & 2.88 & \textbf{\color{darkgreen}2.33} \\
    Seq3 & 37.90 & 8.66 & 7.70 & \textbf{\color{darkgreen}18.67} & 3.31 & 2.10 & 18.74 & \textbf{\color{darkgreen}3.28} & \textbf{\color{darkgreen}2.05} & 18.74 & \textbf{\color{darkgreen}3.28} & \textbf{\color{darkgreen}2.05} \\
    Seq4 & 32.24 & 9.92 & 8.79 & 13.85 & 2.68 & 2.13 & 13.83 & \textbf{\color{darkgreen}2.65} & \textbf{\color{darkgreen}2.07} & \textbf{\color{darkgreen}12.63} & \textbf{\color{darkgreen}2.65} & \textbf{\color{darkgreen}2.07} \\
    Seq5 & 41.07 & 6.98 & 6.09 & \textbf{\color{darkgreen}6.49} & 1.28 & 1.03 & 6.51 & \textbf{\color{darkgreen}1.27} & \textbf{\color{darkgreen}1.00} & 6.51 & \textbf{\color{darkgreen}1.27} & \textbf{\color{darkgreen}1.00} \\
    Seq6 & 34.65 & 8.84 & 7.37 & 15.16 & 3.50 & 3.14 & \textbf{\color{darkgreen}15.12} & \textbf{\color{darkgreen}3.47} & \textbf{\color{darkgreen}3.06} & \textbf{\color{darkgreen}15.12} & 3.49 & 3.11 \\
    Seq7 & 35.43 & 8.35 & 6.72 & 18.15 & 2.37 & 1.51 & 18.16 & \textbf{\color{darkgreen}2.27} & \textbf{\color{darkgreen}1.46} & \textbf{\color{darkgreen}17.85} & \textbf{\color{darkgreen}2.27} & \textbf{\color{darkgreen}1.46} \\
    Seq8 & 95.26 & 7.78 & 6.16 & \textbf{\color{darkgreen}11.75} & 1.53 & 1.14 & 11.76 & \textbf{\color{darkgreen}1.49} & \textbf{\color{darkgreen}1.13} & 11.76 & \textbf{\color{darkgreen}1.49} & \textbf{\color{darkgreen}1.13} \\
    Seq9 & 27.24 & 6.78 & 6.05 & 10.58 & 1.96 & 1.72 & \textbf{\color{darkgreen}10.23} & \textbf{\color{darkgreen}1.94} & \textbf{\color{darkgreen}1.69} & \textbf{\color{darkgreen}10.23} & \textbf{\color{darkgreen}1.94} & \textbf{\color{darkgreen}1.69} \\
    \hline
    \textbf{Avg.} & 108.66 & 9.69 & 7.59 & 14.35 & 2.55 & 2.01 & 13.35 & \textbf{\color{darkgreen}2.51} & \textbf{\color{darkgreen}1.96} & \textbf{\color{darkgreen}12.63} & \textbf{\color{darkgreen}2.51} & 1.97 \\
    \midrule
    \bottomrule
    \end{tabular}%
    }
    \label{tab:rpe-trans}
\end{table*}

\begin{figure*}[hbtp]
    \centering
    \begin{subfigure}[b]{0.18\textwidth}
        \centering
        \includegraphics[width=\linewidth]{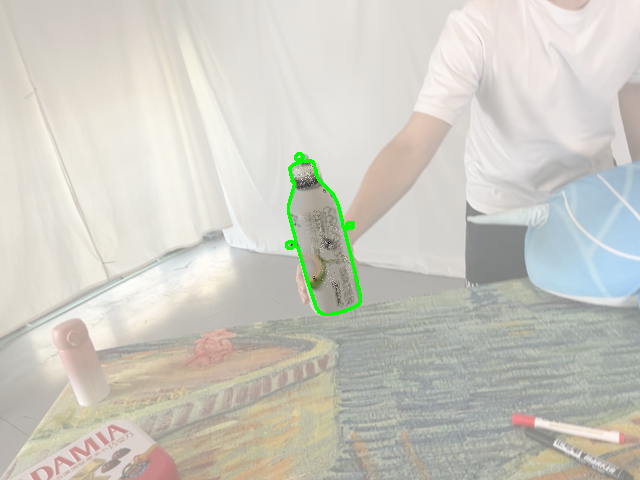}
        \caption{Seq0 - Mocap}
        \label{fig:seq0_method1}
    \end{subfigure}
    \hfill
    \begin{subfigure}[b]{0.18\textwidth}
        \centering
        \includegraphics[width=\linewidth]{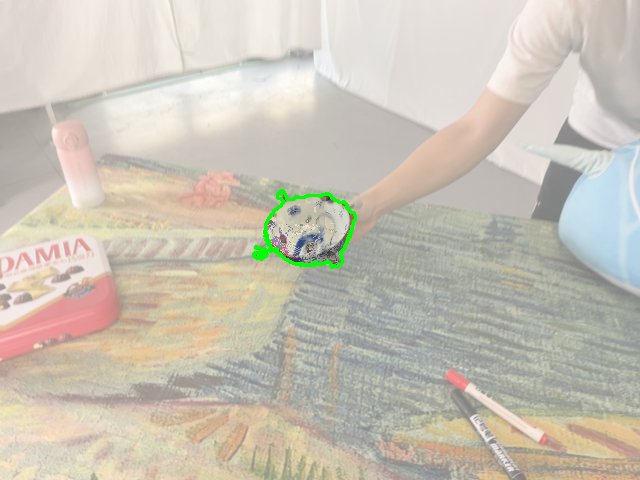}
        \caption{Seq1 - Mocap}
        \label{fig:seq1_method1}
    \end{subfigure}
    \hfill
    \begin{subfigure}[b]{0.18\textwidth}
        \centering
        \includegraphics[width=\linewidth]{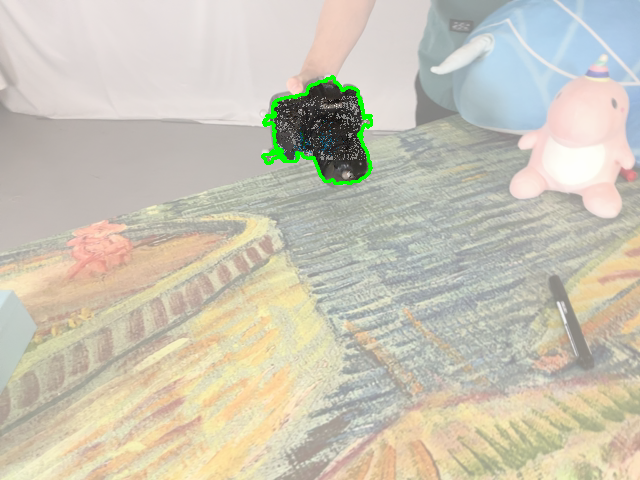}
        \caption{Seq2 - Mocap}
        \label{fig:seq2_method1}
    \end{subfigure}
    \hfill
    \begin{subfigure}[b]{0.18\textwidth}
        \centering
        \includegraphics[width=\linewidth]{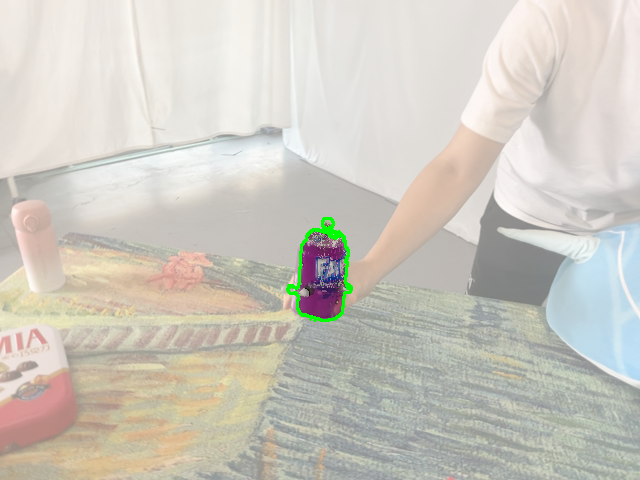}
        \caption{Seq3 - Mocap}
        \label{fig:seq3_method1}
    \end{subfigure}
    \hfill
    \begin{subfigure}[b]{0.18\textwidth}
        \centering
        \includegraphics[width=\linewidth]{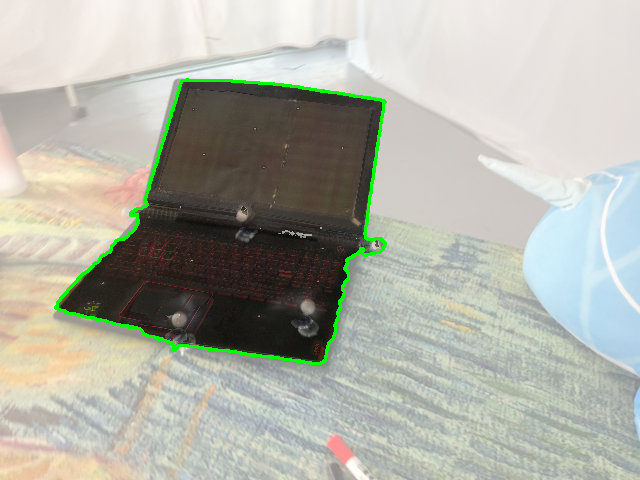}
        \caption{Seq4 - Mocap}
        \label{fig:seq4_method1}
    \end{subfigure}
    
    \vspace{0.5cm}
    
    \begin{subfigure}[b]{0.18\textwidth}
        \centering
        \includegraphics[width=\linewidth]{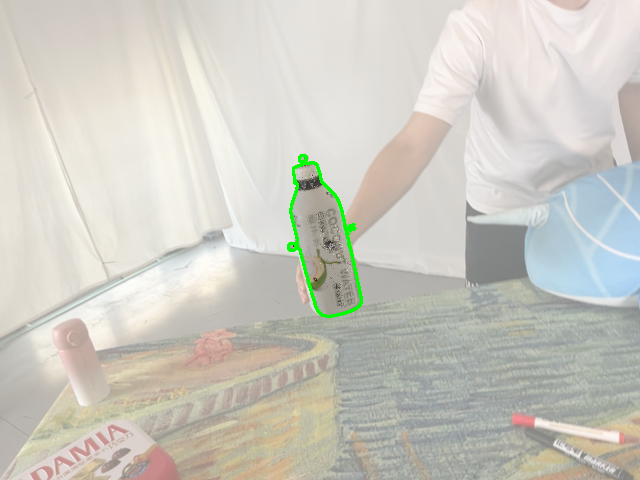}
        \caption{Seq0 - PGO}
        \label{fig:seq0_method2}
    \end{subfigure}
    \hfill
    \begin{subfigure}[b]{0.18\textwidth}
        \centering
        \includegraphics[width=\linewidth]{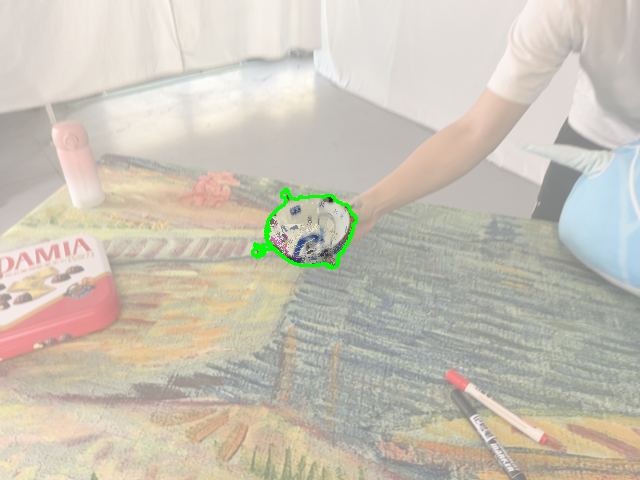}
        \caption{Seq1 - PGO}
        \label{fig:seq1_method2}
    \end{subfigure}
    \hfill
    \begin{subfigure}[b]{0.18\textwidth}
        \centering
        \includegraphics[width=\linewidth]{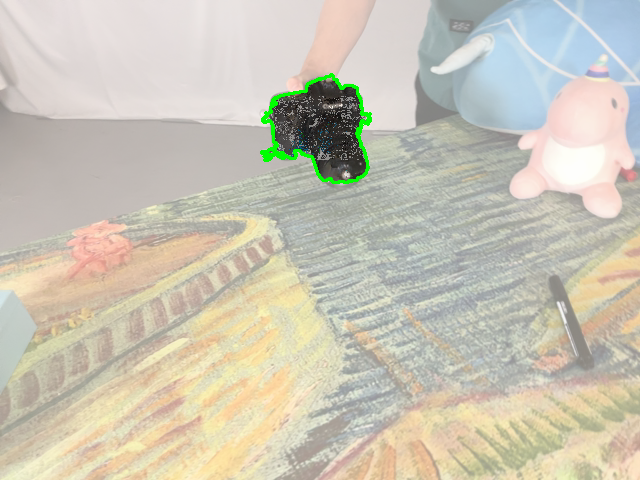}
        \caption{Seq2 - PGO}
        \label{fig:seq2_method2}
    \end{subfigure}
    \hfill
    \begin{subfigure}[b]{0.18\textwidth}
        \centering
        \includegraphics[width=\linewidth]{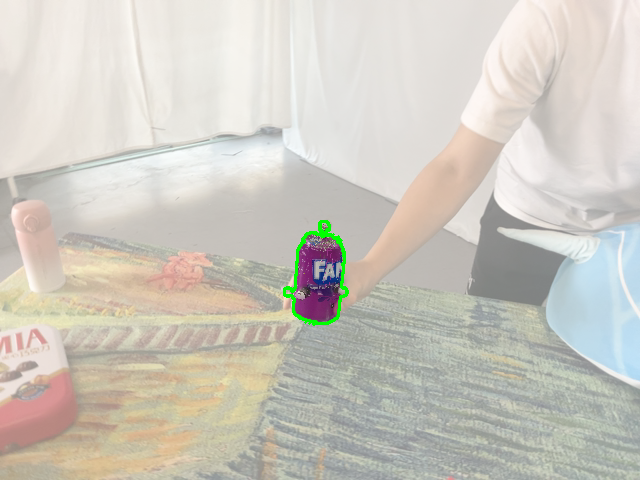}
        \caption{Seq3 - PGO}
        \label{fig:seq3_method2}
    \end{subfigure}
    \hfill
    \begin{subfigure}[b]{0.18\textwidth}
        \centering
        \includegraphics[width=\linewidth]{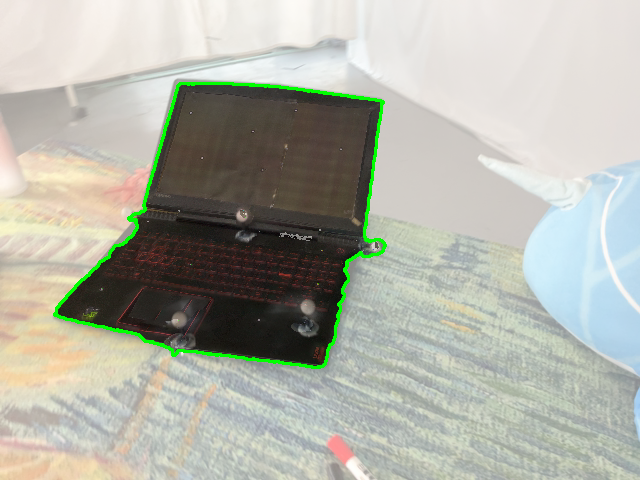}
        \caption{Seq4 - PGO}
        \label{fig:seq4_method2}
    \end{subfigure}
    
    \caption{Comparison of algorithm performance across Seq0 - Seq4 sequences. The top row shows Mocap results, and the bottom row shows PGO results for the same sequences.}
    \label{fig:algorithm_comparison}
\end{figure*}

\begin{figure*}[hbtp]
    \centering
    \begin{subfigure}[b]{0.18\textwidth}
        \centering
        \includegraphics[width=\linewidth]{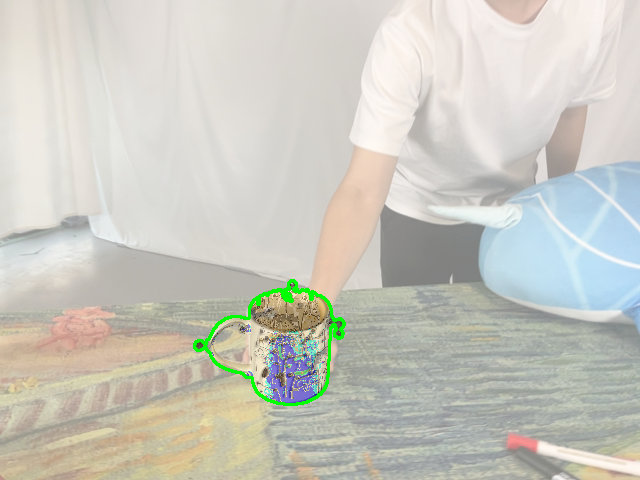}
        \caption{Seq5 - Mocap}
        \label{fig:seq5_method1}
    \end{subfigure}
    \hfill
    \begin{subfigure}[b]{0.18\textwidth}
        \centering
        \includegraphics[width=\linewidth]{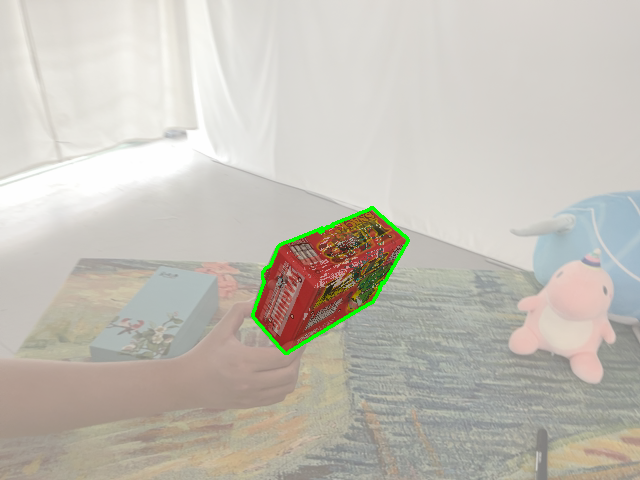}
        \caption{Seq6 - Mocap}
        \label{fig:seq6_method1}
    \end{subfigure}
    \hfill
    \begin{subfigure}[b]{0.18\textwidth}
        \centering
        \includegraphics[width=\linewidth]{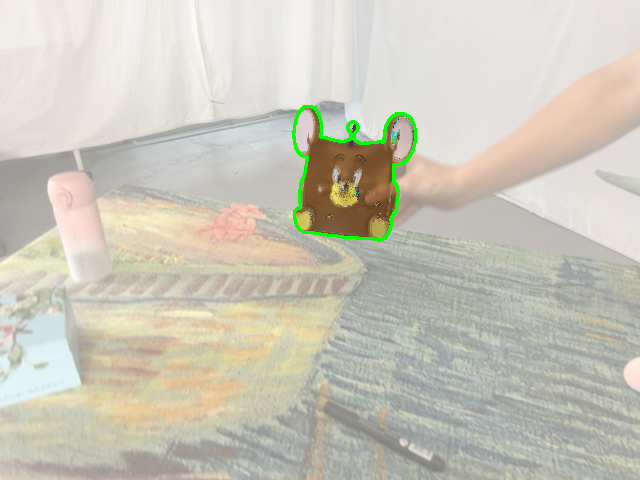}
        \caption{Seq7 - Mocap}
        \label{fig:seq7_method1}
    \end{subfigure}
    \hfill
    \begin{subfigure}[b]{0.18\textwidth}
        \centering
        \includegraphics[width=\linewidth]{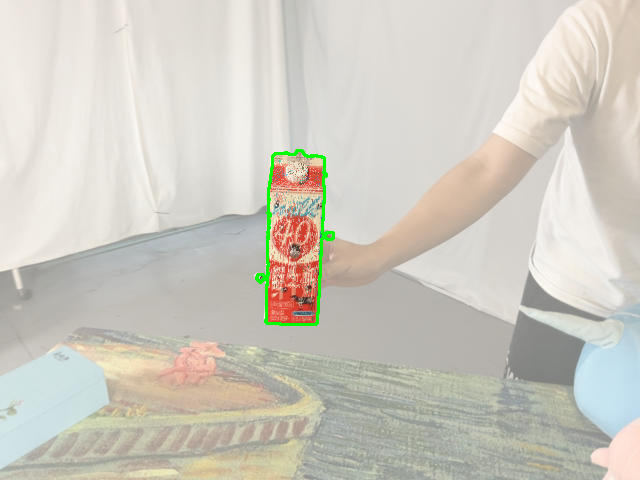}
        \caption{Seq8 - Mocap}
        \label{fig:seq8_method1}
    \end{subfigure}
    \hfill
    \begin{subfigure}[b]{0.18\textwidth}
        \centering
        \includegraphics[width=\linewidth]{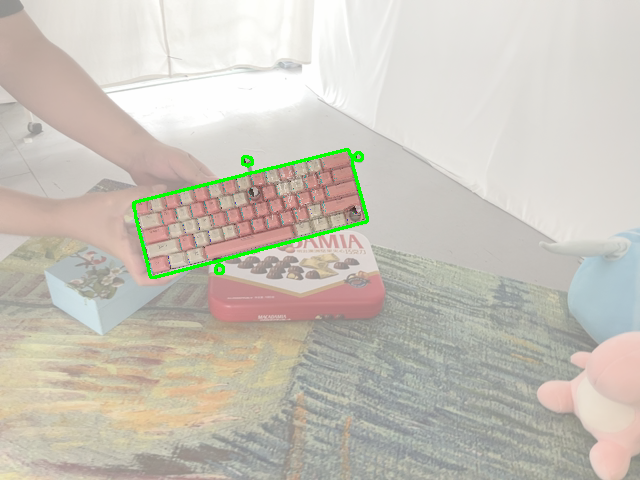}
        \caption{Seq9 - Mocap}
        \label{fig:seq9_method1}
    \end{subfigure}
    
    \vspace{0.5cm}
    
    \begin{subfigure}[b]{0.18\textwidth}
        \centering
        \includegraphics[width=\linewidth]{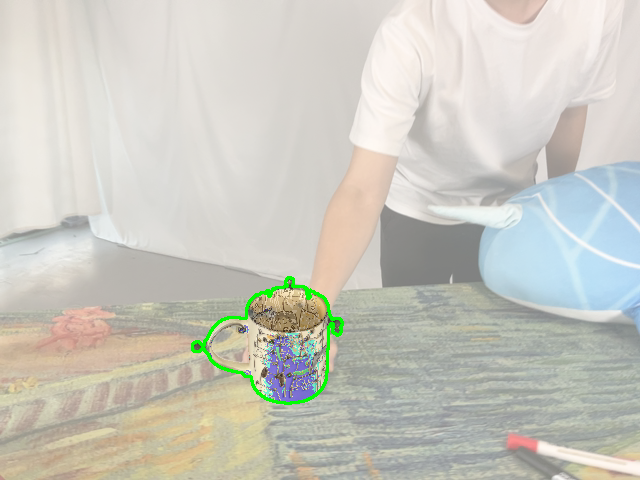}
        \caption{Seq5 - PGO}
        \label{fig:seq5_method2}
    \end{subfigure}
    \hfill
    \begin{subfigure}[b]{0.18\textwidth}
        \centering
        \includegraphics[width=\linewidth]{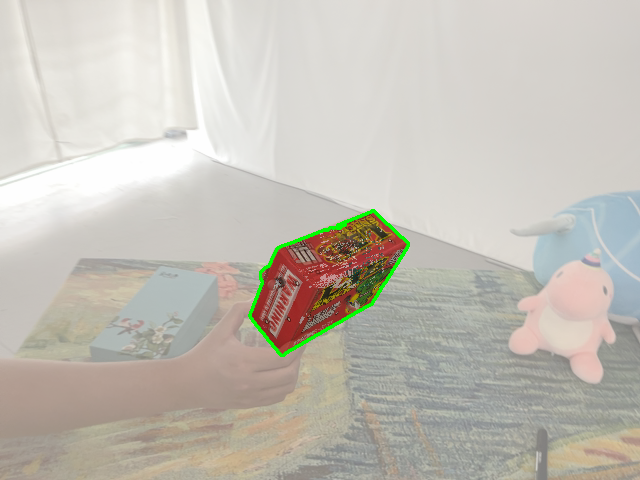}
        \caption{Seq6 - PGO}
        \label{fig:seq6_method2}
    \end{subfigure}
    \hfill
    \begin{subfigure}[b]{0.18\textwidth}
        \centering
        \includegraphics[width=\linewidth]{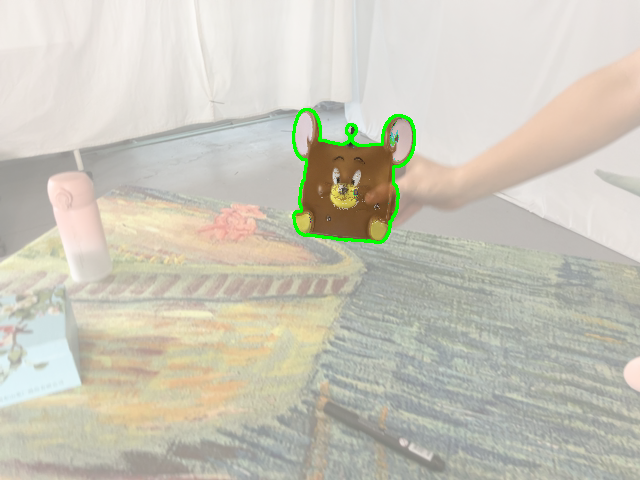}
        \caption{Seq7 - PGO}
        \label{fig:seq7_method2}
    \end{subfigure}
    \hfill
    \begin{subfigure}[b]{0.18\textwidth}
        \centering
        \includegraphics[width=\linewidth]{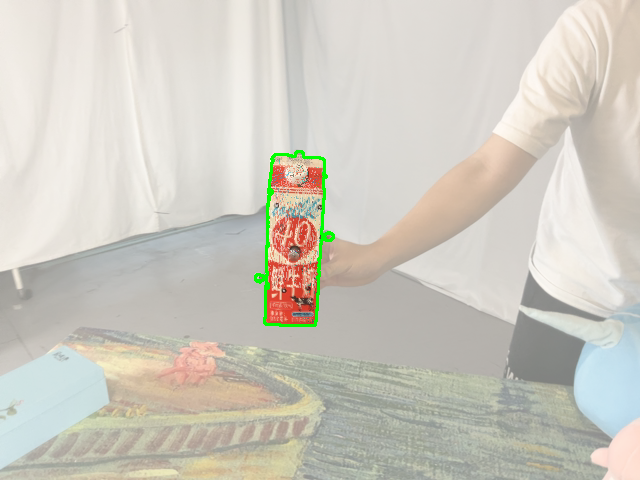}
        \caption{Seq8 - PGO}
        \label{fig:seq8_method2}
    \end{subfigure}
    \hfill
    \begin{subfigure}[b]{0.18\textwidth}
        \centering
        \includegraphics[width=\linewidth]{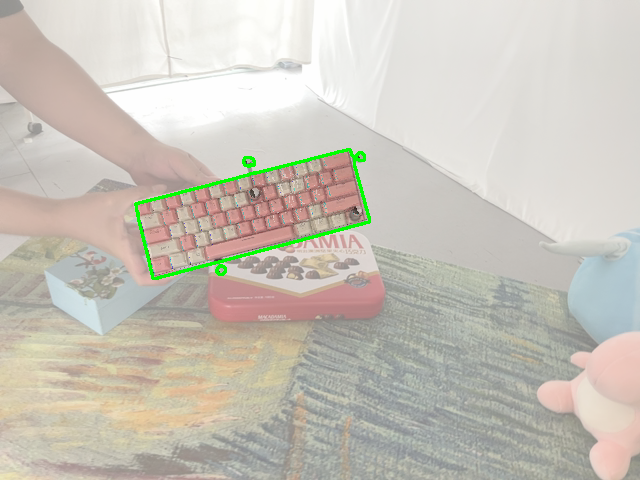}
        \caption{Seq9 - PGO}
        \label{fig:seq9_method2}
    \end{subfigure}
    
    \caption{Comparison of algorithm performance across Seqs5 - Seqs9 sequences. Top row shows Mocap results, and the bottom row shows PGO results for the same sequences.}
    \label{fig:algorithm_comparison}
\end{figure*}

\subsubsection{Results}
Here are our experimental comparison results, where we report the mean, max, and median for ATE shown in~\cref{tab:ate} and RPE shown in~\cref{tab:rpe-rot} and~\cref{tab:rpe-trans}.
Additionally, we randomly select several frames to present qualitative visualization results.
The results show that our final ATE and RPE are very close to the ground truth, with an average translation error of less than $3$ mm and an average rotation error of less than $0.2$ degree, indicating quite good quality. Furthermore, while the overall results after pose graph optimization didn't change significantly, the maximum error decreases. This aligns with the design goal of this module: effectively compensating for poor estimations in \textbf{Refined Abs-Pose} and preserving the good ones. From the perspective of relative pose estimation itself, its drift is still noticeable, and its overall accuracy cannot match that of \textbf{Refined Abs-Pose}. Therefore, it can only serve to compensate for low-quality absolute poses.

\subsection{Analysis of Unseen Object Pose Estimation Methods}

\begin{figure}[thbp]
    \centering
    \includegraphics[width=0.4\textwidth]{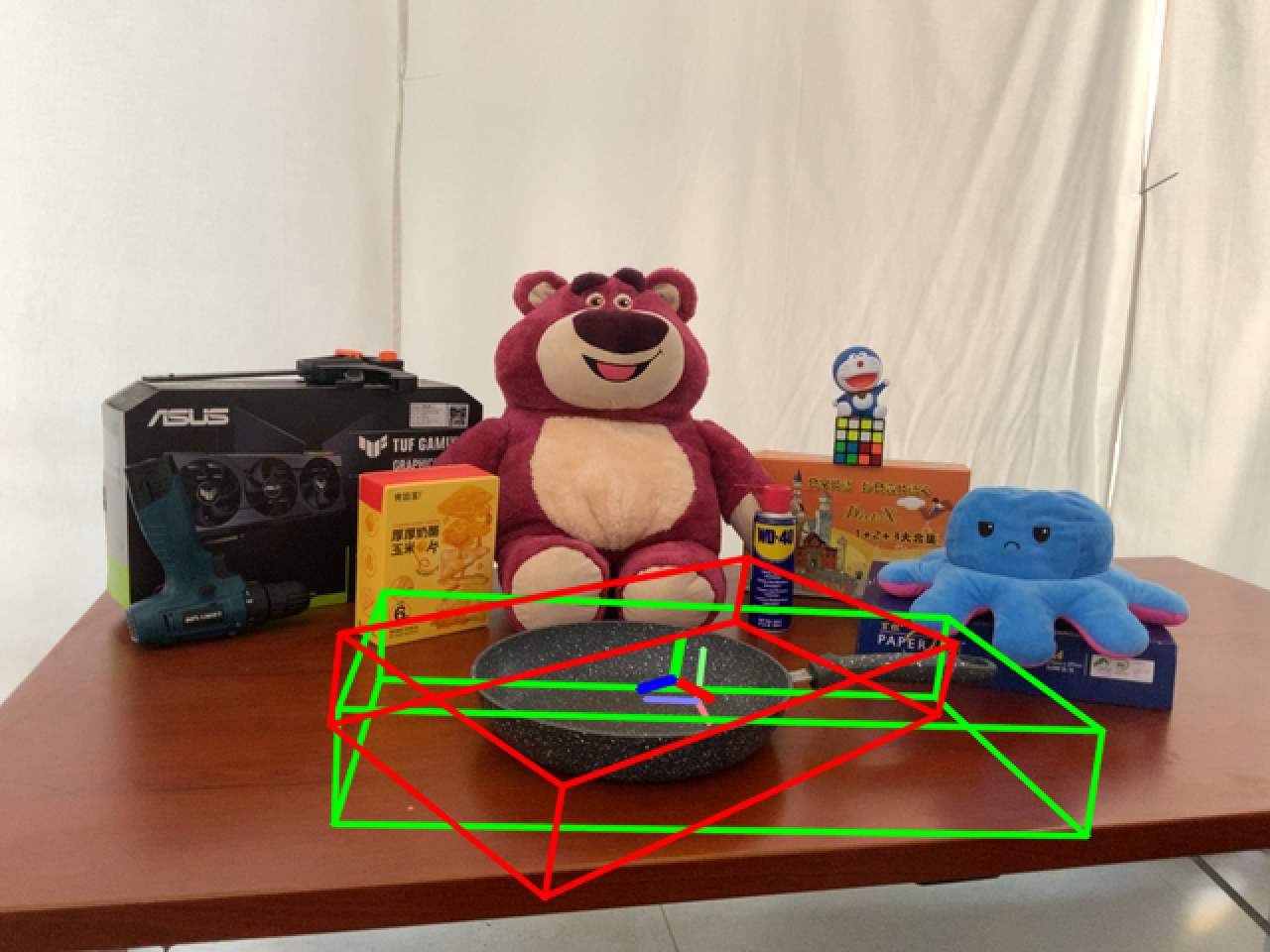}
    \caption{FoundPose\cite{ornek2024foundpose}* failure case.}
    \label{fig:UOPE_Foundpose}
\end{figure}

Illustrated in \cref{fig:UOPE_Foundpose}, \cite{ornek2024foundpose}* heavily relies on features extracted by DINOv2~\cite{oquab2023dinov2} to establish 2D-3D correspondences, which makes it susceptible to pose estimation errors for low-texture or highly symmetric objects. \cref{fig:UOPE_Foundpose} shows that the object (a frying pan) is textureless, and during the detection process, the right-side handle was mistakenly ignored. As a result, the object was erroneously treated as symmetric, leading to incorrect pose estimates.

\begin{figure}[thbp]
    \centering
    \includegraphics[width=0.4\textwidth]{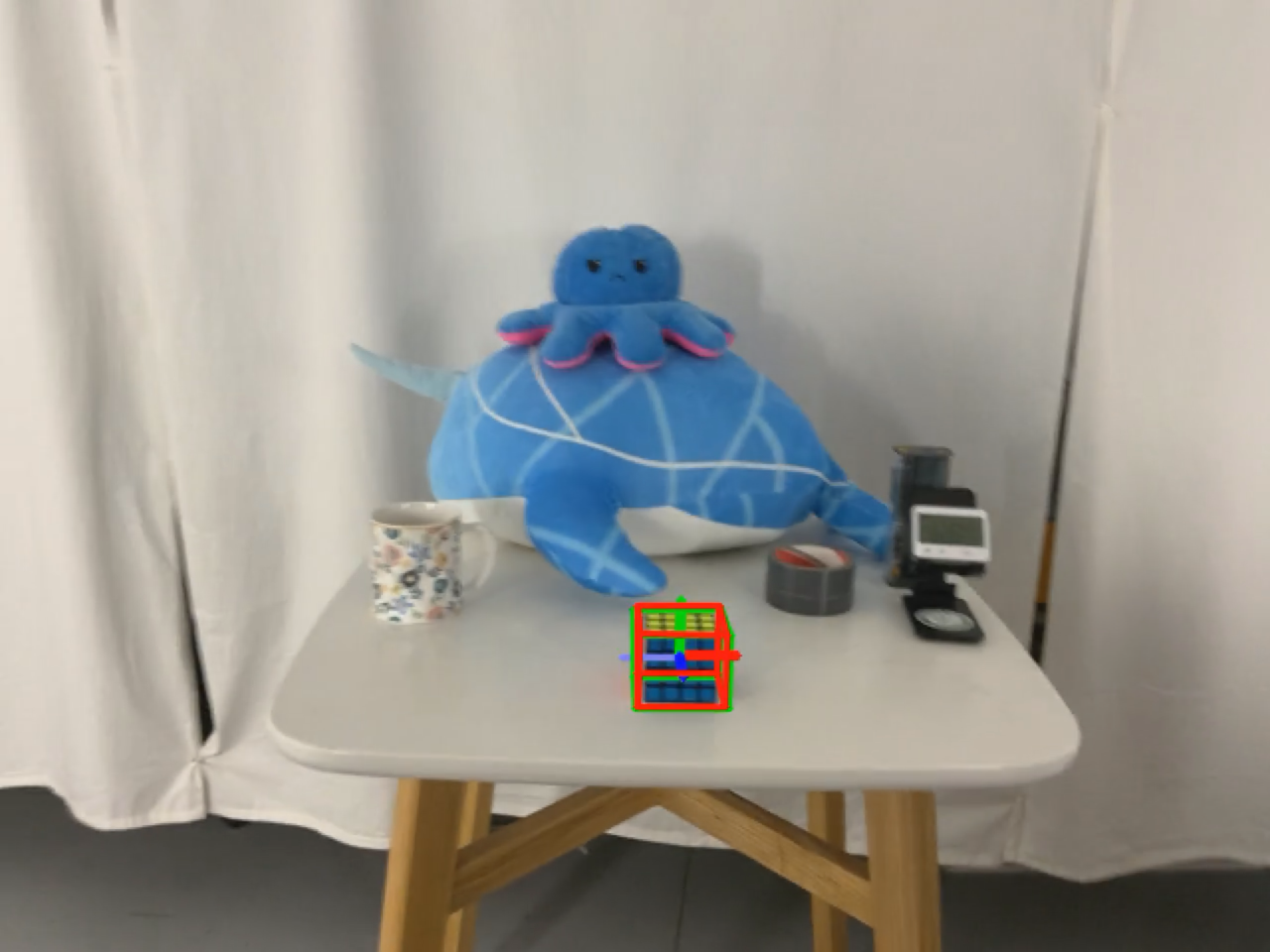}
    \caption{FoundationPose\cite{wen2024foundationpose} failure case. }
    \label{fig:UOPE_Fdpose}
\end{figure}

As shown in \cref{fig:UOPE_Fdpose}, we directly employ the model-based \cite{wen2024foundationpose}. It primarily relies on the input RGB-D data. When the camera undergoes rapid motion, causing motion blur, the pose estimation results becomes error-prone as well. This issue is especially prominent for flat, textureless, and symmetric objects, where the network may incorrectly estimate the orientation of the coordinate axes. 

\begin{figure}[thbp]
    \centering
    \includegraphics[width=0.4\textwidth]{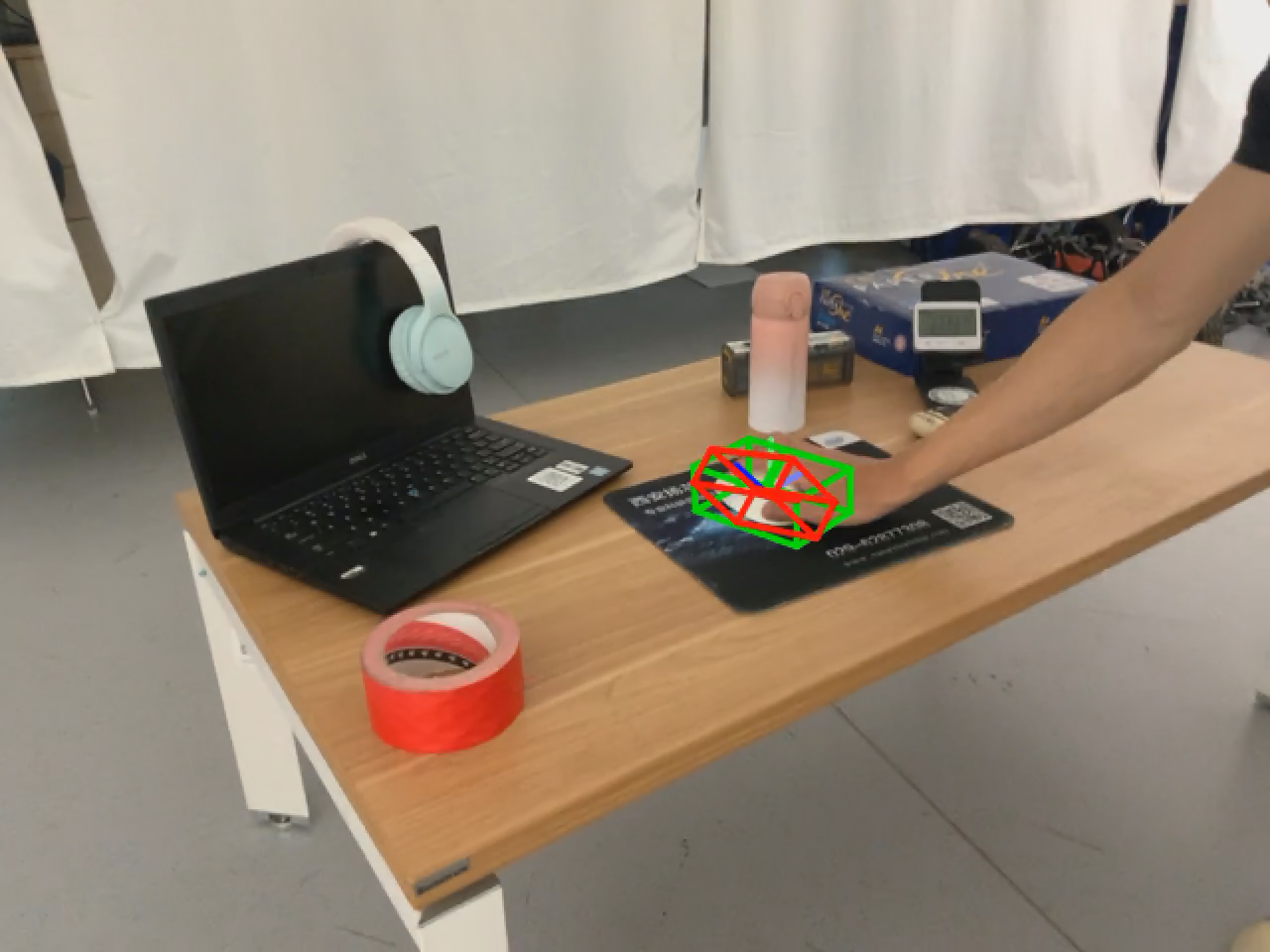}
    \caption{GigaPose~\cite{nguyen2024gigapose} failure case. }
    \label{fig:UOPE_Gigapose}
\end{figure}

The performance of GigaPose~\cite{nguyen2024gigapose} benefits from its time-intensive pre-processing step, where it renders the input CAD model extensively before performing pose estimation. This process allows it to establish better correspondences during object pose estimation. However, since the method relies solely on RGB information and performs feature matching across different viewpoints of the CAD model, it faces significant challenges when dealing with occlusions or textureless objects as shown in \cref{fig:UOPE_Gigapose}, such limitations result in a failure case where the person’s hand holding the mouse causes the estimation to break down.

\begin{figure}[htbp]
    \centering
        \begin{subfigure}[b]{0.2\textwidth}
            \includegraphics[width=\linewidth]{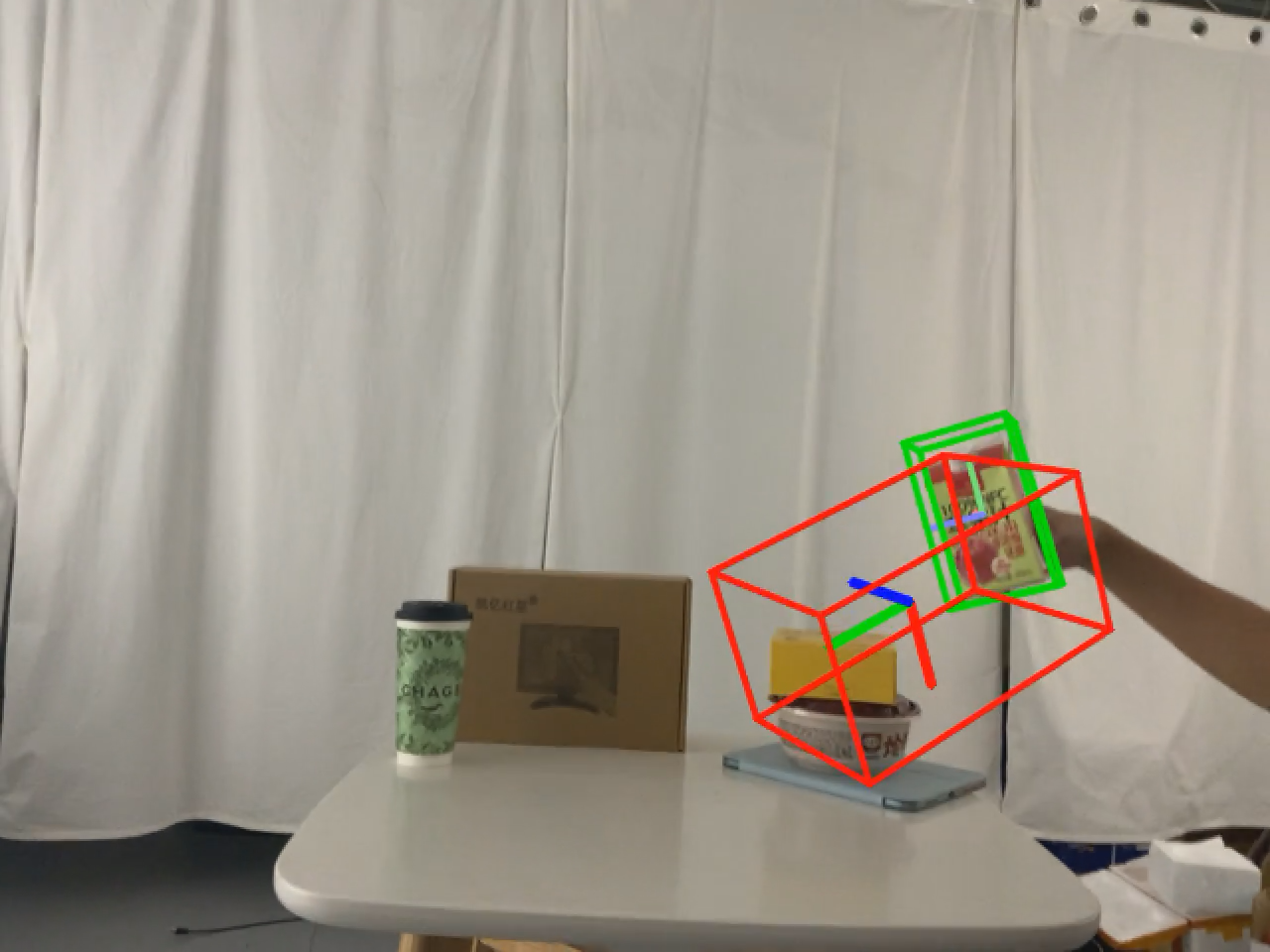}
            \caption{MegaPose (RGB)}
            
        \end{subfigure}
        \begin{subfigure}[b]{0.2\textwidth}
            \includegraphics[width=\linewidth]{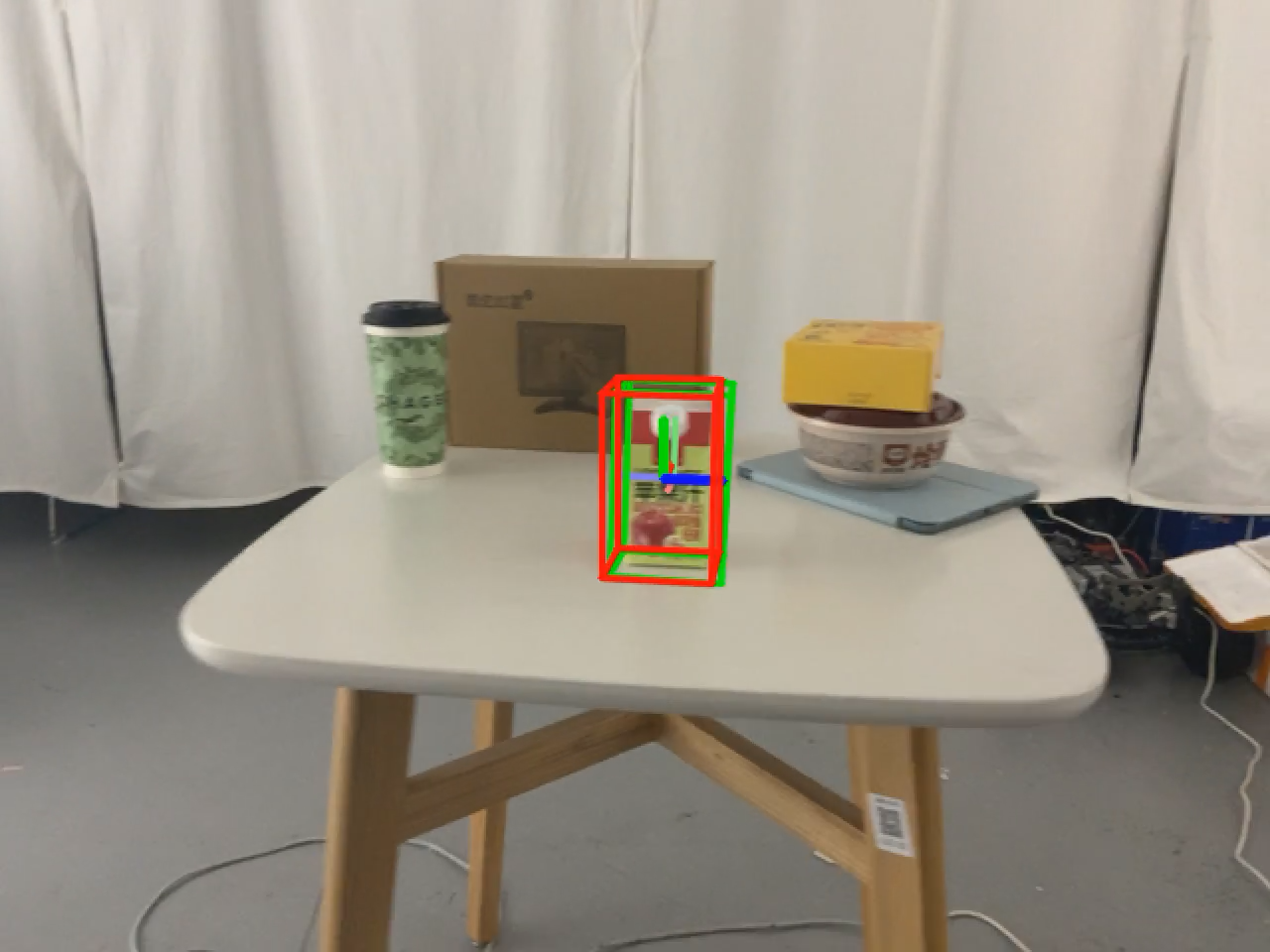}
            \caption{MegaPose (RGB-D)}
            
        \end{subfigure}
        \caption{MegaPose\cite{labbe2022megapose} failure case.}
        \label{fig:mega}
\end{figure}  

As shown in \cref{fig:mega}, here we evaluate both the RGB-based and RGB-D-based variants of \cite{labbe2022megapose}. This method takes a cropped image (RGB or RGB-D) as input along with the corresponding CAD model. Multiple poses are rendered from the CAD model and then passed to the coarse estimator to obtain the initial poses, which are subsequently refined by the refinement network to produce the final outputs. Note that the final results heavily depend on the coarse estimator network. If the initial pose estimation is inaccurate, then the refinement stage is generally unable to recover the correct poses. Therefore, both RGB and RGB-D approaches are highly dependent on the quality of the multiview input poses. For geometrically symmetric or regular-shaped objects such as the juice box shown in \cref{fig:mega}, the use of multiple poses from the CAD model can easily lead to incorrect initial guesses. Consequently, failure cases are observed in both the RGB-based and RGB-D-based versions of the method, as illustrated in the figure~\cref{fig:mega}.

\begin{figure}[thbp]
    \centering
    \includegraphics[width=0.4\textwidth]{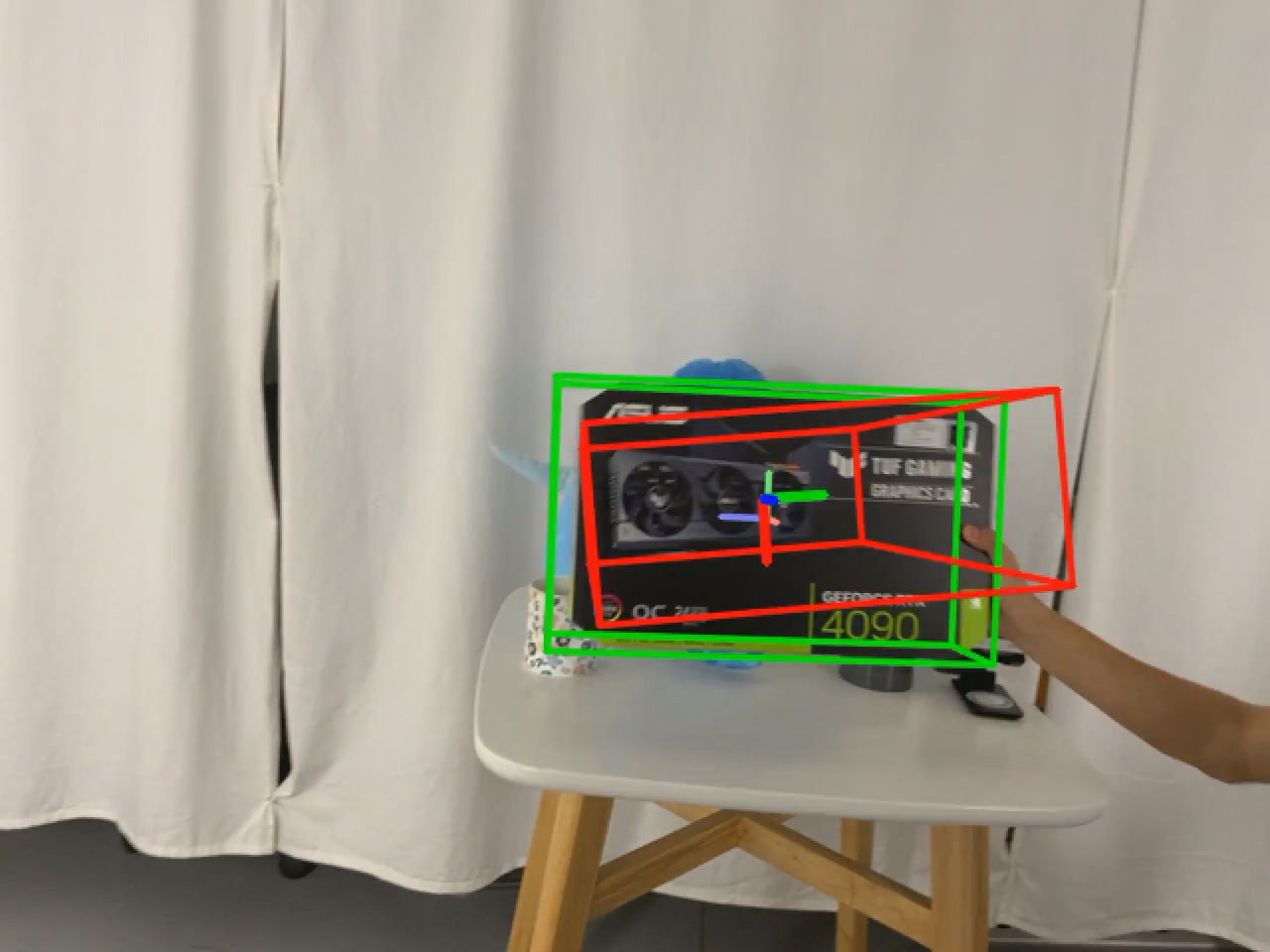}
    \caption{SAM6D\cite{lin2024sam} failure case. }
    \label{fig:UOPE_SAM6D}
\end{figure}
SAM6D~\cite{lin2024sam} is a multi-stage method. It first segments any objects using SAM\cite{kirillov2023segment}, and then associate the segmentation mask to the target object. And through the \cite{qin2023geotransformer} to estimate the object pose. Incorrect registration may happen in symmetric and textureless object cases, resulting in failure cases such as the one shown in \cref{fig:UOPE_SAM6D}.

\subsection{Analysis of Object Pose Tracking Methods}
\begin{figure}[thbp]
    \centering
    \includegraphics[width=0.4\textwidth]{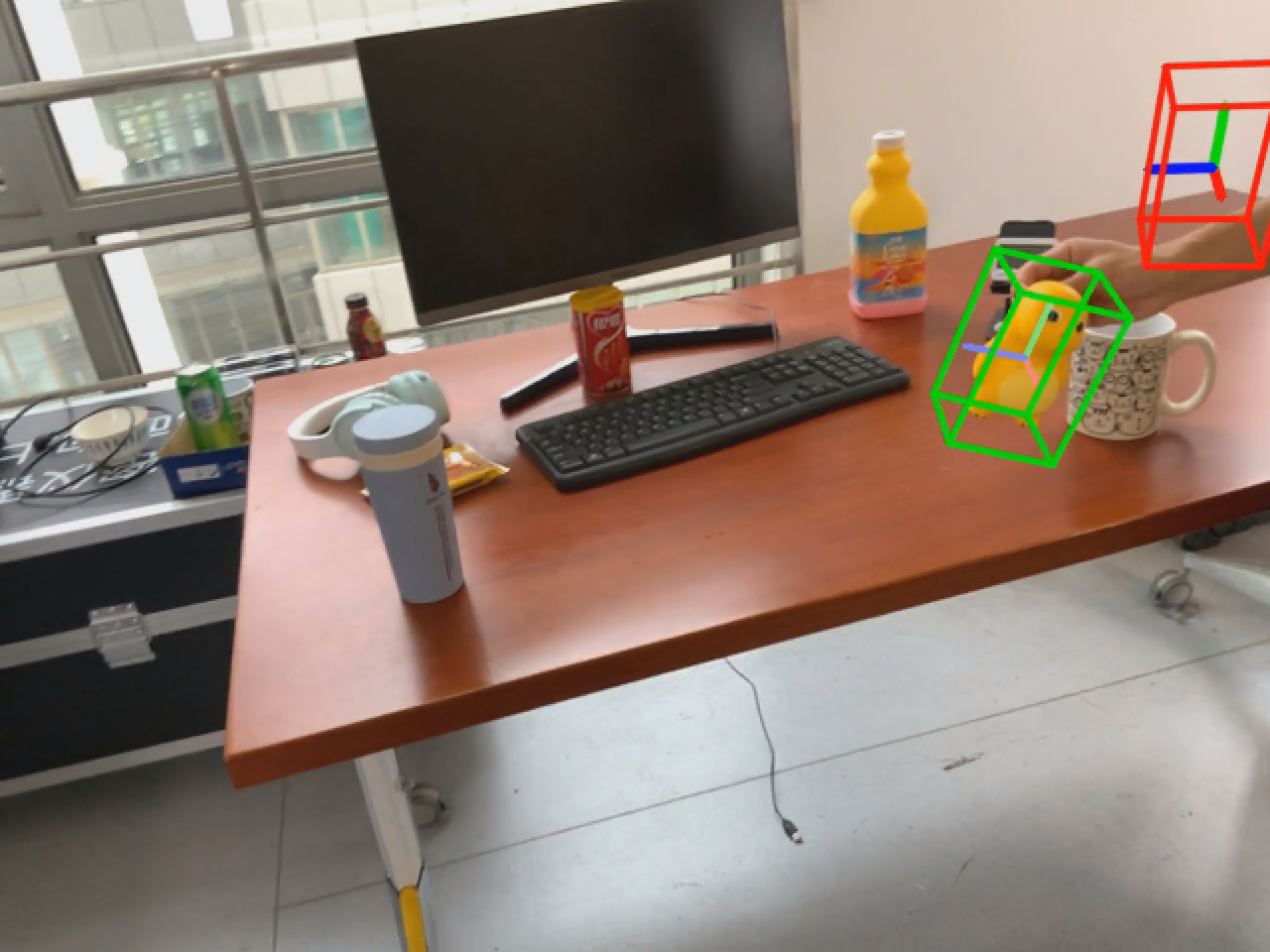}
    \caption{MaskFusion~\cite{runz2018maskfusion} failure case.}
    \label{fig:OPT_maskfusion}
\end{figure}

For \cite{runz2018maskfusion}, to ensure a fair benchmark, we replaced its original mask results with masks generated by the state-of-the-art method SAM2~\cite{ravi2024sam}. However, when intensive motion occurs, the object tracking algoritm tends to reinitialize its state. Consequently, this leads to noticeable drift. An example failure case is shown in \cref{fig:OPT_maskfusion}

\begin{figure}[thbp]
    \centering
    \includegraphics[width=0.4\textwidth]{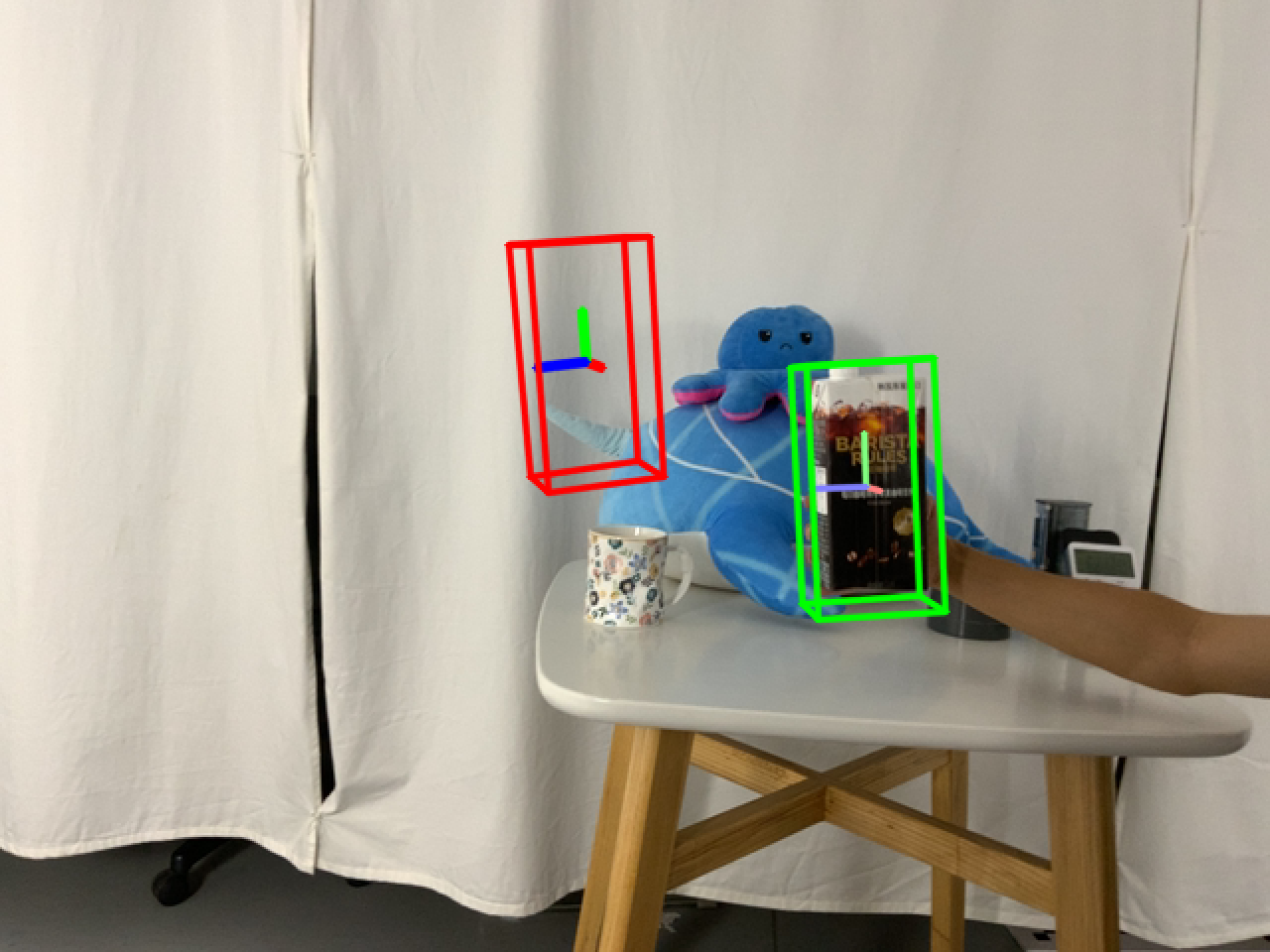}
    \caption{BundleTrack~\cite{wen2021bundletrack} failure case. }
    \label{fig:OPT_Bundletrack}
\end{figure}
\cite{wen2021bundletrack} takes as input an RGB-D sequence along with object segmentation results. It then uses traditional keypoint matching methods to estimate the relative poses between consecutive frames, which are subsequently optimized through an object pose graph. However, as \cref{fig:OPT_Bundletrack} shows, its main drawback is that for objects with sparse texture, the relative pose estimations often become inaccurate, leading to accumulated drift in subsequent frames.

\begin{figure}[thbp]
    \centering
    \includegraphics[width=0.4\textwidth]{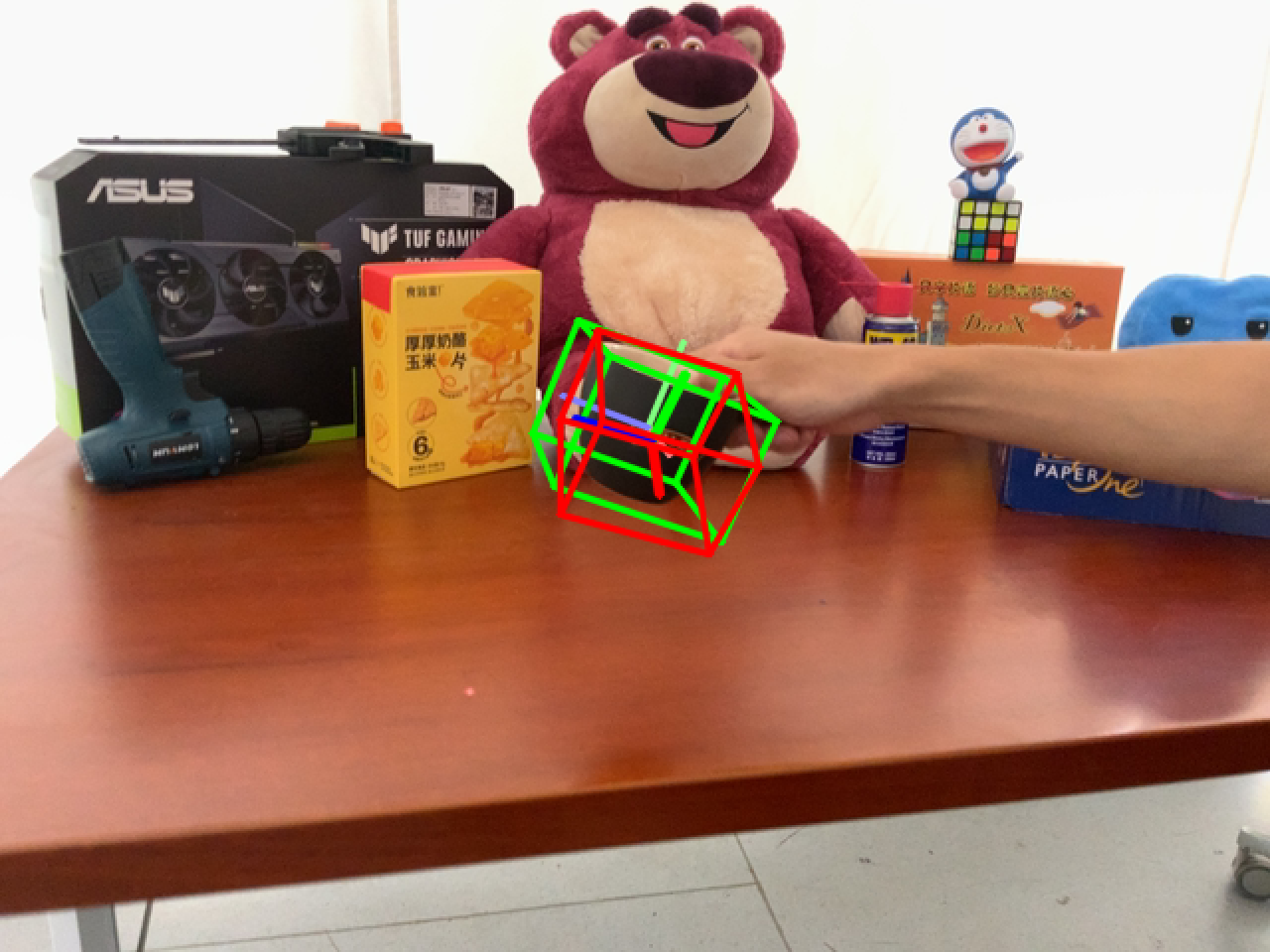}
    \caption{BundleSDF~\cite{wen2023bundlesdf} failure case. }
    \label{fig:OPT_BundleSDF}
\end{figure}

\cref{fig:OPT_BundleSDF}) demonstrates overall robust performance without obvious drift or failure cases. However, it still faces considerable challenges when dealing with textureless and symmetric objects. Moreover, as \cite{wen2023bundlesdf} is a Neural Object Field-based approach, it simultaneously estimates the object’s 6DoF pose and reconstructs the object’s mesh. Consequently, the runtime for processing an entire sequence is considerably limited.

\begin{figure}[thbp]
    \centering
    \includegraphics[width=0.4\textwidth]{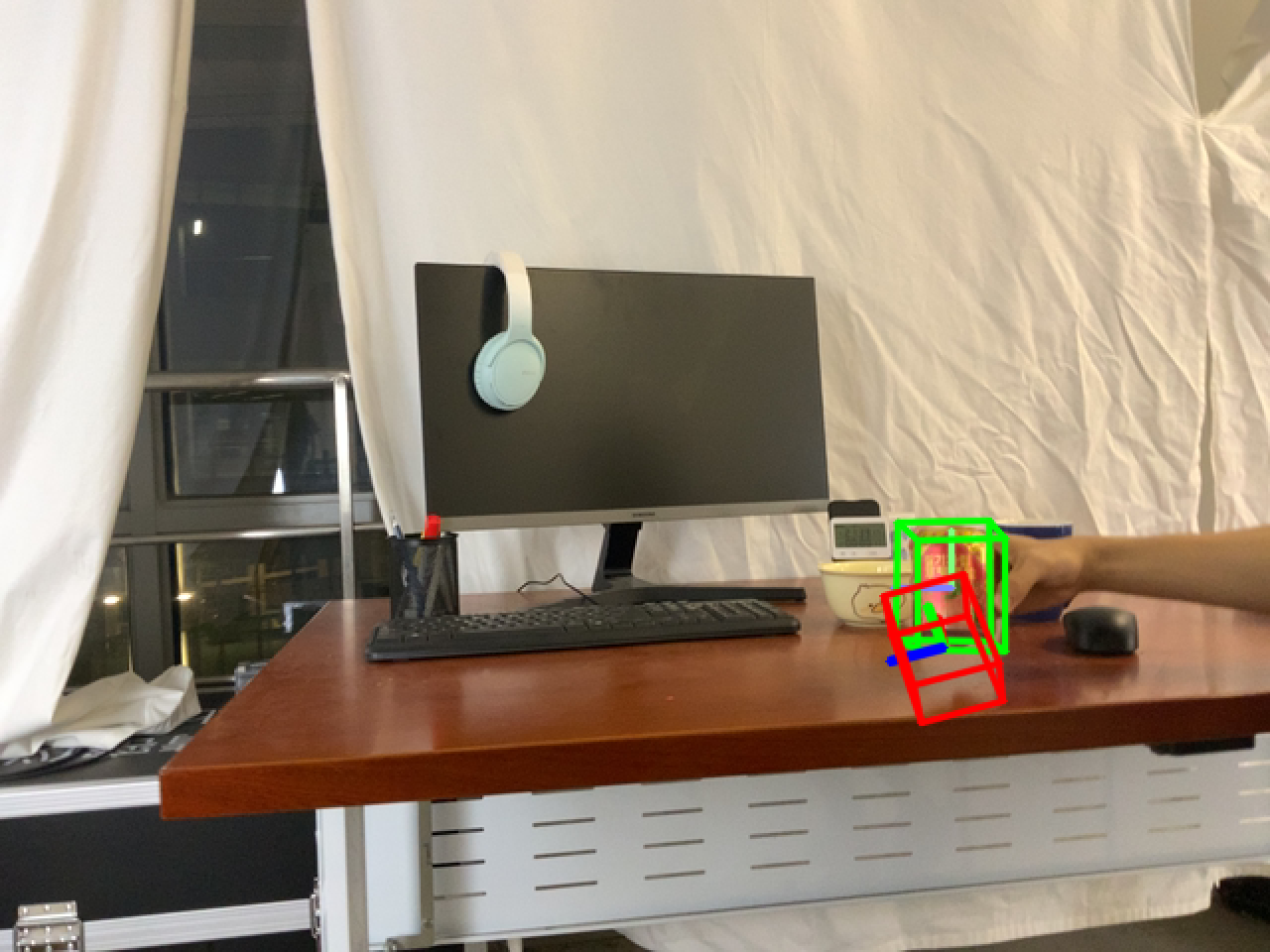}
    \caption{DROID-SLAM~\cite{teed2021droid} failure case.}
    \label{fig:OPT_Droid-SLAM}
\end{figure}

\cite{teed2021droid}, as a popular SLAM framework, takes RGB-D inputs and was originally proposed for camera pose tracking and scene reconstruction. When provided with segmented images and fed into its DBA layer, it operates in an object-centric setting, allowing us to obtain the object pose for each frame. However, since the masked objects provide limited input information, any error in camera pose estimation can lead to significant drift(Fig.\ref{fig:OPT_Droid-SLAM}).

\begin{figure}[htbp]
    \centering
    \includegraphics[width=0.44\textwidth]{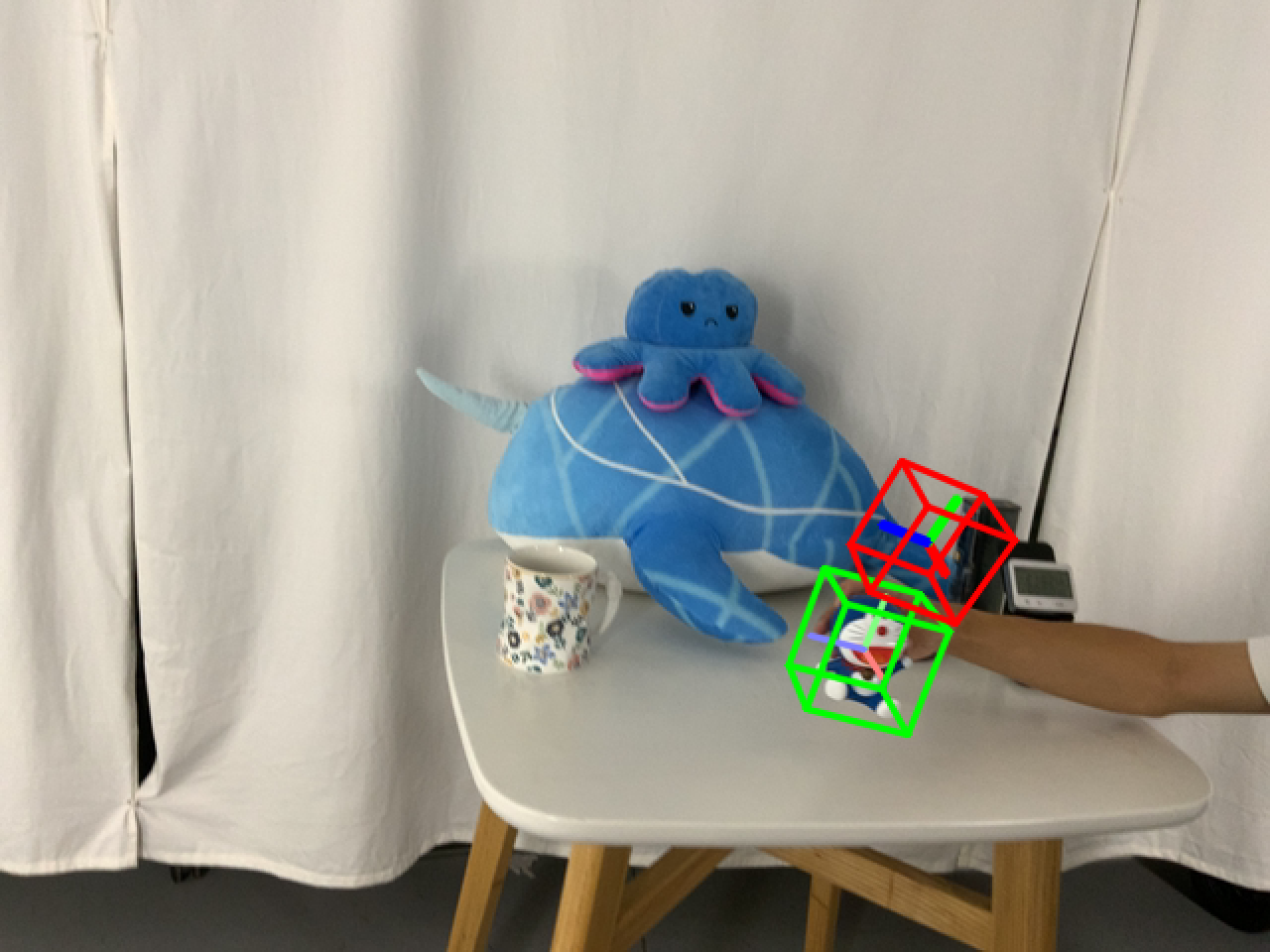}
    \caption{FoundationPose~\cite{wen2024foundationpose} (tracking) failure case.}
    \label{fig:OPT_fdpose}
\end{figure}

\cite{wen2024foundationpose} supports both object pose estimation and object pose tracking. We benchmark its tracking mode. As shown in \cref{fig:OPT_fdpose}, the tracking results in drift failure and tracking loss in rapid motion.

\subsection{Analysis of Category-level Object Estimation Methods}

\begin{figure}[t]
    \centering
        \begin{subfigure}[b]{0.22\textwidth}
            \includegraphics[width=\linewidth]{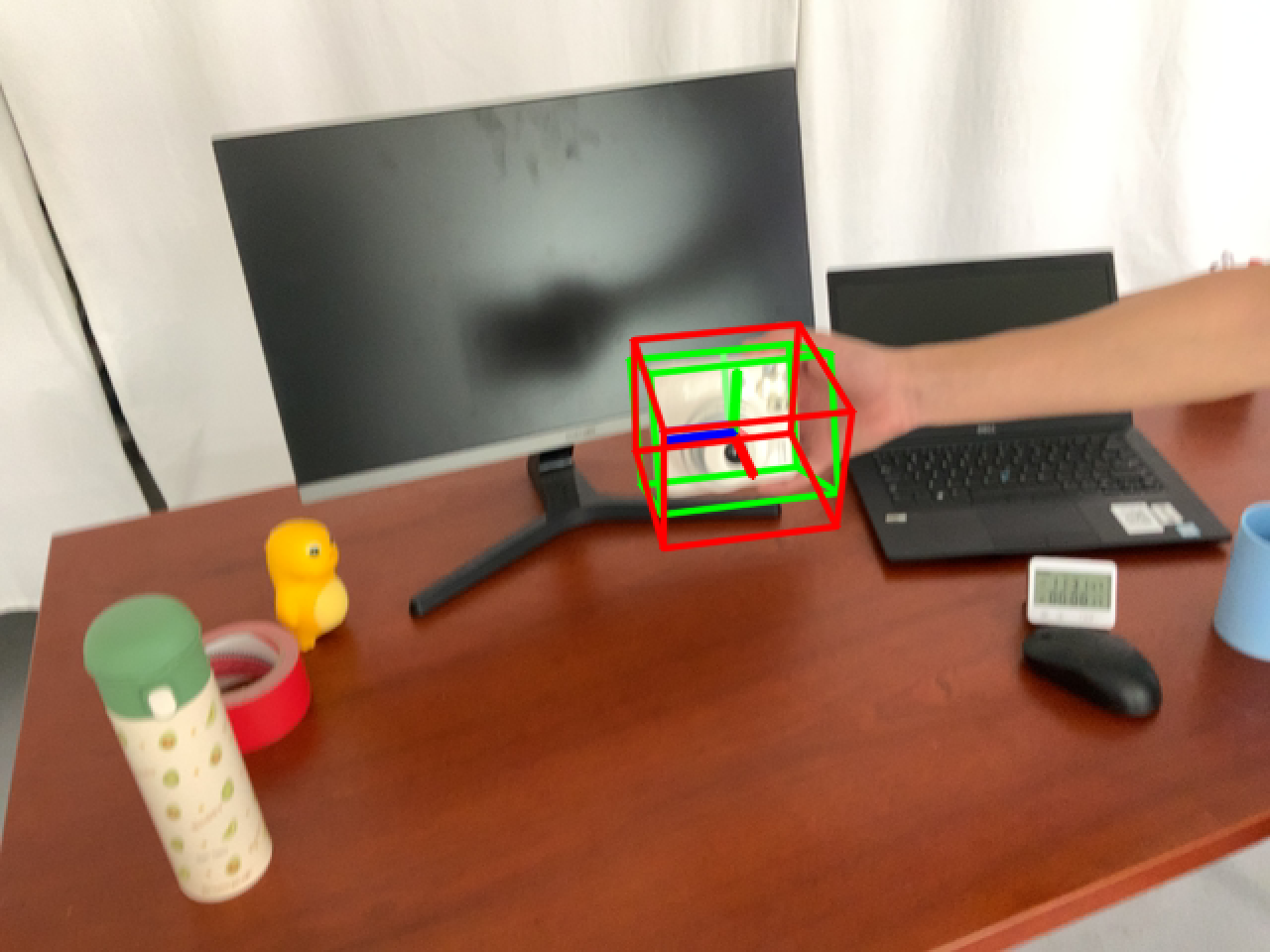}
            \caption{Original AG-Pose~\cite{lin2024instance}.}
            
        \end{subfigure}
        \begin{subfigure}[b]{0.22\textwidth}
            \includegraphics[width=\linewidth]{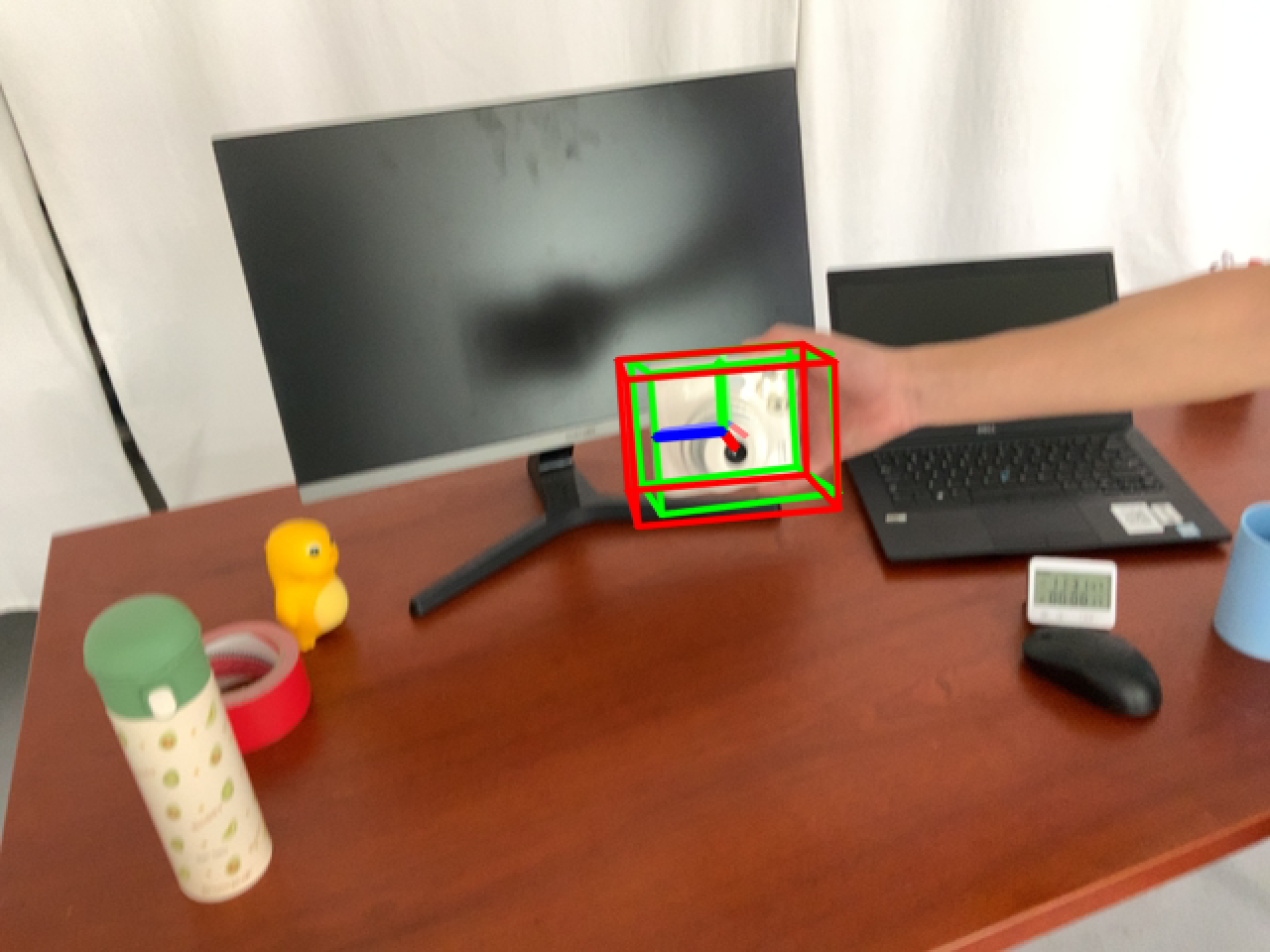}
            \caption{Fine-tuned AG-Pose~\cite{lin2024instance}.}
            
        \end{subfigure}
        \caption{AG-Pose~\cite{lin2024instance}.}
        \label{finetune-agpose}
\end{figure}  

\begin{figure}[t]
    \centering
        \begin{subfigure}[b]{0.22\textwidth}
            \includegraphics[width=\linewidth]{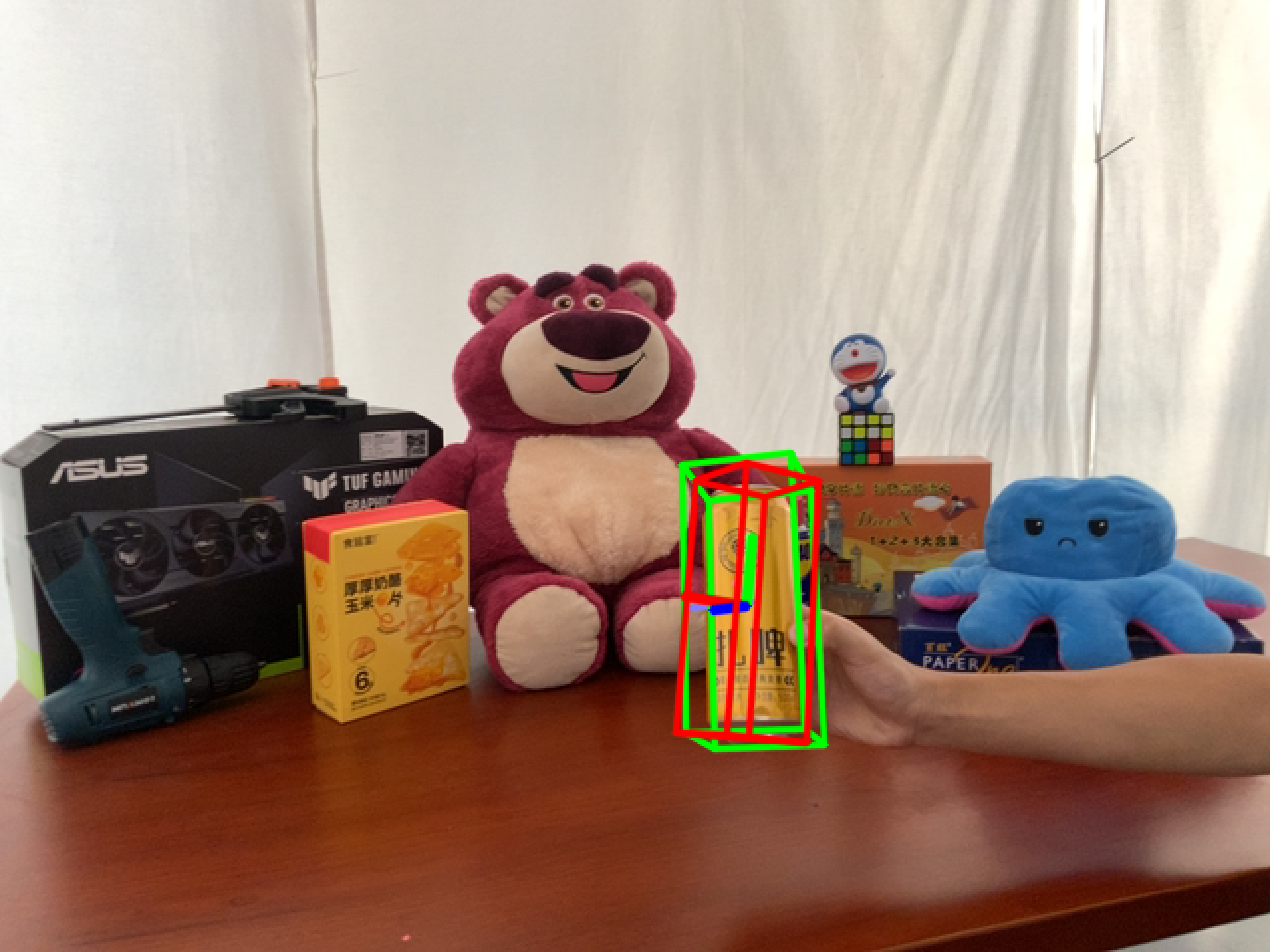}
            \caption{Original SecondPose~\cite{chen2024secondpose}}
            
        \end{subfigure}
        \begin{subfigure}[b]{0.22\textwidth}
            \includegraphics[width=\linewidth]{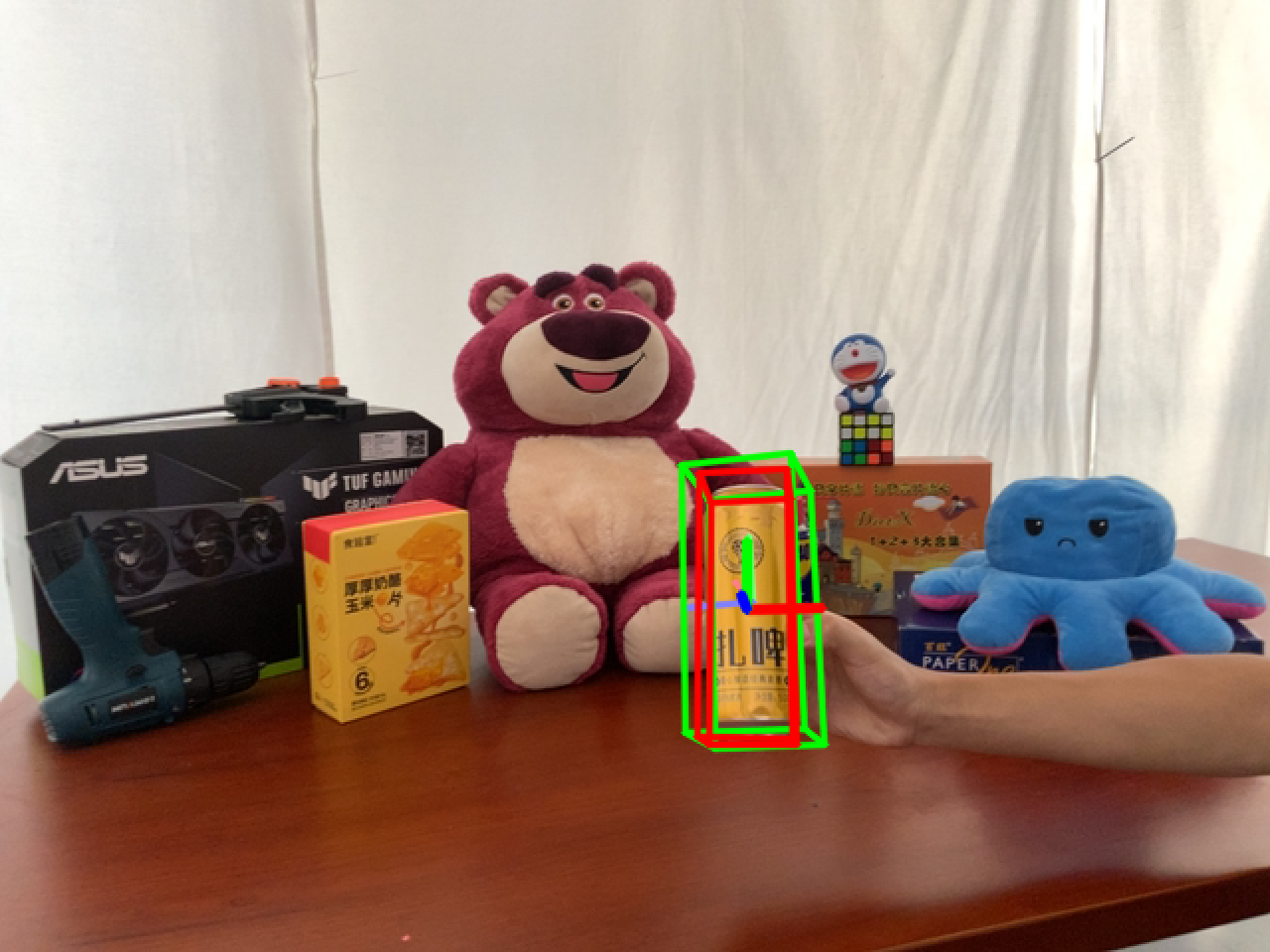}
            \caption{Fine-tuned SecondPose~\cite{chen2024secondpose}.}
        \end{subfigure}
        \caption{SecondPose~\cite{chen2024secondpose}.}
        \label{finetune-secondpose}
\end{figure}

\subsubsection{Fine-tuning}

Moreover, under our \textit{Dynamic Object with Moving Camera scenarios}, the dataset presents different challenges compared to static real-world datasets such as REAL275~\cite{wang2019normalized} or HouseCat6D~\cite{jung2024housecat6d}. Our dataset captures novel viewpoints induced by object rotations along different axes while being held in hand—viewpoints that are difficult to observe in static scenes. These challenges manifest in pose estimation failures for various methods, such as GCASP~\cite{li2023generative} when a bowl is lifted upward, AG-Pose~\cite{lin2024instance} and GenPose~\cite{zhang2023generative} when the camera is moved upward, and DiffusionNOCS~\cite{ikeda2024diffusionnocs} when a can is lifted vertically.
Furthermore, in our fine-tuning experiments (see Section ~\ref{cope_finetuned_results}), it is evident that fine-tuning on our dataset leads to substantial improvements. For example, in Fig.\ref{finetune-agpose} and Fig.\ref{finetune-secondpose}, when comparing SecondPose~\cite{chen2024secondpose} and AG-Pose~\cite{lin2024instance} before and after fine-tuning, the previously mentioned pose estimation failures caused by vertical lifting of objects are significantly mitigated.

\begin{figure*}[tbp]
    \centering
    \begin{subfigure}[b]{0.28\textwidth}
        \includegraphics[width=\linewidth]{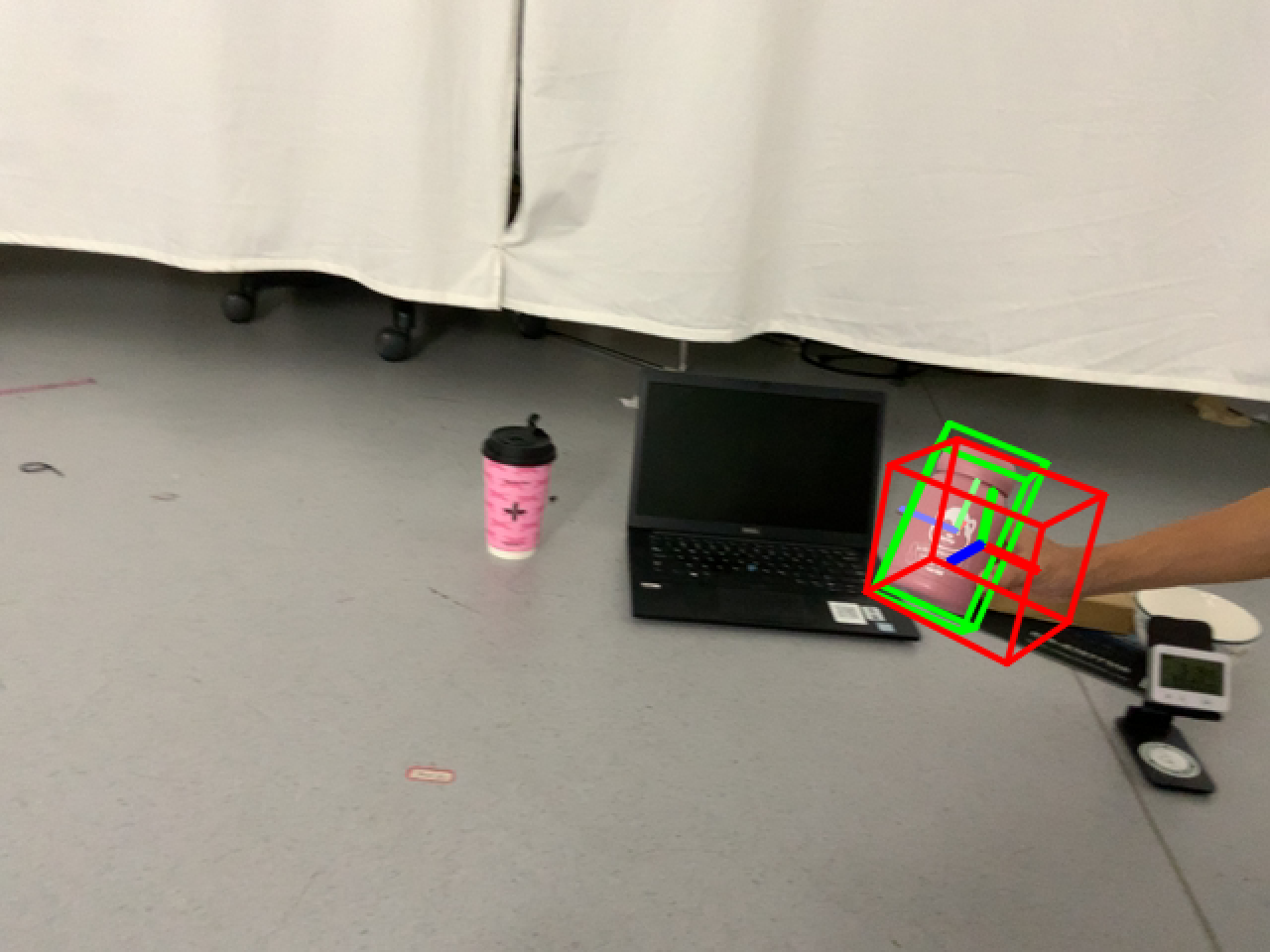}
        \caption{NOCS~\cite{wang2019densefusion}.}
    \end{subfigure}
    \hfill
    \begin{subfigure}[b]{0.28\textwidth}
        \includegraphics[width=\linewidth]{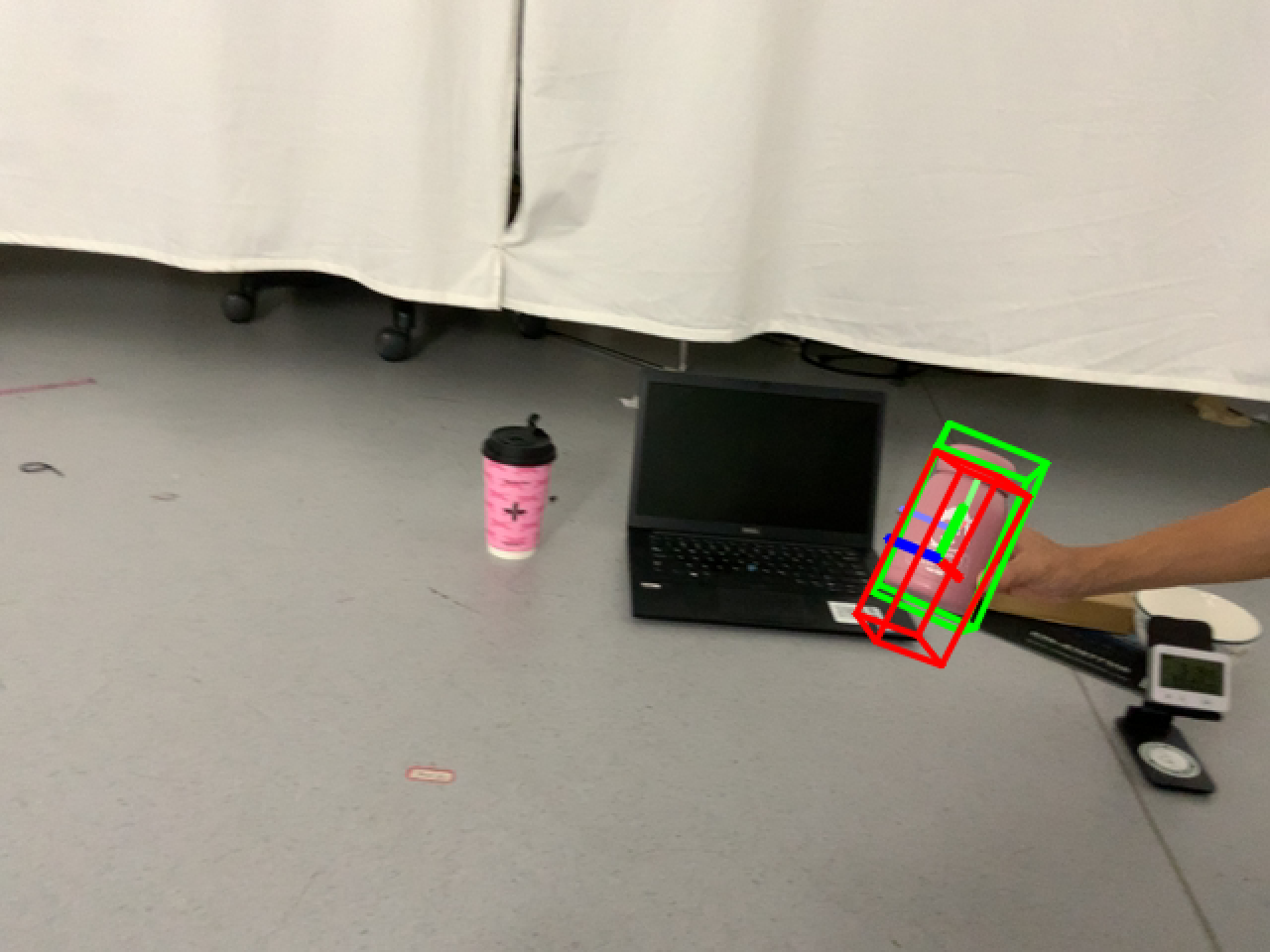}
        \caption{NOCS (MV-ROPE)~\cite{yang2024mv}.}
    \end{subfigure}
    \hfill
    \begin{subfigure}[b]{0.28\textwidth}
        \includegraphics[width=\linewidth]{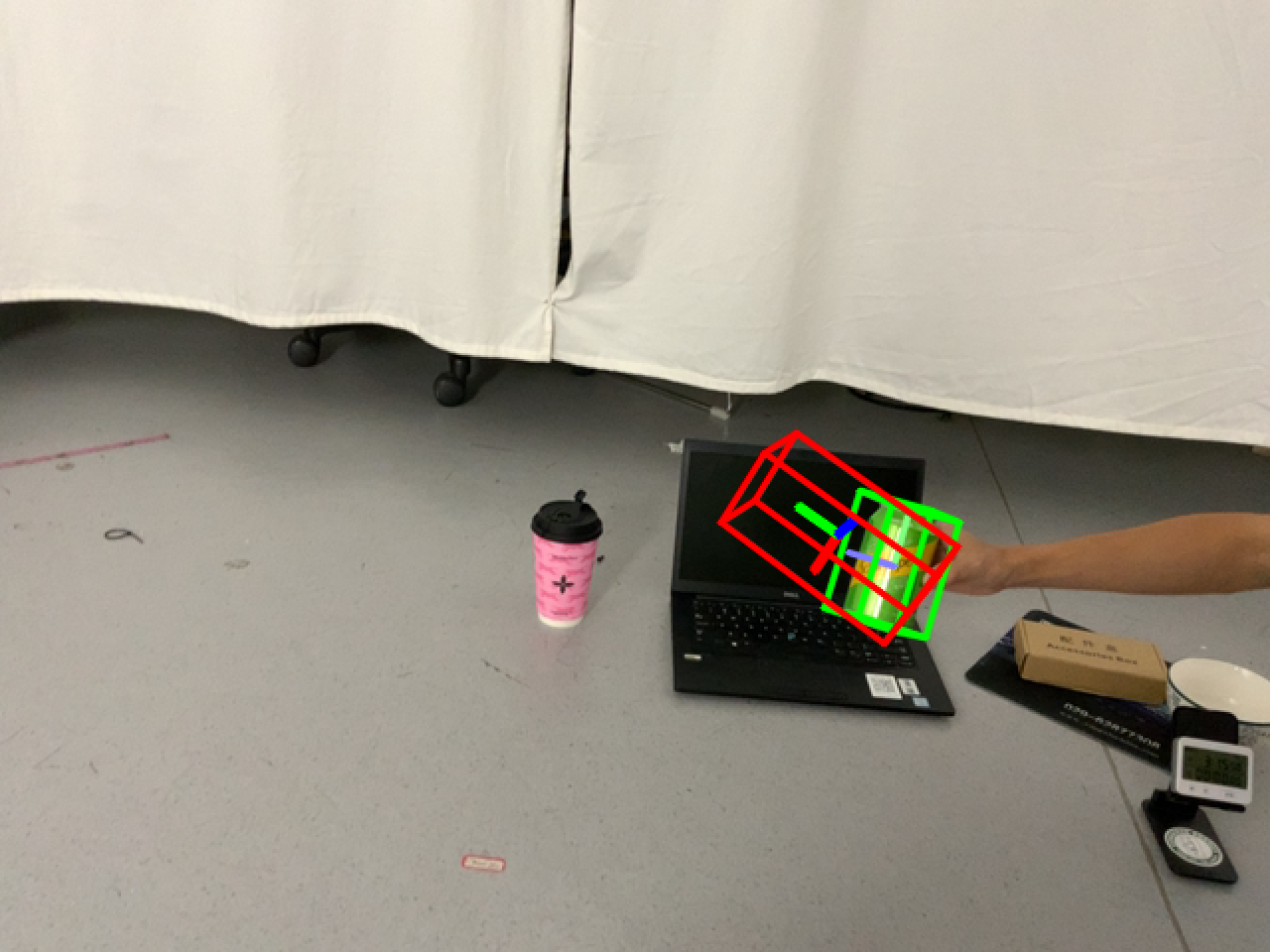}
        \caption{DiffusionNOCS~\cite{ikeda2024diffusionnocs}.}
    \end{subfigure}

    \vspace{0.5em}

    \begin{subfigure}[b]{0.28\textwidth}
        \includegraphics[width=\linewidth]{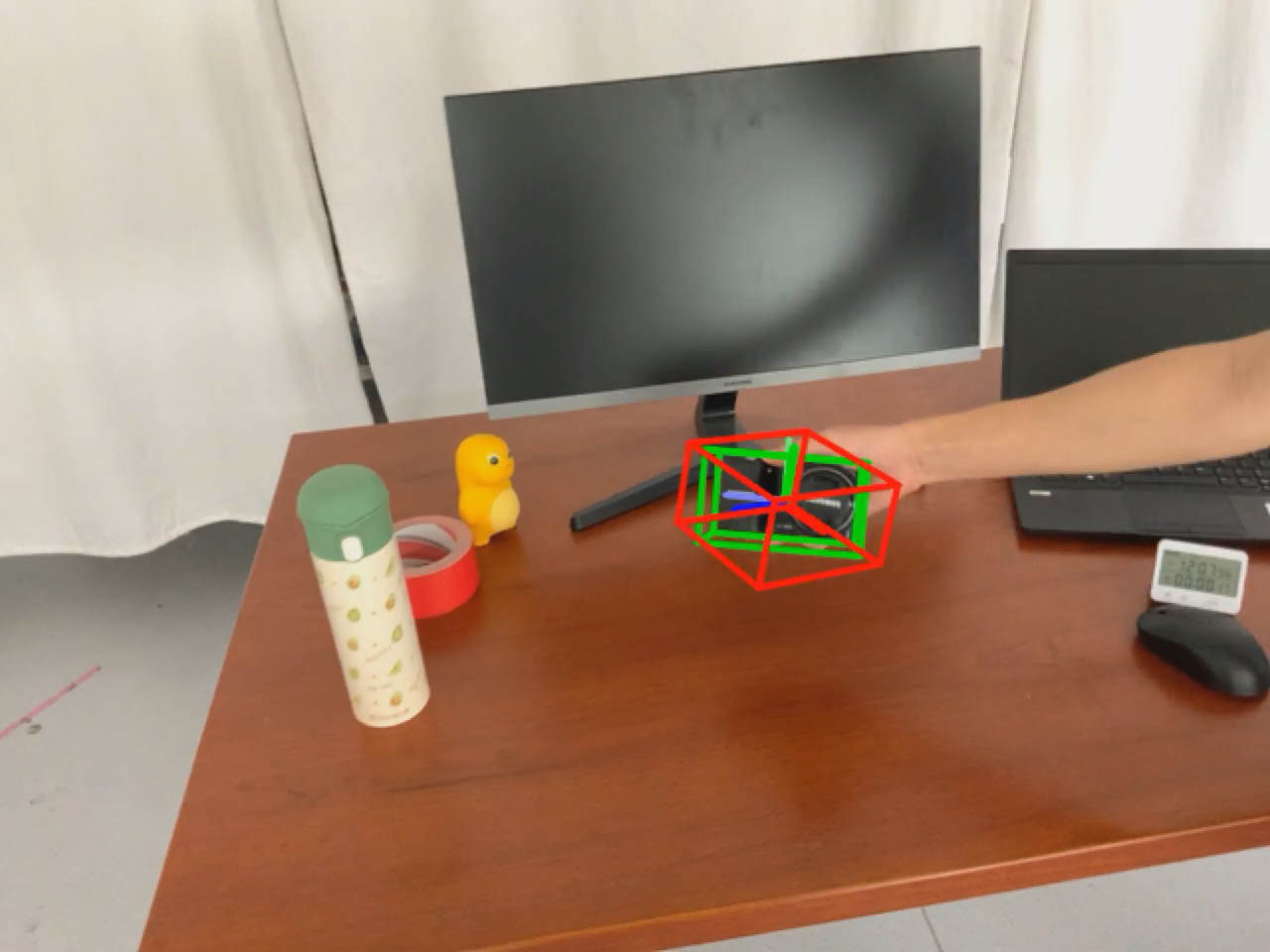}
        \caption{AG-Pose~\cite{lin2024instance}.}
    \end{subfigure}
    \hfill
    \begin{subfigure}[b]{0.28\textwidth}
        \includegraphics[width=\linewidth]{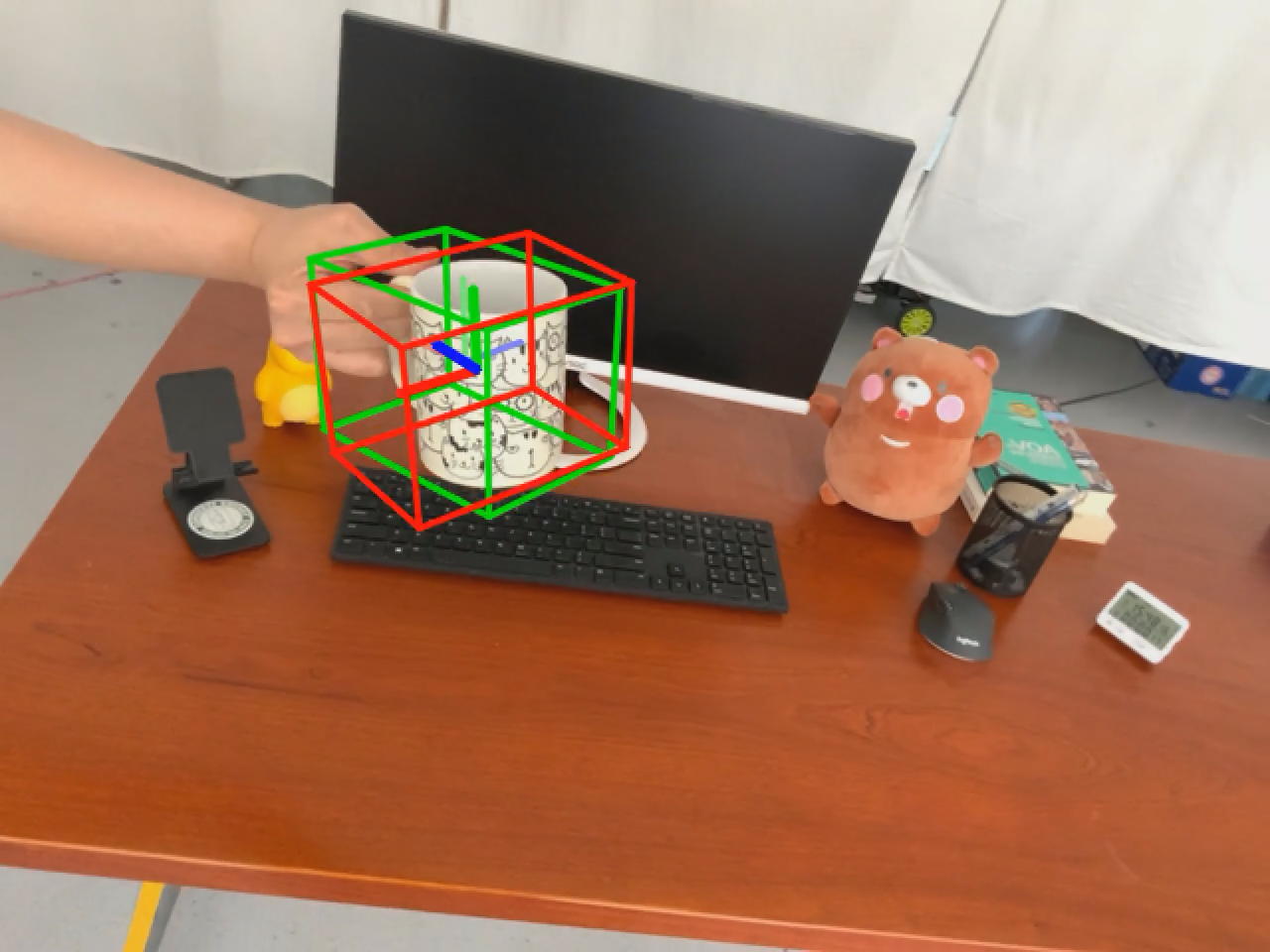}
        \caption{SecondPose~\cite{chen2024secondpose}.}
    \end{subfigure}
    \hfill
    \begin{subfigure}[b]{0.28\textwidth}
        \includegraphics[width=\linewidth]{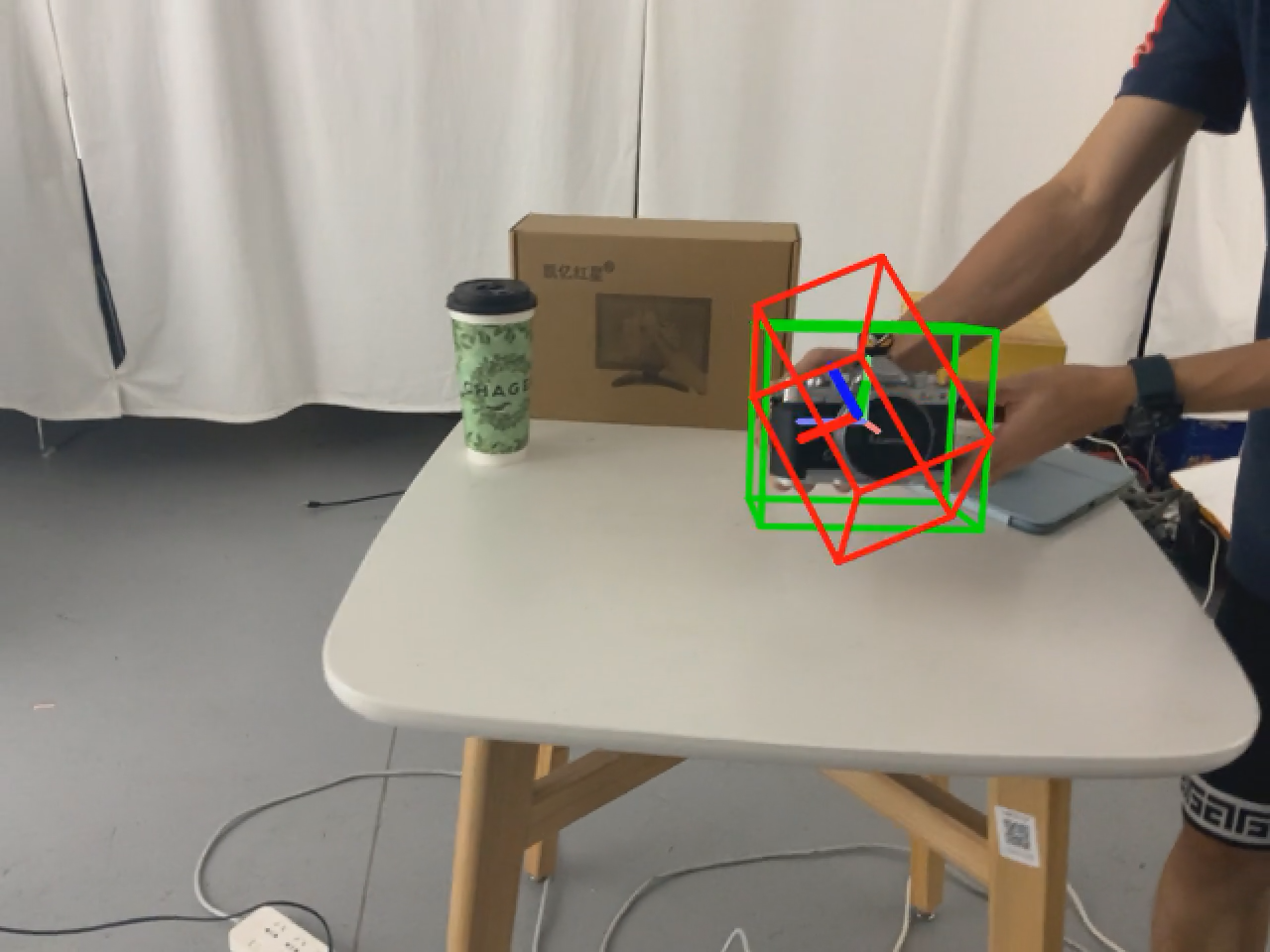}
        \caption{IST-Net~\cite{liu2023net}}
    \end{subfigure}

    \vspace{0.5em} 

    \begin{subfigure}[b]{0.28\textwidth}
        \includegraphics[width=\linewidth]{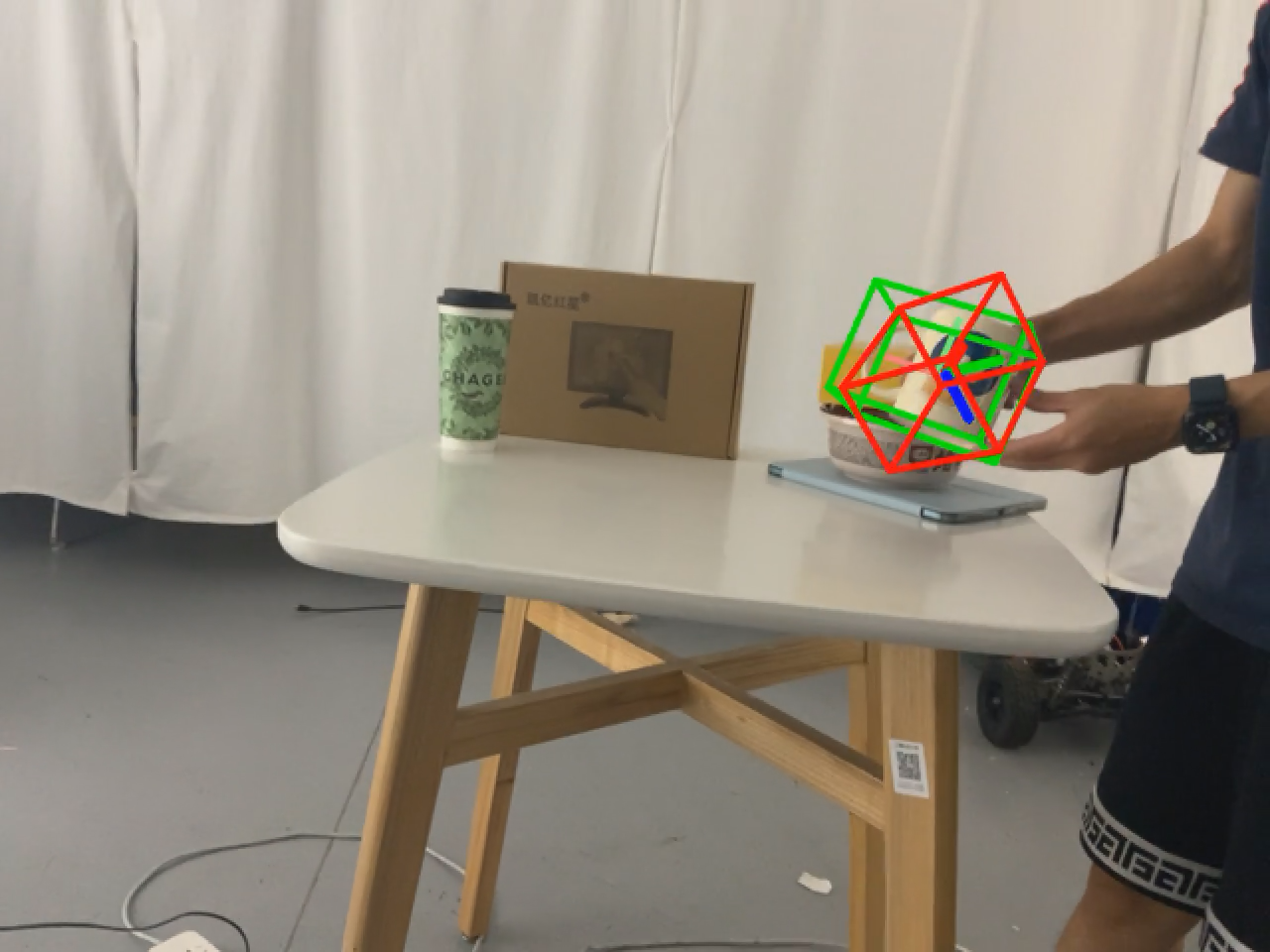}
        \caption{VI-Net~\cite{lin2023vi}.}
    \end{subfigure}
    \hfill
    \begin{subfigure}[b]{0.28\textwidth}
        \includegraphics[width=\linewidth]{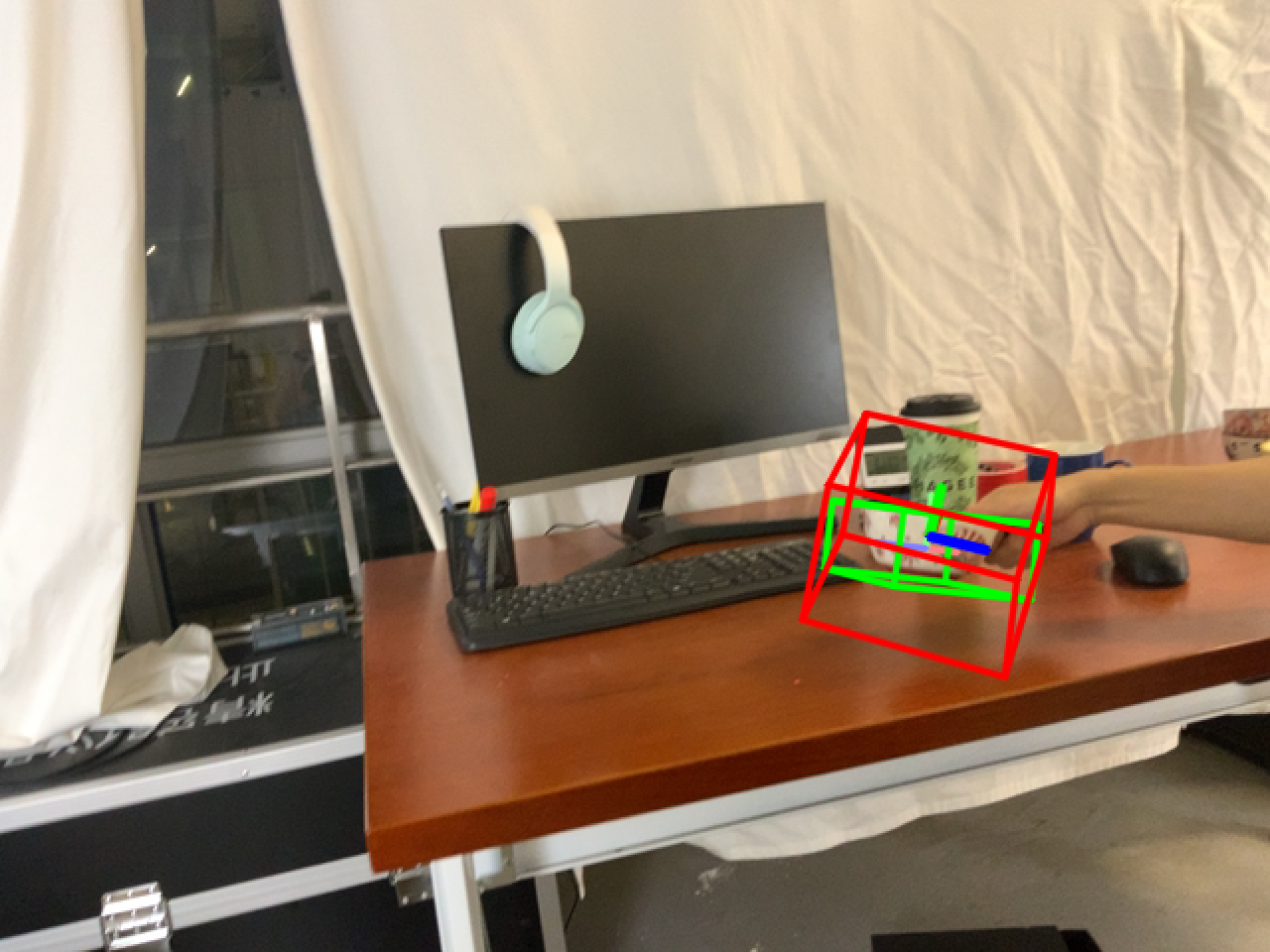}
        \caption{GCASP~\cite{li2023generative}.}
    \end{subfigure}
    \hfill
    \begin{subfigure}[b]{0.28\textwidth}
        \includegraphics[width=\linewidth]{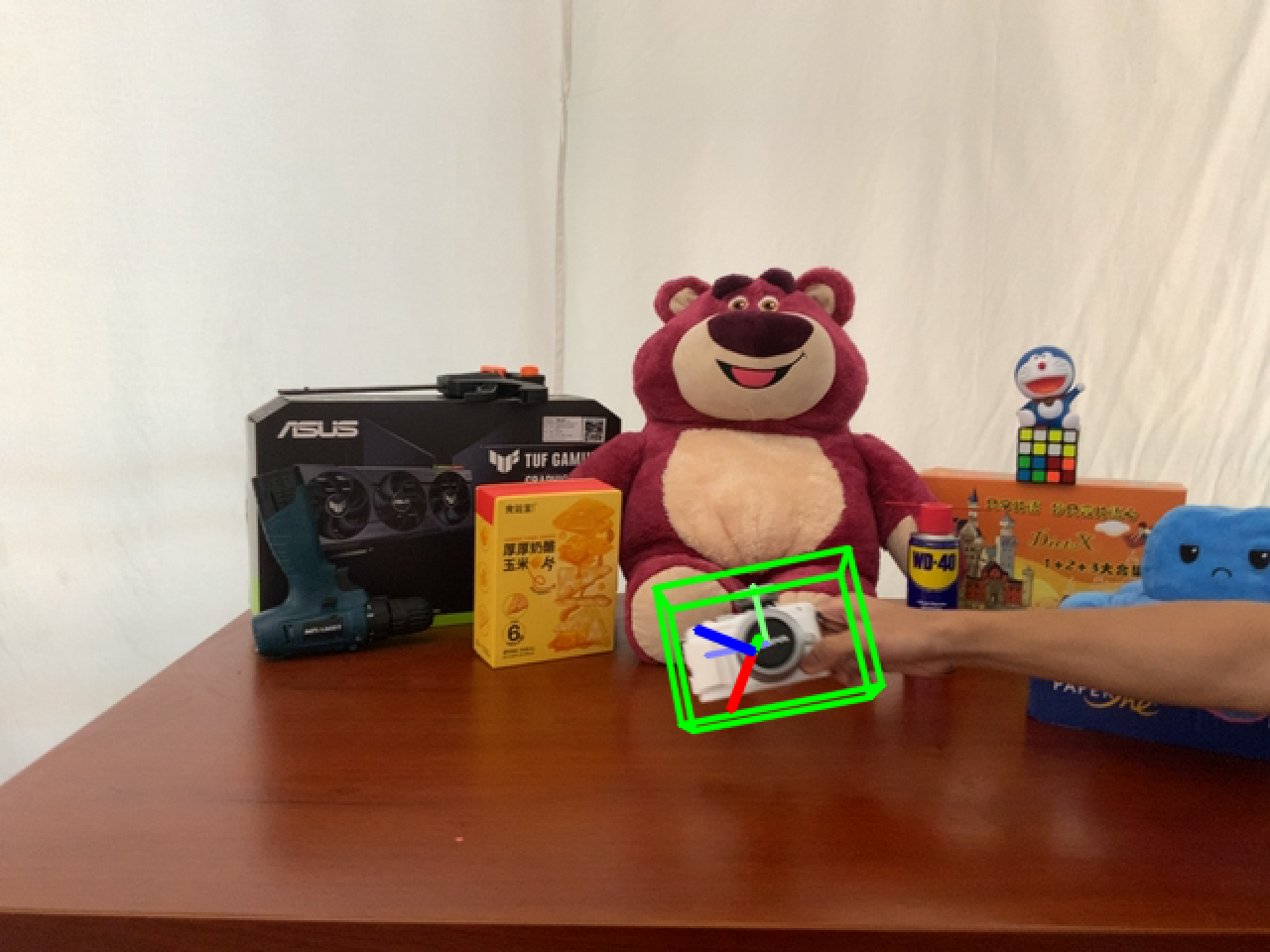}
        \caption{GenPose~\cite{zhang2023generative}.}
    \end{subfigure}

    \caption{COPE methods failure cases.}
    \label{fig:COPE}
\end{figure*}

\subsection{COPE methods failure cases.}
\cite{wang2019normalized} predicts normalized object points and then apply the Umeyama algorithm~\cite{umeyama1991least} to recover the object pose and size. However \cite{umeyama1991least} requires high-quality NOCS map and depth map to recover accurate object pose and size, otherwise, its accuracy is limited. Note that in the original NOCS implementation, its object detection and NOCS map prediction module is coupled. For fairness in benchmarking, we adopt the settings from MV-ROPE~\cite{yang2024mv}, where consistent segmentation results similar to other benchmarked methods are provided as input.

Overall, the methods that rely on predicting a NOCS map, such as \cite{wang2019normalized, ikeda2024diffusionnocs, yang2024mv} tend to exhibit significantly higher translation errors compared to methods that directly predict translation, such as \cite{liu2023net,li2023generative, lin2023vi, chen2024secondpose,lin2024instance,zhang2023generative}. This performance gap can be attributed to the fact that most of the latter methods, including \cite{liu2023net, lin2023vi, chen2024secondpose,lin2024instance}, perform translation prediction using PointNet-based architectures\cite{qi2017pointnet} and decouple it from rotation prediction. Compared to rotation estimation, translation prediction techniques are relatively more mature and stable.

Regarding COPE algorithms, we found that challenging cases tend to concentrate in two categories: the camera category and the mug category, so we focused our analysis there.

\begin{enumerate}
    \item The camera category has long been a known challenge in COPE tasks due to the large intra-class variation among commonly seen camera models. Cameras in real-world scenarios exhibit a wide range of appearances, posing significant challenges for the generalization ability of pose estimation networks.
    \item The mug category, by contrast, tends to have more consistent geometric shapes. However, failure cases often occur when a mug is being held by a human hand. Such occlusion frequently leads to errors in rotation estimation. Another possible cause of failure is when the mug’s opening is not visible in the input image, under such conditions, it becomes difficult for the network to distinguish the upright orientation of the mug.
\end{enumerate}

\begin{figure}[tbp]
    \centering
        \begin{subfigure}[b]{0.235\textwidth}
            \includegraphics[width=\linewidth]{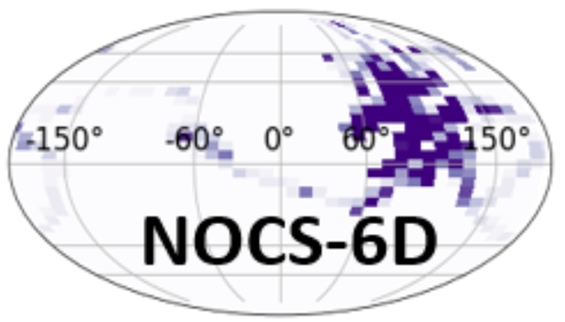}
            \caption{}
            
        \end{subfigure}
        \begin{subfigure}[b]{0.24\textwidth}
            \includegraphics[width=\linewidth]{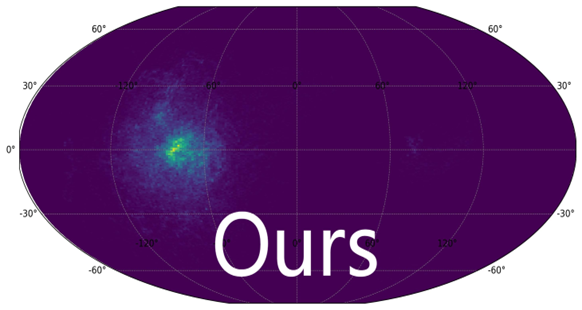}
            \caption{}
            
        \end{subfigure}
        \caption{Viewpoint coverage map. Most of our benchmarked COPE algorithm is trained on NOCS-6D~\cite{wang2019densefusion} dataset. Differ from NOCS-6D~\cite{wang2019normalized}, our viewpoint coverage is more balanced between top-down and bottom-up views. Note that the left figure is referred to \cite{jung2024housecat6d}.}
        \label{fig:nocs6d}
\end{figure}  

Another issue is that most benchmarked COPE methods are trained in NOCS~\cite{wang2019normalized} dataset, whose pose distribution is different from ours, which can be another reason for the failure cases in our test split.

\clearpage
{
\printbibliography[heading=bibintoc,title={References}]

@inproceedings{guo2023handal,
  title={Handal: A dataset of real-world manipulable object categories with pose annotations, affordances, and reconstructions},
  author={Guo, Andrew and Wen, Bowen and Yuan, Jianhe and Tremblay, Jonathan and Tyree, Stephen and Smith, Jeffrey and Birchfield, Stan},
  booktitle={2023 IEEE/RSJ International Conference on Intelligent Robots and Systems (IROS)},
  pages={11428--11435},
  year={2023},
  organization={IEEE}
}

@article{fu2022category,
  title={Category-level 6d object pose estimation in the wild: A semi-supervised learning approach and a new dataset},
  author={Fu, Yang and Wang, Xiaolong},
  journal={Advances in Neural Information Processing Systems},
  volume={35},
  pages={27469--27483},
  year={2022}
}

@inproceedings{wang2022phocal,
  title={Phocal: A multi-modal dataset for category-level object pose estimation with photometrically challenging objects},
  author={Wang, Pengyuan and Jung, HyunJun and Li, Yitong and Shen, Siyuan and Srikanth, Rahul Parthasarathy and Garattoni, Lorenzo and Meier, Sven and Navab, Nassir and Busam, Benjamin},
  booktitle={Proceedings of the IEEE/CVF conference on computer vision and pattern recognition},
  pages={21222--21231},
  year={2022}
}

@inproceedings{wang2019normalized,
  title={Normalized object coordinate space for category-level 6d object pose and size estimation},
  author={Wang, He and Sridhar, Srinath and Huang, Jingwei and Valentin, Julien and Song, Shuran and Guibas, Leonidas J},
  booktitle={Proceedings of the IEEE/CVF Conference on Computer Vision and Pattern Recognition},
  pages={2642--2651},
  year={2019}
}

@inproceedings{liu2021stereobj,
  title={Stereobj-1m: Large-scale stereo image dataset for 6d object pose estimation},
  author={Liu, Xingyu and Iwase, Shun and Kitani, Kris M},
  booktitle={Proceedings of the IEEE/CVF International Conference on Computer Vision},
  pages={10870--10879},
  year={2021}
}

@article{hodan2023bop,
  title={{BOP} Challenge 2023 on Detection, Segmentation and Pose Estimation of Seen and Unseen Rigid Objects},
  author={Hoda{\v{n}}, Tom{\'a}{\v{s}} and Sundermeyer, Martin and Labb{\'e}, Yann and Nguyen, Van Nguyen and Wang, Gu and Brachmann, Eric and Drost, Bertram and Lepetit, Vincent and Rother, Carsten and Matas, Ji{\v{r}}{\'i}},
  journal={Conference on Computer Vision and Pattern Recognition Workshops (CVPRW)},
  year={2024}
}

@inproceedings{wen2020se,
  title={se (3)-tracknet: Data-driven 6d pose tracking by calibrating image residuals in synthetic domains},
  author={Wen, Bowen and Mitash, Chaitanya and Ren, Baozhang and Bekris, Kostas E},
  booktitle={2020 IEEE/RSJ International Conference on Intelligent Robots and Systems (IROS)},
  pages={10367--10373},
  year={2020},
  organization={IEEE}
}

@inproceedings{hinterstoisser2013model,
  title={Model based training, detection and pose estimation of texture-less 3d objects in heavily cluttered scenes},
  author={Hinterstoisser, Stefan and Lepetit, Vincent and Ilic, Slobodan and Holzer, Stefan and Bradski, Gary and Konolige, Kurt and Navab, Nassir},
  booktitle={Computer Vision--ACCV 2012: 11th Asian Conference on Computer Vision, Daejeon, Korea, November 5-9, 2012, Revised Selected Papers, Part I 11},
  pages={548--562},
  year={2013},
  organization={Springer}
}

@inproceedings{hodan2017t,
  title={T-LESS: An RGB-D dataset for 6D pose estimation of texture-less objects},
  author={Hodan, Tom{\'a}{\v{s}} and Haluza, Pavel and Obdr{\v{z}}{\'a}lek, {\v{S}}tep{\'a}n and Matas, Jiri and Lourakis, Manolis and Zabulis, Xenophon},
  booktitle={2017 IEEE Winter Conference on Applications of Computer Vision (WACV)},
  pages={880--888},
  year={2017},
  organization={IEEE}
}

@inproceedings{drost2017introducing,
  title={Introducing mvtec itodd-a dataset for 3d object recognition in industry},
  author={Drost, Bertram and Ulrich, Markus and Bergmann, Paul and Hartinger, Philipp and Steger, Carsten},
  booktitle={Proceedings of the IEEE international conference on computer vision workshops},
  pages={2200--2208},
  year={2017}
}

@inproceedings{jung2024housecat6d,
  title={HouseCat6D-A Large-Scale Multi-Modal Category Level 6D Object Perception Dataset with Household Objects in Realistic Scenarios},
  author={Jung, HyunJun and Wu, Shun-Cheng and Ruhkamp, Patrick and Zhai, Guangyao and Schieber, Hannah and Rizzoli, Giulia and Wang, Pengyuan and Zhao, Hongcheng and Garattoni, Lorenzo and Meier, Sven and others},
  booktitle={Proceedings of the IEEE/CVF Conference on Computer Vision and Pattern Recognition},
  pages={22498--22508},
  year={2024}
}

@inproceedings{wen2024foundationpose,
  title={Foundationpose: Unified 6d pose estimation and tracking of novel objects},
  author={Wen, Bowen and Yang, Wei and Kautz, Jan and Birchfield, Stan},
  booktitle={Proceedings of the IEEE/CVF Conference on Computer Vision and Pattern Recognition},
  pages={17868--17879},
  year={2024}
}

@article{ravi2024sam,
  title={Sam 2: Segment anything in images and videos},
  author={Ravi, Nikhila and Gabeur, Valentin and Hu, Yuan-Ting and Hu, Ronghang and Ryali, Chaitanya and Ma, Tengyu and Khedr, Haitham and R{\"a}dle, Roman and Rolland, Chloe and Gustafson, Laura and others},
  journal={arXiv preprint arXiv:2408.00714},
  year={2024}
}

@inproceedings{wen2021bundletrack,
  title={Bundletrack: 6d pose tracking for novel objects without instance or category-level 3d models},
  author={Wen, Bowen and Bekris, Kostas},
  booktitle={2021 IEEE/RSJ International Conference on Intelligent Robots and Systems (IROS)},
  pages={8067--8074},
  year={2021},
  organization={IEEE}
}

@misc{structuresensor,
  author       = {Occipital, Inc.},
  title        = {{Structure Sensor}},
  howpublished = {\url{https://structure.io/}},
  note         = {Accessed: 2023-10-05},
}

@article{karaev2024cotracker3,
  title={CoTracker3: Simpler and better point tracking by pseudo-labelling real videos},
  author={Karaev, Nikita and Makarov, Iurii and Wang, Jianyuan and Neverova, Natalia and Vedaldi, Andrea and Rupprecht, Christian},
  journal={arXiv preprint arXiv:2410.11831},
  year={2024}
}

@inproceedings{ikeda2024diffusionnocs,
  title={Diffusionnocs: Managing symmetry and uncertainty in sim2real multi-modal category-level pose estimation},
  author={Ikeda, Takuya and Zakharov, Sergey and Ko, Tianyi and Irshad, Muhammad Zubair and Lee, Robert and Liu, Katherine and Ambrus, Rares and Nishiwaki, Koichi},
  booktitle={2024 IEEE/RSJ International Conference on Intelligent Robots and Systems (IROS)},
  pages={7406--7413},
  year={2024},
  organization={IEEE}
}

@inproceedings{yang2024mv,
  title={MV-ROPE: Multi-view Constraints for Robust Category-level Object Pose and Size Estimation},
  author={Yang, Jiaqi and Chen, Yucong and Meng, Xiangting and Yan, Chenxin and Li, Min and Cheng, Ran and Liu, Lige and Sun, Tao and Kneip, Laurent},
  booktitle={2024 IEEE/RSJ International Conference on Intelligent Robots and Systems (IROS)},
  pages={7588--7595},
  year={2024},
  organization={IEEE}
}

@inproceedings{krishnan2024omninocs,
  title={OmniNOCS: A unified NOCS dataset and model for 3D lifting of 2D objects},
  author={Krishnan, Akshay and Kundu, Abhijit and Maninis, Kevis-Kokitsi and Hays, James and Brown, Matthew},
  booktitle={European Conference on Computer Vision},
  pages={127--145},
  year={2024},
  organization={Springer}
}

@article{liu2024deep,
  title={Deep learning-based object pose estimation: A comprehensive survey},
  author={Liu, Jian and Sun, Wei and Yang, Hui and Zeng, Zhiwen and Liu, Chongpei and Zheng, Jin and Liu, Xingyu and Rahmani, Hossein and Sebe, Nicu and Mian, Ajmal},
  journal={arXiv preprint arXiv:2405.07801},
  year={2024}
}

@inproceedings{meng2023kgnet,
  title={Kgnet: Knowledge-guided networks for category-level 6d object pose and size estimation},
  author={Meng, Qiwei and Gu, Jason and Zhu, Shiqiang and Liao, Jianfeng and Jin, Tianlei and Guo, Fangtai and Wang, Wen and Song, Wei},
  booktitle={2023 IEEE International Conference on Robotics and Automation (ICRA)},
  pages={6102--6108},
  year={2023},
  organization={IEEE}
}

@inproceedings{wang2023query6dof,
  title={Query6dof: Learning sparse queries as implicit shape prior for category-level 6dof pose estimation},
  author={Wang, Ruiqi and Wang, Xinggang and Li, Te and Yang, Rong and Wan, Minhong and Liu, Wenyu},
  booktitle={Proceedings of the IEEE/CVF international conference on computer vision},
  pages={14055--14064},
  year={2023}
}

@inproceedings{li2023generative,
  title={Generative category-level shape and pose estimation with semantic primitives},
  author={Li, Guanglin and Li, Yifeng and Ye, Zhichao and Zhang, Qihang and Kong, Tao and Cui, Zhaopeng and Zhang, Guofeng},
  booktitle={Conference on Robot Learning},
  pages={1390--1400},
  year={2023},
  organization={PMLR}
}

@inproceedings{lin2023vi,
  title={Vi-net: Boosting category-level 6d object pose estimation via learning decoupled rotations on the spherical representations},
  author={Lin, Jiehong and Wei, Zewei and Zhang, Yabin and Jia, Kui},
  booktitle={Proceedings of the IEEE/CVF international conference on computer vision},
  pages={14001--14011},
  year={2023}
}

@inproceedings{liu2023net,
  title={Ist-net: Prior-free category-level pose estimation with implicit space transformation},
  author={Liu, Jianhui and Chen, Yukang and Ye, Xiaoqing and Qi, Xiaojuan},
  booktitle={Proceedings of the IEEE/CVF International Conference on Computer Vision},
  pages={13978--13988},
  year={2023}
}

@article{zhang2023generative,
  title={Generative category-level object pose estimation via diffusion models},
  author={Zhang, Jiyao and Wu, Mingdong and Dong, Hao},
  journal={Advances in Neural Information Processing Systems},
  volume={36},
  pages={54627--54644},
  year={2023}
}

@inproceedings{chen2024secondpose,
  title={Secondpose: Se (3)-consistent dual-stream feature fusion for category-level pose estimation},
  author={Chen, Yamei and Di, Yan and Zhai, Guangyao and Manhardt, Fabian and Zhang, Chenyangguang and Zhang, Ruida and Tombari, Federico and Navab, Nassir and Busam, Benjamin},
  booktitle={Proceedings of the IEEE/CVF Conference on Computer Vision and Pattern Recognition},
  pages={9959--9969},
  year={2024}
}

@inproceedings{lin2024instance,
  title={Instance-adaptive and geometric-aware keypoint learning for category-level 6d object pose estimation},
  author={Lin, Xiao and Yang, Wenfei and Gao, Yuan and Zhang, Tianzhu},
  booktitle={Proceedings of the IEEE/CVF Conference on Computer Vision and Pattern Recognition},
  pages={21040--21049},
  year={2024}
}

@inproceedings{ornek2024foundpose,
  title={Foundpose: Unseen object pose estimation with foundation features},
  author={{\"O}rnek, Evin P{\i}nar and Labb{\'e}, Yann and Tekin, Bugra and Ma, Lingni and Keskin, Cem and Forster, Christian and Hodan, Tomas},
  booktitle={European Conference on Computer Vision},
  pages={163--182},
  year={2024},
  organization={Springer}
}

@inproceedings{labbe2022megapose,
    title     = {MegaPose: 6D Pose Estimation of Novel Objects via Render \& Compare},
    author    = {Labb\'e, Yann and Manuelli, Lucas and Mousavian, Arsalan and Tyree, Stephen and Birchfield, Stan and Tremblay, Jonathan and Carpentier, Justin and Aubry, Mathieu and Fox, Dieter and Sivic, Josef},
    booktitle = {Proceedings of the 6th Conference on Robot Learning (CoRL)},
    year      = {2022},
  }

@inproceedings{nguyen2024gigapose,
  title={Gigapose: Fast and robust novel object pose estimation via one correspondence},
  author={Nguyen, Van Nguyen and Groueix, Thibault and Salzmann, Mathieu and Lepetit, Vincent},
  booktitle={Proceedings of the IEEE/CVF Conference on Computer Vision and Pattern Recognition},
  pages={9903--9913},
  year={2024}
}

@inproceedings{lin2024sam,
  title={Sam-6d: Segment anything model meets zero-shot 6d object pose estimation},
  author={Lin, Jiehong and Liu, Lihua and Lu, Dekun and Jia, Kui},
  booktitle={Proceedings of the IEEE/CVF Conference on Computer Vision and Pattern Recognition},
  pages={27906--27916},
  year={2024}
}

@article{li2025gce,
  title={GCE-Pose: Global Context Enhancement for Category-level Object Pose Estimation},
  author={Li, Weihang and Xu, Hongli and Huang, Junwen and Jung, Hyunjun and Yu, Peter KT and Navab, Nassir and Busam, Benjamin},
  journal={arXiv preprint arXiv:2502.04293},
  year={2025}
}

@inproceedings{wang2024gs,
  title={Gs-pose: Category-level object pose estimation via geometric and semantic correspondence},
  author={Wang, Pengyuan and Ikeda, Takuya and Lee, Robert and Nishiwaki, Koichi},
  booktitle={European Conference on Computer Vision},
  pages={108--126},
  year={2024},
  organization={Springer}
}

@inproceedings{wen2023bundlesdf,
  title={Bundlesdf: Neural 6-dof tracking and 3d reconstruction of unknown objects},
  author={Wen, Bowen and Tremblay, Jonathan and Blukis, Valts and Tyree, Stephen and M{\"u}ller, Thomas and Evans, Alex and Fox, Dieter and Kautz, Jan and Birchfield, Stan},
  booktitle={Proceedings of the IEEE/CVF Conference on Computer Vision and Pattern Recognition},
  pages={606--617},
  year={2023}
}

@article{chen2023zeropose,
  title={ZeroPose: CAD-model-based zero-shot pose estimation},
  author={Chen, Jianqiu and Sun, Mingshan and Bao, Tianpeng and Zhao, Rui and Wu, Liwei and He, Zhenyu},
  journal={arXiv e-prints},
  pages={arXiv--2305},
  year={2023}
}

@inproceedings{huang2024matchu,
  title={Matchu: Matching unseen objects for 6d pose estimation from rgb-d images},
  author={Huang, Junwen and Yu, Hao and Yu, Kuan-Ting and Navab, Nassir and Ilic, Slobodan and Busam, Benjamin},
  booktitle={Proceedings of the IEEE/CVF Conference on Computer Vision and Pattern Recognition},
  pages={10095--10105},
  year={2024}
}

@inproceedings{caraffa2024freeze,
  title={Freeze: Training-free zero-shot 6d pose estimation with geometric and vision foundation models},
  author={Caraffa, Andrea and Boscaini, Davide and Hamza, Amir and Poiesi, Fabio},
  booktitle={European Conference on Computer Vision},
  pages={414--431},
  year={2024},
  organization={Springer}
}

@inproceedings{nguyen2023cnos,
  title={Cnos: A strong baseline for cad-based novel object segmentation},
  author={Nguyen, Van Nguyen and Groueix, Thibault and Ponimatkin, Georgy and Lepetit, Vincent and Hodan, Tomas},
  booktitle={Proceedings of the IEEE/CVF International Conference on Computer Vision},
  pages={2134--2140},
  year={2023}
}

@article{zhang2024omni6dpose,
  title={Omni6DPose: A Benchmark and Model for Universal 6D Object Pose Estimation and Tracking},
  author={Zhang, Jiyao and Huang, Weiyao and Peng, Bo and Wu, Mingdong and Hu, Fei and Chen, Zijian and Zhao, Bo and Dong, Hao},
  booktitle={European Conference on Computer Vision},
  year={2024},
  organization={Springer}
}

@inproceedings{xiang2018posecnn,
    Author = {Yu Xiang and Tanner Schmidt and Venkatraman Narayanan and Dieter Fox},
    Title = {{PoseCNN}: A Convolutional Neural Network for {6D} Object Pose Estimation in Cluttered Scenes},
    booktitle = {Robotics: Science and Systems (RSS)},
    Year = {2018}
}

@inproceedings{zhang2024omni6d,
  title={Omni6D: Large-Vocabulary 3D Object Dataset for Category-Level 6D Object Pose Estimation},
  author={Zhang, Mengchen and Wu, Tong and Wang, Tai and Wang, Tengfei and Liu, Ziwei and Lin, Dahua},
  booktitle={European Conference on Computer Vision},
  pages={216--232},
  year={2024},
  organization={Springer}
}

@article{teed2021droid,
  title={Droid-slam: Deep visual slam for monocular, stereo, and rgb-d cameras},
  author={Teed, Zachary and Deng, Jia},
  journal={Advances in neural information processing systems},
  volume={34},
  pages={16558--16569},
  year={2021}
}

@inproceedings{runz2018maskfusion,
  title={Maskfusion: Real-time recognition, tracking and reconstruction of multiple moving objects},
  author={Runz, Martin and Buffier, Maud and Agapito, Lourdes},
  booktitle={2018 IEEE international symposium on mixed and augmented reality (ISMAR)},
  pages={10--20},
  year={2018},
  organization={IEEE}
}

@inproceedings{he2022oneposeplusplus,
    title={OnePose++: Keypoint-Free One-Shot Object Pose Estimation without {CAD} Models},
    author={Xingyi He and Jiaming Sun and Yuang Wang and Di Huang and Hujun Bao and Xiaowei Zhou},
    booktitle={Advances in Neural Information Processing Systems},
    year={2022}
}

@article{rauch1965maximum,
  title={Maximum likelihood estimates of linear dynamic systems},
  author={Rauch, Herbert E and Tung, F and Striebel, Charlotte T},
  journal={AIAA journal},
  volume={3},
  number={8},
  pages={1445--1450},
  year={1965}
}

@inproceedings{sturm2012benchmark,
  title={A benchmark for the evaluation of RGB-D SLAM systems},
  author={Sturm, J{\"u}rgen and Engelhard, Nikolas and Endres, Felix and Burgard, Wolfram and Cremers, Daniel},
  booktitle={2012 IEEE/RSJ international conference on intelligent robots and systems},
  pages={573--580},
  year={2012},
  organization={IEEE}
}

@inproceedings{wang2019densefusion,
  title={Densefusion: 6d object pose estimation by iterative dense fusion},
  author={Wang, Chen and Xu, Danfei and Zhu, Yuke and Mart{\'\i}n-Mart{\'\i}n, Roberto and Lu, Cewu and Fei-Fei, Li and Savarese, Silvio},
  booktitle={Proceedings of the IEEE/CVF conference on computer vision and pattern recognition},
  pages={3343--3352},
  year={2019}
}

@inproceedings{xu2022rnnpose,
  title={Rnnpose: Recurrent 6-dof object pose refinement with robust correspondence field estimation and pose optimization},
  author={Xu, Yan and Lin, Kwan-Yee and Zhang, Guofeng and Wang, Xiaogang and Li, Hongsheng},
  booktitle={Proceedings of the IEEE/CVF conference on computer vision and pattern recognition},
  pages={14880--14890},
  year={2022}
}

@inproceedings{zhou2023deep,
  title={Deep fusion transformer network with weighted vector-wise keypoints voting for robust 6d object pose estimation},
  author={Zhou, Jun and Chen, Kai and Xu, Linlin and Dou, Qi and Qin, Jing},
  booktitle={Proceedings of the IEEE/CVF International Conference on Computer Vision},
  pages={13967--13977},
  year={2023}
}

@inproceedings{dang2022learning,
  title={Learning-based point cloud registration for 6d object pose estimation in the real world},
  author={Dang, Zheng and Wang, Lizhou and Guo, Yu and Salzmann, Mathieu},
  booktitle={European conference on computer vision},
  pages={19--37},
  year={2022},
  organization={Springer}
}

@article{bishop2001introduction,
  title={An introduction to the kalman filter},
  author={Bishop, Gary and Welch, Greg and others},
  journal={Proc of SIGGRAPH, Course},
  volume={8},
  number={27599-23175},
  pages={41},
  year={2001}
}

@inproceedings{olson2011apriltag,
  title={AprilTag: A robust and flexible visual fiducial system},
  author={Olson, Edwin},
  booktitle={2011 IEEE international conference on robotics and automation},
  pages={3400--3407},
  year={2011},
  organization={IEEE}
}

@article{ubellacker2024high,
  title={High-speed aerial grasping using a soft drone with onboard perception},
  author={Ubellacker, Samuel and Ray, Aaron and Bern, James M and Strader, Jared and Carlone, Luca},
  journal={npj Robotics},
  volume={2},
  number={1},
  pages={5},
  year={2024},
  publisher={Nature Publishing Group UK London}
}

@article{oquab2023dinov2,
  title={Dinov2: Learning robust visual features without supervision},
  author={Oquab, Maxime and Darcet, Timoth{\'e}e and Moutakanni, Th{\'e}o and Vo, Huy and Szafraniec, Marc and Khalidov, Vasil and Fernandez, Pierre and Haziza, Daniel and Massa, Francisco and El-Nouby, Alaaeldin and others},
  journal={arXiv preprint arXiv:2304.07193},
  year={2023}
}

@inproceedings{kirillov2023segment,
  title={Segment anything},
  author={Kirillov, Alexander and Mintun, Eric and Ravi, Nikhila and Mao, Hanzi and Rolland, Chloe and Gustafson, Laura and Xiao, Tete and Whitehead, Spencer and Berg, Alexander C and Lo, Wan-Yen and others},
  booktitle={Proceedings of the IEEE/CVF international conference on computer vision},
  pages={4015--4026},
  year={2023}
}

@article{qin2023geotransformer,
  title={Geotransformer: Fast and robust point cloud registration with geometric transformer},
  author={Qin, Zheng and Yu, Hao and Wang, Changjian and Guo, Yulan and Peng, Yuxing and Ilic, Slobodan and Hu, Dewen and Xu, Kai},
  journal={IEEE Transactions on Pattern Analysis and Machine Intelligence},
  volume={45},
  number={8},
  pages={9806--9821},
  year={2023},
  publisher={IEEE}
}

@article{umeyama1991least,
  title={Least-squares estimation of transformation parameters between two point patterns},
  author={Umeyama, Shinji},
  journal={IEEE Transactions on Pattern Analysis \& Machine Intelligence},
  volume={13},
  number={04},
  pages={376--380},
  year={1991},
  publisher={IEEE Computer Society}
}

@inproceedings{qi2017pointnet,
  title={Pointnet: Deep learning on point sets for 3d classification and segmentation},
  author={Qi, Charles R and Su, Hao and Mo, Kaichun and Guibas, Leonidas J},
  booktitle={Proceedings of the IEEE conference on computer vision and pattern recognition},
  pages={652--660},
  year={2017}
}

@inproceedings{banerjee2025hot3d,
  title={Hot3d: Hand and object tracking in 3d from egocentric multi-view videos},
  author={Banerjee, Prithviraj and Shkodrani, Sindi and Moulon, Pierre and Hampali, Shreyas and Han, Shangchen and Zhang, Fan and Zhang, Linguang and Fountain, Jade and Miller, Edward and Basol, Selen and others},
  booktitle={Proceedings of the Computer Vision and Pattern Recognition Conference},
  pages={7061--7071},
  year={2025}
}
}
\end{refsection}

\end{document}